\documentclass{article} 
\usepackage{iclr2027_conference,times}

\usepackage{amsmath,amsfonts,bm}

\def\eqref#1{equation~\ref{#1}}

\def\1{\bm{1}}

\def\va{{\bm{a}}}

\def\vh{{\bm{h}}}

\def\vs{{\bm{s}}}

\def\vx{{\bm{x}}}

\def\vz{{\bm{z}}}

\def\evz{{z}}

\def\mA{{\bm{A}}}
\def\mB{{\bm{B}}}
\def\mC{{\bm{C}}}
\def\mD{{\bm{D}}}

\def\mI{{\bm{I}}}

\def\mM{{\bm{M}}}

\def\mW{{\bm{W}}}

\DeclareMathAlphabet{\mathsfit}{\encodingdefault}{\sfdefault}{m}{sl}
\SetMathAlphabet{\mathsfit}{bold}{\encodingdefault}{\sfdefault}{bx}{n}

\def\sU{{\mathbb{U}}}

\usepackage{hyperref}
\usepackage{url}
\usepackage{graphicx}
\usepackage{enumitem}
\usepackage{caption}
\usepackage{multirow}

\title{Continual Learning of Dynamical Systems in Recurrent Neural Networks through Recyclable Unit Gating}

\author{
Sima Hashemi$^{1}$ \qquad
Daniel Durstewitz$^{1,2}$ \qquad
Georgia Koppe$^{1,3,4}$ \\[1ex]
\\
$^{1}$Faculty of Mathematics and Computer Science, Interdisciplinary Center for Scientific Computing,\\
Heidelberg University, Heidelberg, Germany \\[0.5ex]
$^{2}$Department of Theoretical Neuroscience, Central Institute of Mental Health (CIMH), Medical Faculty \\
Mannheim, Heidelberg University, Heidelberg, Germany \\[0.5ex]
$^{3}$Hector Institute for AI in Psychiatry \&
Department of Psychiatry and Psychotherapy, CIMH, \\
Medical Faculty Mannheim, Heidelberg University, Heidelberg, Germany \\[0.5ex]
$^{4}$Hertie Institute for AI in Brain Health, University of Tübingen, Tübingen, Germany \\[1ex]
\texttt{sima.hashemi@iwr.uni-heidelberg.de}
}

\iclrfinalcopy
\begin{document}

\maketitle

\begin{abstract}
Dynamical Systems Reconstruction (DSR) aims to infer models from observed time series that reproduce a system's qualitative long-term behavior. Continual DSR (cDSR) requires learning new systems while preserving previously learned dynamics, yet even small parameter updates in recurrent models can qualitatively alter their behavior over long autonomous rollouts.  We benchmark  established continual learning (CL) methods spanning parameter regularization, replay, and parameter isolation on the fully trainable and interpretable Almost-Linear RNN (AL-RNN).  Parameter isolation preserves earlier dynamics most effectively, but excessive task-specific allocations can rapidly exhaust a fixed-size network. We therefore introduce Continually-Recyclable Unit-Gating (CRUG), which conserves capacity through compact allocation and forward transfer. Differentiable gates trained with an $L_0$-based penalty select task-specific units, while unused units are recycled for subsequent tasks. Directed connections allow later tasks to reuse earlier representations without affecting the dynamics of previously committed units. CRUG achieves the strongest reconstruction--capacity trade-off among the tested methods with zero forgetting and reliably learns a heterogeneous sequence of nonlinear and chaotic systems. Furthermore, we show that forward transfer is more pronounced and useful when tasks share similar underlying dynamics. Lastly, we demonstrate that CRUG's advantages extend beyond autonomous cDSR to sequential cognitive tasks.

\end{abstract}

\section{Introduction}
Across fields such as neuroscience, ecology, and climate science, the goal of modeling time-series data is often not merely to predict the next observation, but to recover the dynamical organization underlying the observed behavior---including its attractors, regimes, and transitions. Dynamical systems reconstruction (DSR) addresses this goal by inferring a generative model whose autonomous evolution reproduces the system's \emph{long-term} behavior \citep{Durstewitz2023}. This distinction is crucial for nonlinear and chaotic systems, where nearby trajectories can rapidly diverge, such that accurate short-term forecasts alone do not establish that the underlying dynamics have been recovered \citep{Wood2010,Koppe2019}. DSR is therefore evaluated by whether freely generated trajectories reproduce the geometry and temporal statistics of the observed system.

Most DSR approaches assume that all relevant data are available during a single training phase, yet distinct systems or dynamical regimes are often encountered sequentially. In neuroscience, for example, population recordings accumulate across sessions as animals learn new tasks or adapt to changing behavioral contexts, potentially revealing distinct dynamical regimes that share computational motifs \citep{DurstewitzAverbeckKoppe2025,Driscoll2024}. Learning from these changes while retaining earlier knowledge is a central goal of continual learning (CL) and a hallmark of biological intelligence \citep{Parisi2019}, but artificial neural networks often suffer from catastrophic forgetting (CF) as updates for new tasks disrupt previously learned representations \citep{McCloskey1989}. Whereas fitting separate models precludes knowledge accumulation within a single model and joint retraining requires access to earlier data, continual DSR (cDSR) aims to incrementally learn new regimes while preserving previously reconstructed dynamics and, where possible, reusing shared dynamical structure. 

Although CL has been studied extensively, most methods and benchmarks concern feedforward classification. Existing studies of CL in recurrent neural networks (RNNs) primarily consider sequence classification, supervised prediction, or working-memory tasks \citep{Cossu2021,Ehret2021}. The few approaches developed specifically for changing dynamical systems (DS) either keep the recurrent reservoir fixed and learn competing readout heads \citep{Bereska2022}, or use an external memory to represent sequentially emerging dynamical modes \citep{Akgul2024}. These studies leave open how established CL strategies behave in a fully trainable recurrent model when success requires preserving autonomous long-term dynamics. This distinction is important because recurrent parameters are applied repeatedly during free simulation, and each small parameter change can qualitatively alter the generated dynamics \citep{hemmer2024}.

We investigate cDSR using the Almost-Linear RNN (AL-RNN), an interpretable piecewise-linear architecture designed for DSR \citep{Brenner2024}. We systematically compare representative methods from three major CL families---parameter regularization, replay, and parameter isolation---based on their ability to preserve a system's autonomous long-term dynamics. 

\paragraph{Contributions.}
Our contributions are threefold: \textbf{(i) Benchmark.} We provide a systematic comparison of established CL methods for cDSR in a fully trainable recurrent model. \textbf{(ii) Method.} We introduce Continually-Recyclable Unit-Gating (CRUG), which learns compact task-specific unit allocations, protects committed dynamics, recycles unused capacity, and permits forward transfer. \textbf{(iii) Findings.} CRUG achieves the strongest reconstruction--capacity trade-off among the tested methods with zero forgetting, while forward transfer improves reconstruction and capacity efficiency when successive systems share dynamical structure. We validate CRUG on synthetic dynamical systems and real-world time series, and show that its benefits extend beyond cDSR to sequential cognitive tasks.

\section{Methods}
We introduce CRUG, a CL method for RNNs that partitions the latent space into disjoint, task-specific subspaces, thereby preventing interference between sequentially learned tasks by construction. Building on prior parameter-isolation methods \citep{Serra2018,Golkar2019,Abati2020}, CRUG combines unit-level gating to learn compact task-specific allocations, protection of earlier dynamics while permitting forward transfer, and recycling of unused units. In principle, this framework is applicable to different recurrent architectures, provided that their parameters and connectivity can be selectively constrained to preserve previously learned dynamics. Here, we instantiate CRUG in the AL-RNN, a piecewise-linear architecture that achieves accurate DSR with parsimonious nonlinearity \citep{Brenner2024, Brenner2026}, while its dynamics are mathematically highly tractable \citep{Brenner2024, Eisenmann2023, Eisenmann2026}. 
We briefly review the architecture before describing CRUG in detail.

\subsection{Preliminaries: AL-RNN architecture}
\label{sec:al-rnn}
The AL-RNN is a recurrent state-space model for reconstructing DS from time-series data \citep{Brenner2024}. It separates a latent dynamical model from an observation mapping that links the latent states to the measured data. Combined with generalized teacher forcing (GTF;\citet{Hess2023}), AL-RNNs can be reliably trained on nonlinear and chaotic dynamics. We use an AL-RNN with $M$ latent units, of which $P$ are nonlinear (ReLU) and the remaining $M - P$ are linear. The latent state $\vz_t \in \mathbb{R}^M$ evolves according to
\begin{align}
\vz_{t+1} &= \mA \vz_t + \mW \Phi^*(\vz_t) + \mC \vs_t + \vh,
\label{eq:al-rnn} \qquad \text{with}\\
\Phi^*(\vz_t) &= \left[
\evz_{1,t}, \dots, \evz_{M-P,t},
\phi(\evz_{M-P+1,t}), \dots, \phi(\evz_{M,t})
\right]^T,
\end{align}
where $\mA = \operatorname{diag}(\va)$, with $\va \in \mathbb{R}^{M}$, is a diagonal matrix; $\mW \in \mathbb{R}^{M \times M}$ contains the recurrent interactions; $\vs_t \in \mathbb{R}^Q$ denotes an optional external input, which is mapped into the latent space by $\mC \in \mathbb{R}^{M \times Q}$; and $\vh \in \mathbb{R}^M$ is a bias vector. The partially nonlinear transformation $\Phi^*(\cdot)$ applies the ReLU activation $\phi(\cdot)$ only to the last $P$ latent units, while the first $M-P$ units remain linear. For CL, at every training step of task $k$, a shared, fixed random encoder $\mB$ maps the initial observation $\vx^{(k)}_0 \in \mathbb{R}^{N_k}$ into the $M$-dimensional latent space, $\vz_0 = \mB\vx^{(k)}_0$ (Fig.~\ref{fig:schematic}D).
The encoder serves only for initialization; subsequent latent states evolve through the recurrent dynamics (see Appendix~\ref{AP:model_initiallization}). Each task $k$ is assigned $N_k$ readout units, and the model output is obtained from their activity via a fixed, task-specific decoder, $\hat{\vx}_t = \mD_k^{\top}\vz_t$ (Fig.~\ref{fig:schematic}D). Neither the encoder nor the decoders are trained.

To stabilize training on chaotic dynamics, we use Sparse Teacher Forcing (STF) with forcing interval $\tau_{\mathrm{STF}}=16$ \citep{Mikhaeil2022,Hess2023}. For task $k$, the control state is defined as
$\tilde{\vz}^{(k)}_t
= \mD_k\vx^{(k)}_t
+ (\mI_M-\mD_k\mD_k^\top)\vz^{(k)}_t$.
At times $t=n\,\tau_{\mathrm{STF}}$, $n \in \mathbb{N}$, the recurrent update uses $\tilde{\vz}^{(k)}_t$ instead of $\vz^{(k)}_t$. This periodically replaces the current task's readout coordinates with the ground-truth observation, leaving all other latent coordinates unchanged and limiting the accumulation of trajectory deviations. 
More generally, forcing methods guide recurrent dynamics using observed states or prediction errors and can improve learning over long temporal horizons \citep{Sagtekin2025,Herz2026}. At test time, the model receives only the initial observation for initialization and subsequently evolves autonomously, without further access to observations and STF.

\subsection{Continually-Recyclable Unit-Gating}
\label{sec:crug}

\begin{figure}[t]
    \centering
    \includegraphics[width=\linewidth]{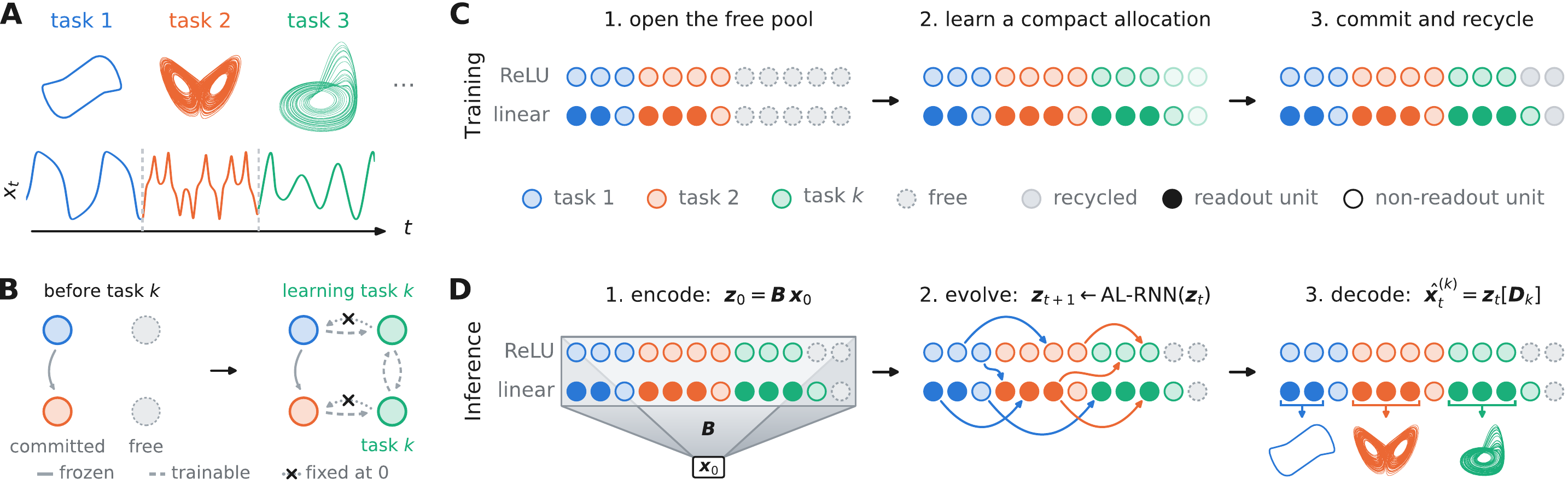}
    \caption{\textbf{Continually-Recyclable Unit-Gating (CRUG).}
\textbf{A)} Dynamical systems are learned sequentially while retaining previously learned dynamics. 
\textbf{B)} During training on task $k$, the connections of previously committed units remain frozen. Forward transfer connections from committed to free units are trainable, whereas reverse connections are held at zero to protect earlier dynamics from interference.
\textbf{C)} All free units initially participate in learning task $k$. During training, the $L_0$ penalty closes unnecessary gates. At the end of the task's training phase, units with gate values above a threshold are committed and frozen, while the remaining units are reinitialized and recycled for task $k+1$. Darker shades indicate higher gate values.
\textbf{D)} For inference after all tasks have been learned, the model is initialized from a given starting point and evolved for $T$ time steps. After the initial transient, all learned dynamical systems run concurrently within the model. Each task's reconstruction can be read out from its corresponding set of readout units. For clarity, only transfer connections are shown; connections among units within each task are omitted.
}
    \label{fig:schematic}
\end{figure}

CRUG is a CL framework that equips the AL-RNN introduced above with adaptive, task-specific unit allocation. For each new task, all uncommitted units form a free pool whose participation is controlled by trainable gates, while an \(L_0\) penalty encourages a compact allocation. After training, selected units are frozen, and the rest are recycled by reinitializing them for subsequent tasks. Committed units may provide input to units learned for later tasks, enabling forward transfer, whereas connections in the reverse direction are fixed to zero to protect earlier dynamics \citep{Golkar2019}. Fig.~\ref{fig:schematic}C illustrates this allocation, commitment, and recycling process.

\paragraph{Unit-level gates.}
To determine how many units a new task requires, CRUG assigns trainable gates to all units still available for allocation (Fig.~\ref{fig:schematic}C). At the start of task $k$, let $\sU_{k-1}^{\mathrm{committed}}$ denote the units reserved for
earlier tasks. The remaining units form the free pool, $\sU_k^{\mathrm{free}}
    =
    \{1,\ldots,M\}
    \setminus
    \sU_{k-1}^{\mathrm{committed}}$,
with $\sU_0^{\mathrm{committed}}=\emptyset$. The free-unit pool consists of a linear-unit pool and a ReLU-unit pool, $\sU_k^{\mathrm{free}} = \sU_k^{\mathrm{lin}} \cup \sU_k^{\mathrm{ReLU}}$.

For a task $k$, the first $N_k$ free linear units are reserved as protected readout units ($\sU^{\mathrm{dec}}_k=\sU_k^{\mathrm{lin}}[0:N_k]$). The fixed task-specific decoder $\mD_k\in\mathbb{R}^{M\times N_k}$ is defined as $(\mD_k)_{j,i}=1$ if $j=\sU^{\mathrm{dec}}_k[i]$ and $0$ otherwise, for $i\in[0,N_k-1]$. Thus, $\hat{\vx}_t^{(k)}=\vz_t\mD_k$ reads out the selected latent coordinates. The gates of these readout units are not trainable and set to $g_i=1.0$. 
All other free units 
receive a trainable gate $g_i \in [0,1]$, whereas committed units retain fixed gates $g_i = 1$, keeping them available for forward transfer to later tasks. 

For each free unit, the gate is parameterized by a trainable logit \(\log\alpha_i\) using a stretched and hard-clipped sigmoid
\begin{align}
    g_i = \min\Big(1,\ \max\big(0,\ \sigma(\log\alpha_i)(\zeta-\gamma)+\gamma\big)\Big),
    \label{eq:hard-concrete}
\end{align}
where $\sigma(\cdot)$ denotes the logistic sigmoid and $\gamma=-0.1$ and $\zeta=1.1$ determine the stretch interval \citep{louizos2018}. At the beginning of each task, we initialize $\log\alpha_i=2$, corresponding to $g_i \approx0.96$. 
Because the gates are initialized near one, all free units initially contribute almost fully to the recurrent dynamics. At the same time, the gate parameters remain away from saturation, allowing the capacity penalty defined below to drive unnecessary gates toward zero.

Each gate acts on the corresponding latent unit and its incident connections. Specifically, the effective parameters used in the AL-RNN update are
\begin{align}
\widetilde A_{ii}
&= g_i A_{ii},
&
\widetilde W_{ij}
&= g_i g_j W_{ij},
&
\widetilde C_{iq}
&= g_i C_{iq},
&
\widetilde h_i
&= g_i h_i.
\label{eq}
\end{align}
The factor $g_i$ suppresses the update of target unit $i$, while the factor $g_j$ suppresses its contribution as a source to other recurrent units. Hence, setting $g_i=0$ completely removes unit $i$ from the recurrent computation. During training, the gated parameters
\(\widetilde{\theta}
=
\bigl(
\widetilde{\mA},
\widetilde{\mW},
\widetilde{\mC},
\widetilde{\vh}
\bigr)
\)
replace $\theta=(\mA,\mW,\mC,\vh)$ in the AL-RNN updates.

\paragraph{Capacity regularization.}
To encourage each task to use fewer units, we add an $L_0$-based penalty on the gates that encourages gates to close unless keeping them open improves task performance \citep{louizos2018}. We use separate penalty strengths for the free linear and ReLU units:
\begin{align}
    \mathcal{L}_{\mathrm{cap}}^k
    &=
    \lambda_{L_0}^{\mathrm{lin}}
    \sum_{i\in\sU_k^{\mathrm{lin}}\setminus
    \sU_{k}^{\mathrm{dec}}}
    \sigma\left(
        \log\alpha_i-\log\frac{-\gamma}{\zeta}
    \right)
    +    \lambda_{L_0}^{\mathrm{ReLU}}
    \sum_{i\in\sU_k^{\mathrm{ReLU}}}
    \sigma\left(
        \log\alpha_i-\log\frac{-\gamma}{\zeta}
    \right).
    \label{eq:L0-loss}
\end{align}
Minimizing the capacity penalty lowers \(\log\alpha_i\), moving \(g_i\) toward zero, whereas the task loss keeps the gate open when unit \(i\) contributes to accurate reconstruction. Thus, only units whose contribution outweighs their capacity cost tend to remain above a commitment threshold. 
Because a uniform penalty may favor the more expressive ReLU units, we penalize them more strongly to limit nonlinearity and simplify the resulting state-space partition.

\paragraph{Protection and forward transfer.}
During training, CRUG preserves the dynamics of previously committed units while allowing the new task to use their activity (Fig.~\ref{fig:schematic}B). Connections from committed source units to free target units remain trainable, while connections in the reverse direction are set to zero \citep{Golkar2019}. All parameters governing the updates of committed units are frozen. 

Let 
$\mathbf{f}_k\in\{0,1\}^M$ be the indicator vector of $\sU_k^{\mathrm{free}}$. Because $\mW$ is indexed as $[\mathrm{target},\mathrm{source}]$, only rows corresponding to free target units remain plastic. We apply the gradient masks
\begin{align}
    \nabla_{\va}\mathcal{L}
    &\leftarrow
    \nabla_{\va}\mathcal{L}\odot\mathbf{f}_k,
    &
    \nabla_{\mW}\mathcal{L}
    &\leftarrow
    \nabla_{\mW}\mathcal{L}
    \odot
    \left(\mathbf{f}_k\mathbf{1}_M^\top\right),
    \nonumber\\
    \nabla_{\mC}\mathcal{L}
    &\leftarrow
    \nabla_{\mC}\mathcal{L}
    \odot
    \left(\mathbf{f}_k\mathbf{1}_Q^\top\right),
    &
    \nabla_{\vh}\mathcal{L}
    &\leftarrow
    \nabla_{\vh}\mathcal{L}\odot\mathbf{f}_k.
    \label{eq:freeze}
\end{align}

To control the strength of forward transfer, we additionally regularize connections from committed source units to free target units. The transfer from source unit \(j\in\sU_{k-1}^{\mathrm{committed}}\) to target unit \(i\in\sU_k^{\mathrm{free}}\) is defined as $\mathcal{T}_{ij}:=\widetilde W_{ij} = g_i g_j W_{ij} = g_i W_{ij}$ (since the gate of every committed unit is fixed at $g_j=1$). To prevent inputs through transfer connections from becoming excessively large, different regularization schemes can be applied to these weights, including $L_0$, $L_1$, and $L_2$ regularization (see Appendix~\ref{AP:Transfer mechanism}). 
We systematically examine how the choice of transfer regularization affects reconstruction quality and capacity usage in Section~\ref{sec:ResultsForwardTransfer}. Unless stated otherwise, all results reported in this paper are obtained using $L_2$ regularization, with
\begin{align}
\mathcal{L}_{\mathrm{tr}}^k
=
\lambda_{\mathrm{tr}}
\sum_{\substack{
i\in\sU_k^{\mathrm{free}},\
j\in\sU_{k-1}^{\mathrm{committed}}
}}
\left(\mathcal{T}_{ij}\right)^2.
\label{eq}
\end{align}
The complete objective for task $k$ is then
\begin{align}
    \mathcal{L}^k
    =
    \mathcal{L}_{\mathrm{task}}^k
    +
    \mathcal{L}_{\mathrm{cap}}^k
    +
    \mathcal{L}_{\mathrm{tr}}^k
    \label{eq:crug-loss}
\end{align}
where $\mathcal{L}_{\mathrm{task}}^k$ is the reconstruction loss on task
$k$, i.e.\ the mean squared error between the model's predictions and
the observed data.

\paragraph{Commitment and recycling.}
After training task $k$, units with gates exceeding $\tau_g=0.5$ are retained, and the remaining units are released:
\begin{align}
    \mathcal{K}_k
    &=
    \left\{
        i\in\sU_k^{\mathrm{free}}:g_i>\tau_g
    \right\},
    &
    \mathcal{R}_k
    &=
    \sU_k^{\mathrm{free}}\setminus\mathcal{K}_k.
    \label{eq:kept-released}
\end{align}
For retained units, we preserve the learned gate magnitudes because they contribute to the trained dynamics. We therefore define 
\begin{align}
    \bar g_i
    =
    \begin{cases}
        g_i, & i\in\mathcal{K}_k,\\
        0, & i\in\mathcal{R}_k,\\
        1, & i\in\sU_{k-1}^{\mathrm{committed}},
    \end{cases}
    \label{eq:final-gates}
\end{align}
and absorb these factors into the stored parameters:
\begin{align}
    A_{ii}
    &\leftarrow\bar g_iA_{ii},
    &
    W_{ij}
    &\leftarrow\bar g_i\bar g_jW_{ij},
    &
    C_{iq}
    &\leftarrow\bar g_iC_{iq},
    &
    h_i
    &\leftarrow\bar g_ih_i.
    \label{eq:bake}
\end{align}
The retained units are added to the committed pool, $\sU_k^{\mathrm{committed}}
    =
    \sU_{k-1}^{\mathrm{committed}}\cup\mathcal{K}_k$.
Released units are reinitialized for subsequent tasks, subject to the connectivity constraints protecting committed units (see Appendix~\ref{AP:model_initiallization} for details of the reinitialization procedure). Specifically, since $\bar g_j=0$ for each released source unit $j$, Eq.\ref{eq:bake} sets its connections into committed target units to zero. These connections are excluded from reinitialization and remain fixed at zero. Connections in the opposite direction, from committed units into recycled units, which constitute the forward-transfer block, are reinitialized to zero and remain available for learning. The released units form the next free pool, $\sU_{k+1}^{\mathrm{free}}=\mathcal{R}_k$, making their capacity available for task $k+1$ (Fig.~\ref{fig:schematic}C).

\section{Related Work}
\label{sec:related_work}
\paragraph{Continual learning of recurrent dynamics.}
Several studies address continual learning in dynamical systems directly. \citet{Bereska2022} use a fixed reservoir with competing readout heads, whereas \citet{Akgul2024} use a Bayesian state-space model with external memory to represent multiple dynamical modes. Their evaluations focus on one-step prediction and finite-horizon forecasting, respectively, rather than preservation of long-term autonomous dynamics. \citet{Liu2026} investigate the emergence of orthogonal task manifolds through a local, error-driven learning rule in RNNs trained on natural movie replay tasks. These approaches leave open how established CL strategies perform when the recurrent dynamics are fully trainable and must remain faithful to previously learned systems over long autonomous rollouts. We therefore compare representative methods from three major CL families---parameter regularization, replay, and parameter isolation---within the same AL-RNN architecture.

\paragraph{Parameter regularization and replay.}
Parameter-regularization methods mitigate forgetting by constraining changes to parameters considered important for previous tasks. Elastic Weight Consolidation (EWC) estimates parameter importance using the Fisher information and penalizes deviations from earlier solutions \citep{Kirkpatrick2017}, whereas Synaptic Intelligence (SI) estimates importance online from each parameter's contribution to reducing the training loss \citep{Zenke2017}. Both methods seek a balance between retaining earlier solutions and preserving sufficient plasticity for subsequent tasks. Replay methods instead interleave data from previous tasks with current-task training. Experience Replay (ER) stores and reuses earlier observations \citep{Chaudhry2019}, whereas Generative Replay (GR) replaces stored data with samples generated by a learned model \citep{shin2017}. Here, the AL-RNN itself provides the generative model required for GR.

\paragraph{Parameter isolation and capacity allocation.} Rather than constraining updates or replaying earlier data, parameter-isolation methods reduce interference by allocating and protecting distinct subsets of
network parameters for different tasks 
\citep{Serra2018,Masse2018,Golkar2019,Abati2020,Mallya2018,Yoon2018,Rusu2016}. Approaches differ in how they allocate capacity. Progressive Neural Networks add a new network column for each task \citep{Rusu2016}, Dynamically Expandable Networks expand capacity as needed \citep{Yoon2018}, and PackNet prunes individual weights and freezes those retained for earlier tasks \citep{Mallya2018}.
Other methods gate at the 
unit level. Context-dependent gating (XdG) assigns fixed random masks to tasks 
\citep{Masse2018}, whereas Learned XdG (LXdG) replaces the random draw by learning a gating network to select task-relevant units \citep{Tilley2023}. Hard Attention to the Task (HAT) learns and stores task-specific unit masks using annealed sigmoid gates, with a weighted $L_1$ penalty on previously unused units to promote reuse of existing features \citep{Serra2018}. 
CCGN and CLNP, by contrast, explicitly recycle capacity. CCGN learns input-dependent binary channel gates using a Gumbel-softmax estimator, and reinitializes unused channels after training \citep{Abati2020}. CLNP instead combines $L_1$ weight regularization with post-training pruning of units based on their average activity, freeing unused units for subsequent tasks \citep{Golkar2019}. Beyond reclaiming capacity for later tasks, reinitialization can help maintain plasticity, as demonstrated by continual replacement of low-utility units during prolonged training \citep{Dohare2024}.

\paragraph{Relation to CRUG.}
CRUG belongs to the family of unit-level parameter-isolation methods. Like XdG, it reduces interference by assigning subsets of units to tasks; however, XdG draws its masks at random before training, independently of task difficulty, and tasks may therefore still share units \citep{Masse2018}. CRUG instead learns how many units each task requires, and units committed to earlier tasks cannot be updated by later ones. Similar to HAT, CRUG uses learned gates to protect capacity devoted to earlier tasks \citep{Serra2018}. Unlike HAT, however, CRUG does not store task-specific masks for inference: all tasks are embedded in a single dynamical system and run concurrently. Finally, like CLNP, CRUG explicitly recycles capacity: units left unused after training are reinitialized and made available to subsequent tasks \citep{Golkar2019}. The two methods differ, however, in how this allocation is determined. CRUG learns task-level unit gates jointly with the task loss under an $L_0$-based capacity penalty, whereas CLNP applies $L_1$ weight regularization followed by activation-based pruning. Unlike an $L_1$ weight penalty, the $L_0$ capacity penalty encourages compact allocations without directly shrinking the weights of retained units. This distinction is particularly relevant for RNNs in DSR tasks, in which even small weights can have a large effect on dynamics \citep{hemmer2024}. 

\section{Experiments}

We evaluate cDSR in terms of retention, reconstruction–capacity trade-offs, and forward transfer. We first benchmark CRUG against established CL methods on 
sequences of dynamical systems adapted from \citet{Gilpin2023} (see Appendix~\ref{AP:Dynamical Systems} for tasks definitions). 
Next, we evaluate CRUG on real-world time series and sequential cognitive tasks to assess its generalization beyond synthetic data and cDSR. Finally, we examine whether shared dynamical structure enhances forward transfer.

In each setting, the specified CL method is applied to an AL-RNN trained sequentially across all tasks (see Appendix~\ref{AP:Continual_Learning_Methods} for all tested methods). Each task is evaluated immediately after its own training phase and again after completion of the full sequence. Reconstruction quality is evaluated using two well-established complementary geometrical and temporal measures for DSR tasks: the state-space divergence $D_{\mathrm{stsp}}$ and the Hellinger distance $D_H$ \citep{Koppe2019,Mikhaeil2022}. \(D_{\mathrm{stsp}}\) measures whether long autonomous rollouts visit the same regions of state space with similar frequencies as the observed system, assessing how well the model reproduces its attractor geometry and distribution of states. \(D_H\) compares the power spectra of observed and generated trajectories, assessing whether the model preserves their characteristic frequencies and distribution of power across timescales over long rollouts. Lower values indicate better reconstruction for both measures. Full definitions and implementation details are provided in Appendix~\ref{AP:Evaluation_Meausres}.

\subsection{Benchmarking CRUG against Existing CL Methods}
\label{sec:ResultsBenchmark}
\paragraph{Synthetic dynamical systems} We first evaluate cDSR on a sequence of synthetic dynamical systems: Van der Pol $\rightarrow$ Lorenz–63 $\rightarrow$ R\"ossler $\rightarrow$ Chua. These systems span qualitatively distinct dynamics and attractor geometries, including a stable limit cycle and several forms of chaotic behavior. For this benchmark, we use an AL-RNN with $(M,P)=(160,80)$. As reference points, we trained (1) a model on all four dynamical systems in an \emph{interleaved} setting where CF is less expected, and (2) another model trained sequentially using naive fine-tuning without any CL mechanism, which exhibits extensive CF. Table~\ref{tab:DSR_canonical_tasks} compares reconstruction performance for all tested methods immediately after each task is learned with performance after completion of the full task sequence.

\begin{figure}[htbp!]
\begin{center}
\includegraphics[width=\linewidth]{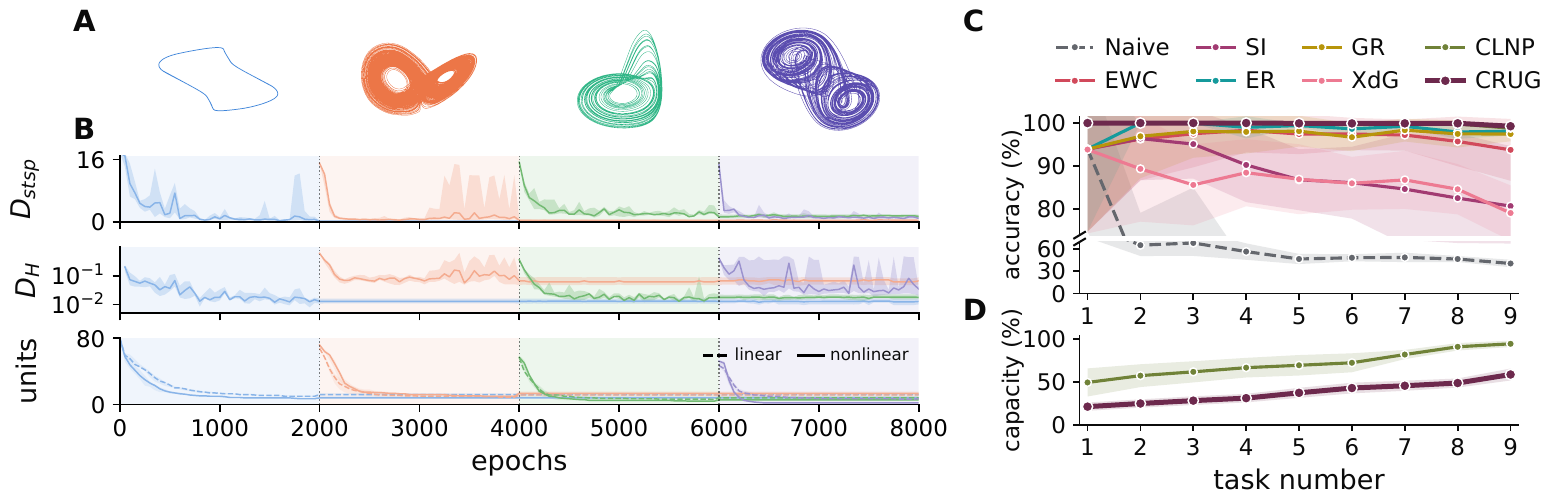}
\end{center}
\caption{\textbf{Benchmarking CRUG against other CL methods.} \textbf{A–B) Dynamical systems reconstruction.} Four dynamical systems, ranging from limit-cycle to chaotic dynamics, are learned sequentially. \textbf{A)} Rollout trajectories generated by the final model for all four tasks. \textbf{B)} $D_{\text{stsp}}$, $D_H$, and the number of committed linear (dashed) and nonlinear (solid) units throughout training. Gray vertical lines mark task boundaries; lines and shaded regions show the median and interquartile range. \textbf{C–D) Cognitive tasks.} Nine cognitive tasks are learned sequentially using each CL method. \textbf{C)} Mean final accuracy as a function of the number of tasks learned. \textbf{D)} Mean capacity usage as a function of the number of tasks learned for the parameter-isolation methods CLNP and CRUG. CRUG achieves the highest final accuracy while using substantially less network capacity.}
\label{fig:CRUG_results}
\end{figure}

The parameter-regularization methods exhibit pronounced forgetting: although EWC and SI learn the earlier tasks, their final reconstructions are poor or divergent. Replay improves retention of the first two systems, but performance deteriorates substantially on R\"ossler, with $D_{\mathrm{stsp}}$  increasing from
$2.67$ to $10.11$ for ER and from $3.09$ to $4.90$ for GR. Although XdG reduces forgetting, it often fails to learn the tasks in the first place. These results indicate that none of these methods achieves a satisfactory stability--plasticity trade-off in this setting. Adaptive parameter isolation methods (CLNP and CRUG) provide better retention, although the two methods differ in their reliability and capacity usage. CRUG achieved lower median overall $D_{\mathrm{stsp}}$ ($1.26$ vs.\ $3.81$, $p=0.0028$), $D_H$ ($0.08$ vs.\ $0.14$, $p=0.0079$), and capacity usage ($63$ vs.\ $123.5$, $p=0.037$). Also, CRUG completed the full task sequence in all 10 runs, whereas CLNP completed seven, with three runs exhausting the free-unit pool.
CRUG also achieved lower $D_{\mathrm{stsp}}$ than interleaved training ($1.26$ vs.\ $3.37$, $p=0.038$), primarily through improved R\"ossler reconstruction (Table~\ref{tab:DSR_canonical_tasks}). Together, these results favor CRUG in reconstruction
quality and reliable sequence completion with limited capacity. Fig.~\ref{fig:CRUG_results}A--B shows example reconstructions of the final model using CRUG method, together with reconstruction measures and capacity usage over the course of training.

To rule out that our findings were related to \emph{task order}, we evaluated three alternative permutations of the original task sequence: Van der Pol, Lorenz–63, R\"ossler, and Chua. CRUG achieved comparable performance across all tested task orders 
(see Appendix~\ref{AP:task order}). To further rule out that our findings were related to \emph{task type}, 
we conducted an additional experiment using four other chaotic systems from the database of \citet{Gilpin2023}: Blasius, Laser, Genesio–Tesi, and Finance, as described in Appendix~\ref{AP:Dynamical Systems}. The failure modes of the first benchmark persist: parameter regularization (EWC and SI) and random gating (XdG) methods either diverge or lose plasticity, while CLNP avoids forgetting at the cost of poor reconstruction and exhausted capacity. CRUG learns the full sequence without pool exhaustion and with zero forgetting, and remains competitive with replay-based methods while committing only ${\sim}60\%$ of the network units and storing no replay data. 
Unlike in the first benchmark, GR achieves slightly lower median $D_{\mathrm{stsp}}$ and $D_H$ values than CRUG, although neither difference is statistically significant and GR uses the full network capacity. Further details are provided in Appendix~\ref{AP:CRUG on Synthetic Dynamical Systems}.

\begin{table}[htbp!]
\caption{\textbf{Benchmarking CRUG against other CL methods on cDSR.}
Each entry in the four first columns reports the $D_{\mathrm{stsp}}$ measure immediately after the task’s
own training phase $\rightarrow$ after all tasks have been learned
sequentially (except for interleaved training). Values are reported in the format
$\text{median}^{\,Q_3-\text{median}}_{\,Q_1-\text{median}}$ across 10 runs
(lower is better). Divergent or incomplete runs were assigned the worst score when computing the median.
Overall performance is the maximum final task $D_{\text{stsp}}$ and  $D_{H}$ within tasks in each
run, summarized across all runs.
``div.'' denotes an infinite median, and ${}^\dagger$ denotes
an infinite upper quartile.
Chua is learned last and therefore has no subsequent forgetting period. For CLNP method $3/10$ runs failed to complete the sequence. }
\label{tab:DSR_canonical_tasks}
\centering
\renewcommand{\arraystretch}{1.5}
\resizebox{\textwidth}{!}{%
\begin{tabular}{l|cccc|cc}

\hline

& \multicolumn{4}{c|}{\textbf{$D_{\mathrm{stsp}}$}}

& \multicolumn{2}{c}{\textbf{Overall performance}} \\

\cline{2-5}\cline{6-7}

\textbf{Method}

& \textbf{Van der Pol}

& \textbf{Lorenz-63}

& \textbf{R\"ossler}

& \textbf{Chua}

& \textbf{$D_{\mathrm{stsp}}\downarrow$}

& \textbf{$D_H\downarrow$} \\

\hline
& & & & & & \\
Interleaved
& $0.05^{+0.18}_{-0.02}$
& $0.25^{+0.06}_{-0.07}$
& $3.37^{+1.69}_{-1.68}$
& $0.90^{+0.48}_{-0.18}$
& $3.37^{+1.69}_{-1.65}$
& $0.05^{+0.00}_{-0.00}$ 
\\

Naive
& $0.02^{+0.01}_{-0.01}\!\rightarrow\!\mathrm{div.}$
& $0.20^{+0.02}_{-0.03}\!\rightarrow\!\mathrm{div.}$
& $1.00^{+0.52}_{-0.13}\!\rightarrow\!\mathrm{div.}$
& $0.69^{+0.10}_{-0.10}$
& $\mathrm{div.}$
&$\mathrm{div.}$
\\

EWC
& $0.02^{+0.01}_{-0.00}\!\rightarrow\!\mathrm{div.}$
& $0.20^{+0.15}_{-0.01}\!\rightarrow\!\mathrm{div.}$
& $3.20^{+1.74}_{-1.38}\!\rightarrow\!15.35^{\dagger}_{-0.07}$
& $1.15^{+0.40}_{-0.25}$
& $\mathrm{div.}$
&$\mathrm{div.}$
\\
SI 
& $0.02^{+0.01}_{-0.01}\!\rightarrow\!0.50^{+0.62}_{-0.24}$
& $15.00^{+0.02}_{-0.03}\!\rightarrow\!14.99^{+0.03}_{-0.02}$
& $15.20^{\dagger}_{-0.70}\!\rightarrow\!15.19^{+0.35}_{-0.66}$
& $14.51^{+0.06}_{-0.06}$
& $15.19^{+0.35}_{-0.19}$
&$\mathrm{div.}$
\\

ER
& $0.02^{+0.01}_{-0.01}\!\rightarrow\!0.04^{+0.17}_{-0.02}$
& $0.24^{+0.04}_{-0.04}\!\rightarrow\!0.21^{+0.07}_{-0.03}$
& $2.67^{+0.94}_{-0.90}\!\rightarrow\!10.11^{+0.66}_{-2.20}$
& $0.82^{+0.37}_{-0.08}$
& $10.11^{+0.66}_{-2.20}$
& $0.04^{+0.01}_{-0.00}$
\\

GR 
& $0.02^{+0.01}_{-0.00}\!\rightarrow\!0.14^{+0.15}_{-0.11}$
& $0.21^{+0.03}_{-0.03}\!\rightarrow\!0.31^{+0.31}_{-0.10}$
& $3.09^{+0.89}_{-1.85}\!\rightarrow\!4.90^{+4.84}_{-3.01}$
& $0.73^{+0.06}_{-0.14}$
& $7.39^{+5.08}_{-5.49}$
& $0.05^{+0.02}_{-0.01}$
\\

XdG
& $14.20^{+2.27}_{-0.33}\!\rightarrow\!17.70^{\dagger}_{-1.22}$
& $10.39^{+3.83}_{-6.17}\!\rightarrow\!10.39^{+3.86}_{-6.17}$
& $4.66^{+1.04}_{-1.15}\!\rightarrow\!4.66^{+1.04}_{-1.15}$
& $14.65^{+0.01}_{-13.95}$
& $17.70^{\dagger}_{-1.22}$
& $\text{div.}$ \\

CLNP& 
$2.59^{\dagger}_{-1.54}\!\rightarrow\! 2.59^{\dagger}_{-1.54}$ & $2.15^{\dagger}_{-0.51}\!\rightarrow\!2.15^{\dagger}_{-0.51}$ & $3.53^{\dagger}_{-1.22}\!\rightarrow\! 3.53^{\dagger}_{-1.29}$ &
$2.44^{\dagger}_{-0.07}$ & $3.81^{\dagger}_{-1.09}$ & 
$0.14^{\dagger}_{-0.01}$\\

\hline
CRUG
& $0.15^{+0.16}_{-0.06}\!\rightarrow\!0.15^{+0.16}_{-0.06}$
& $0.40^{+0.13}_{-0.12}\!\rightarrow\!0.41^{+0.06}_{-0.12}$
& $1.22^{+1.02}_{-0.21}\!\rightarrow\!1.26^{+1.16}_{-0.20}$
& $0.98^{+0.12}_{-0.17}$
& $\mathbf{1.26^{+1.16}_{-0.18}}$
& $\mathbf{0.08^{+0.01}_{-0.01}}$
\\
\hline
\end{tabular}}
\end{table}


\paragraph{Real-world data} We next tested whether CRUG can learn recordings from different subjects sequentially within a single model while preserving earlier reconstructions and using capacity efficiently. We used the \textit{Gait in Neurodegenerative Disease Database}, which contains bilateral foot-sensor recordings from patients with Parkinson's disease, Huntington's disease, and amyotrophic lateral sclerosis, as well as healthy controls \citep{Hausdorff2000,dataset}. We trained an AL-RNN with $(M,P)=(1024,512)$ with CRUG on 12 subjects, three per group, treating each subject as a separate task (Appendix~\ref{AP:Gait_tasks}). CRUG retained all subject-specific gait dynamics without measurable forgetting (median $\widetilde{D}_{\text{stsp}} = 2.12$; Appendix~\ref{AP:Evaluation_Meausres}) while using only $\approx 48\%$ of the total capacity. Compared with four separately trained group-specific models, the joint model achieved $27\%$ better reconstruction with $30\%$ fewer committed units (Table~\ref{tab:gait_joint_vs_group}), showing that forward transfer enables CRUG to accommodate more tasks in a single model without sacrificing performance.

\paragraph{Cognitive tasks} To demonstrate that CRUG is not specific to cDSR, we finally also evaluated it on a sequential suite of nine cognitive benchmark tasks adapted from \citet{Driscoll2024}. These include pro- and anti-response variants of delayed-response, reaction-time, and category-decision tasks requiring context-dependent mirror-image mappings; two context-integration tasks requiring selective attention to one of two sensory modalities; and a Go–NoGo task probing simple threshold detection (see Appendix~\ref{AP:Cognitive_Tasks} for further task details). We used an AL-RNN with $(M,P)=(120,60)$, trained sequentially for 50 epochs per task, and evaluated task accuracy after completing the full sequence. The benchmark results are reported in Fig.~\ref{fig:CRUG_results}C–D and Appendix Table~\ref{tab:cognitive_tasks}. 
Although, as may be expected, all CL methods perform better on the cognitive tasks than on the cDSR benchmark, CRUG achieves the highest overall final accuracy while using only approximately 30\% of the available model capacity (Fig.~\ref{fig:CRUG_results}D). All other methods use the full capacity (except CLNP, which uses approximately 80\%) while achieving lower final accuracy.

\subsection{When does forward transfer improve continual DSR?}
\label{sec:ResultsForwardTransfer}

\begin{figure}[htbp!]
\begin{center}
\includegraphics[width=\linewidth]{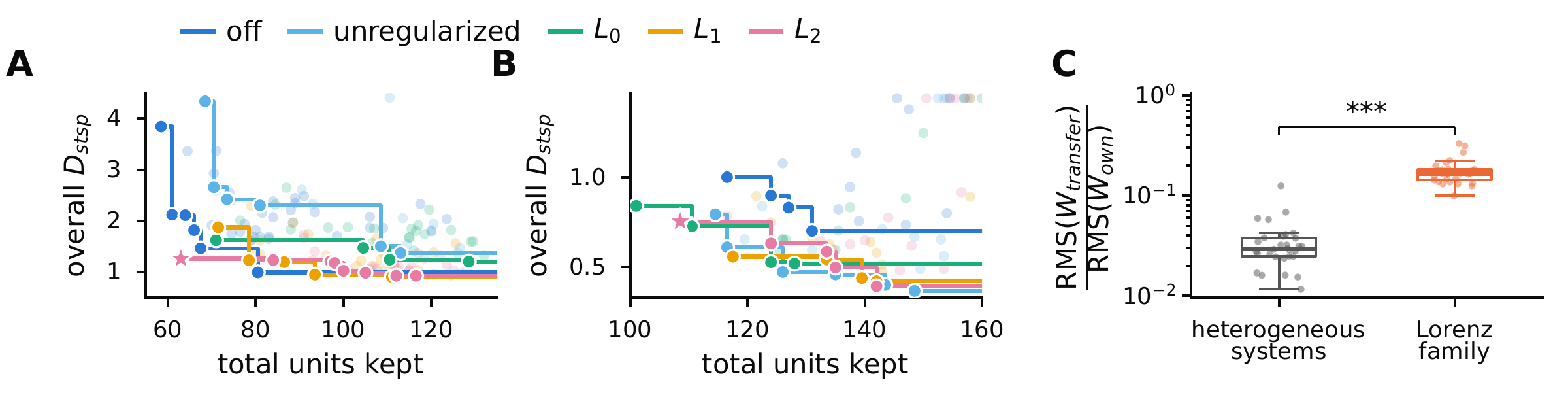}
\end{center}
\caption{\textbf{Forward-transfer mechanisms analysis.}
\textbf{A, B)} Final overall $D_{\text{stsp}}$ (worst task after sequential
training on four tasks) versus total units retained; lower is better.
Faded points show individual hyperparameter configurations (median over
10 seeds), and step curves their Pareto fronts. Stars mark runs analyzed
in C. \textbf{A)} Heterogeneous sequence: transfer provides
no clear advantage. \textbf{B)} Related sequence (Lorenz-63,
$\rho=28\rightarrow313\rightarrow200\rightarrow160$): all transfer
mechanisms outperform no transfer. \textbf{C)} Root-mean-square weights, $\mathrm{RMS}(W)$, of the task-specific and transfer blocks for the starred runs (tasks 2–4; 10 seeds). Bracket indicates two-sided Mann–Whitney $U$ tests; *** $p<0.001$.}
\label{fig:DSR_task_pareto_fronts}
\end{figure}

CRUG prevents later tasks from altering previously committed units while allowing those units to provide input to newly allocated ones. This forward-transfer mechanism enables later tasks to reuse representations and dynamical computations learned from earlier systems \citep{Golkar2019}. The benefit of such reuse is expected to depend on the amount of dynamical structure shared across tasks. To investigate this, we introduced an additional task sequence consisting of four Lorenz–63 systems governed by the same underlying equations but differing in the control parameter, $\rho \in \{28, 313, 200, 160\}$, corresponding to a two-lobe chaotic regime, a single-loop periodic regime, a single-lobe chaotic regime, and a double-loop periodic regime, respectively (see Appendix~\ref{AP:Lorenz_family}). We evaluated both task sequences by comparing CRUG without transfer against four transfer variants: unregularized transfer, $L_0$-gated transfer, $L_1$-regularized transfer, and $L_2$-regularized transfer. The corresponding hyperparameter search ranges are reported in Appendix~\ref{AP:Transfer mechanism}. For each mechanism, we evaluated 40 hyperparameter configurations with $10$ seeds per configuration (Appendix Table~\ref{tab:transfer-hyperparameters}). Because reconstruction quality and retained capacity are competing objectives, we compared the resulting Pareto fronts over final overall $D_{\mathrm{stsp}}$ and the number of committed units in Fig.~\ref{fig:DSR_task_pareto_fronts}. 

For the heterogeneous sequence Van der Pol $\rightarrow$ Lorenz–63 $\rightarrow$ R\"ossler $\rightarrow$ Chua, transfer provides no clear advantage over the no-transfer condition (Fig.~\ref{fig:DSR_task_pareto_fronts}A). In contrast, for the Lorenz-family task sequence, all transfer-based variants achieve more favorable reconstruction–capacity trade-offs than the no-transfer condition (Fig.~\ref{fig:DSR_task_pareto_fronts}B). To further investigate the contribution of the transfer connections in these two task sequences, we selected the best trial using the $L_2$ transfer mechanism for each sequence (indicated by stars in Figs.~\ref{fig:DSR_task_pareto_fronts}A and B). For tasks 2–4, where transfer connections from previously committed units are available, we compared the ratio of root mean square (RMS) of the transfer weights to the within-task weights connecting the task’s own units. This ratio is significantly bigger for the Lorenz family task sequence (Fig.~\ref{fig:DSR_task_pareto_fronts}C). Together with the improved Pareto front, this pattern is consistent with stronger recruitment of forward-transfer pathways when tasks share dynamical structure. 

\section{Discussion} 

Learning to adapt to changing environments without forgetting previously acquired knowledge is a pressing challenge in real-world applications, where retraining from scratch after every change is often costly or infeasible. Motivated by this, we evaluated established CL methods for cDSR and introduced CRUG, which preserves learned dynamics while conserving recurrent capacity. Parameter-isolation methods retained earlier dynamics best, showing the value of protecting recurrent computations rather than merely constraining parameter changes, but at a cost in capacity. CRUG reduces this cost through compact unit allocation and recycling, achieving the best reconstruction–capacity trade-off among the tested methods. Its directed connectivity lets later tasks reuse earlier computations without altering them, which helps most when successive tasks share similar dynamics. Finally, experiments on real-world data and cognitive tasks show that CRUG generalizes beyond synthetic systems and beyond cDSR.

\paragraph{Limitations.}
Our experiments assume task-incremental settings with known task boundaries and identities. Recycling also delays but does not prevent capacity exhaustion, since committed units are never released or consolidated; combining generative replay with consolidation of task-specific subnetworks could recover redundant capacity without retaining original data \citep{Brenner2025,Schwarz2018}. The learned allocation should not be taken as a system's intrinsic minimum capacity, as it depends on optimization, regularization, and previously learned representations. Finally, although improved transfer within the Lorenz family suggests reuse of shared dynamical structure, what exactly is transferred remains unclear; analyzing shared activation patterns and piecewise-linear regimes could clarify this and enable adaptive transfer.

\section{ Acknowledgments}
This work was funded by the Federal Ministry of Research, Technology  and Space (BMFTR) through the NAILIt project under the neuroAI initiative (grant numbers 01GQ2509B and 01GQ2509A), by the Wellcome Trust  (FUTURE-D, Z334349/Z/25/Z), and by the Excellence Strategy of the German Federal and State Governments.

\bibliography{reference}
\bibliographystyle{iclr2027_conference}
 \newpage
\appendix

\section{Appendix}

\subsection{Model Initializations}
\label{AP:model_initiallization}

\paragraph{Shared fixed encoder.}
A single encoder
$\mB\in\mathbb{R}^{M\times N_{\max}}$, where $N_{\max}=\max_k N_k$ (maximal task dimension)
is sampled once per run using the run's random seed, with
independent entries
\begin{equation}
B_{ji}\sim\mathcal{U}\left(
-\frac{1}{\sqrt{N_{\max}}},
\frac{1}{\sqrt{N_{\max}}}
\right),
\end{equation}
where $\mathcal{U}$ denotes the uniform distribution.
Tasks of lower dimension, $N_k<N_{\max}$, share this encoder by
zero-padding: an observation $\vx_t^{(k)}\in\mathbb{R}^{N_k}$ is extended
to
\begin{equation}
\tilde{\vx}_t^{(k)}=
\begin{pmatrix}\vx_t^{(k)}\\ \mathbf{0}_{N_{\max}-N_k}\end{pmatrix}
\in\mathbb{R}^{N_{\max}},
\end{equation}
where $\mathbf{0}_{d}$ is the $d$-dimensional zero vector. The encoder
is then applied as $\vz_t = \mB\tilde{\vx}_t^{(k)}$. The padded coordinates
multiply the last $N_{\max}-N_k$ columns of $\mB$ and contribute
nothing, so each task effectively uses only the first $N_k$ columns of the
same fixed matrix. When $N_k=N_{\max}$, no padding is applied.



\paragraph{Parameter initialization.}

At the beginning of each run, we construct a symmetric positive-definite matrix from an \(M\times M\) matrix of independent standard Gaussian values. We normalize this matrix by its largest eigenvalue and use the resulting diagonal entries to initialize the diagonal of $\mA=\operatorname{diag}(\va)$, setting all off-diagonal entries to zero. This yields positive self-connection coefficients no greater than one, ensuring that the initial self-dynamics do not amplify the latent state. The recurrent interaction matrix, input weights and biases are initialized
to zero:
\begin{equation}
\mW=\mathbf{0},
\qquad
\mC=\mathbf{0}
\qquad 
\vh=\mathbf{0}.
\end{equation}

At the beginning of each task's training phase, we initialize the gate log-parameters to $\log\alpha_i=2$, corresponding to nearly open gates ($g_i\approx0.96$). Gates are applied from the first epoch. To reduce premature pruning during early
learning, we introduce the capacity penalty gradually by multiplying its coefficients by $\min(1,e/E_{\mathrm{warm}})$, where $e$ is the epoch index and $E_{\mathrm{warm}}=0.1*E_{\mathrm{total}}$. Thus, the capacity penalty is zero at $e=0$ and increases linearly to its target strength over the first $E_{\mathrm{warm}}$ epochs.

In contrast to \citet{louizos2018}, we use deterministic gates to avoid introducing additional gate-sampling noise into recurrent training. This choice sacrifices stochastic exploration mechanism for reactivating gates once they enter the zero-gradient region. Near-open initialization and the gradual introduction of the capacity penalty reduce the risk of premature closure but do not guarantee its avoidance.

\paragraph{Reinitialization during recycling.}
After a task is completed and before the next task begins,
units selected for release are returned to their respective
linear or ReLU free pools. Let $\mathcal{R}_k$ denote the set
of released units. For each $i\in\mathcal{R}_k$, we apply
\begin{equation}
a_i\sim\mathcal{U}(0.3,0.9),
\qquad
W_{i,:}\leftarrow\mathbf{0},
\qquad
W_{:,i}\leftarrow\mathbf{0},
\qquad
C_{i,:}\leftarrow\mathbf{0}
\qquad
h_i\leftarrow\mathbf{0}.
\end{equation}
Thus, both incoming and outgoing recurrent connections of a
recycled unit are cleared, together with its constant drive,
while its self-connection coefficient is assigned a newly sampled value. 

\subsection{Dynamical Systems}
\label{AP:Dynamical Systems}
\begin{figure}[htbp!]
\begin{center}
\includegraphics[width=\linewidth]{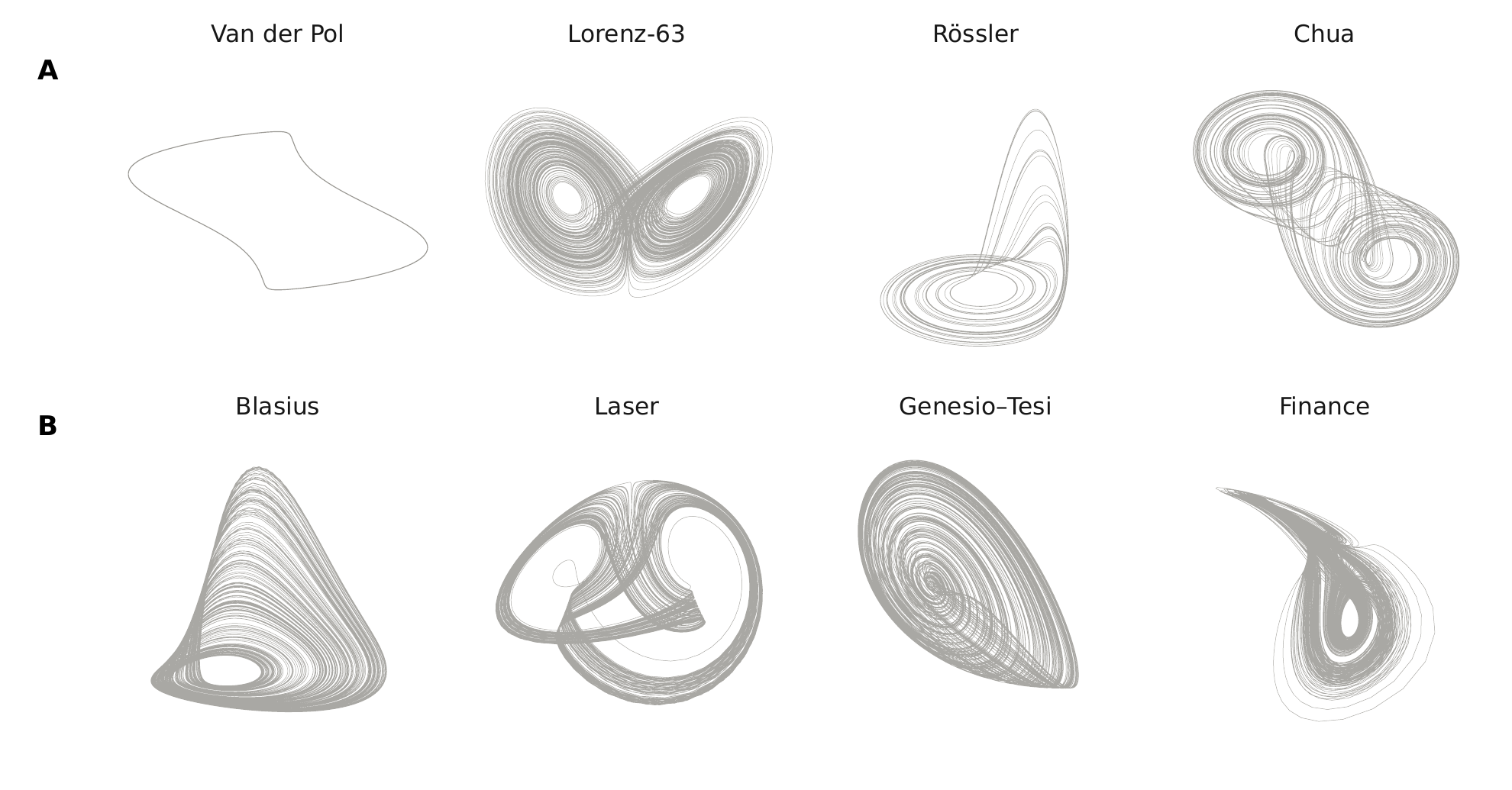}
\end{center}
\caption{\textbf{Dynamical systems used in the cDSR benchmarks.} All eight dynamical systems considered in this study are shown. \textbf{A)} First benchmark: Van der Pol $\rightarrow$ Lorenz-63 $\rightarrow$ R\"ossler $\rightarrow$ Chua. \textbf{B)} Second benchmark: Blasius $\rightarrow$ Laser $\rightarrow$ Genesio--Tesi $\rightarrow$ Finance.}
\label{fig:ground_truth_DS}
\end{figure}

We evaluated the model on two benchmarks, each comprising four nonlinear dynamical systems adapted from the \citet{Gilpin2023} database. The first benchmark comprises the Van der Pol oscillator, the Lorenz-63 system, the R\"ossler system, and Chua’s circuit (Fig.~\ref{fig:ground_truth_DS}A), spanning a periodic limit cycle and three chaotic attractors. The second comprises four chaotic systems: the Blasius food chain, the Laser system, the Genesio–Tesi system, and the Finance system (Fig.~\ref{fig:ground_truth_DS}B).

For the first benchmark, trajectories were generated by numerically
integrating the corresponding differential equations with the
fixed-step fourth-order Runge--Kutta method (RK4), using an integration
and observation interval of $\Delta t=0.01$. Initial states were
sampled as $(\epsilon_1,\epsilon_2)^\top$ for Van der Pol,
$(5\epsilon_1,5\epsilon_2,20+5|\epsilon_3|)^\top$ for Lorenz-63,
$2(\epsilon_1,\epsilon_2,\epsilon_3)^\top$ for R\"ossler, and
$0.1(\epsilon_1,\epsilon_2,\epsilon_3)^\top$ for Chua, where the
$\epsilon_i$ are independent standard Gaussian variables.

The systems of the second benchmark evolve on different time scales,
with dominant periods ranging from approximately $0.95$ (Laser)
to $9.9$ (Finance) time units. We therefore integrated each system
with RK4 using its integration step from the \citet{Gilpin2023} database metadata,
$\Delta t_{\mathrm{int}}\approx T_{\mathrm{dom}}/5000$, where
$T_{\mathrm{dom}}$ is the dominant period, and retained every 50th
state. This yields approximately 100 observations per dominant
period, with observation intervals of $0.066049$ (Blasius),
$0.0094683$ (Laser), $0.059441$ (Genesio--Tesi), and $0.098948$
(Finance). The corresponding integration steps are these intervals
divided by 50. For each seed, the reference initial conditions
from the \citet{Gilpin2023} were perturbed multiplicatively as
$x_{\mathrm{init},i}=x_{\mathrm{ref},i}(1+0.05\epsilon_i)$.
An initial transient of 10,000 sampled steps, corresponding to
approximately 100 dominant periods, was discarded to reduce
the influence of these perturbations.

In both benchmarks, a further burn-in window of 1,000 sampled
steps was excluded before retaining the training and test data. Each observed
dimension was standardized using the mean and standard deviation
of the main generated trajectory, including the 1,000-step
burn-in window and both the training and test portions, but
excluding the preliminary 10,000-step transient of the second
benchmark.

For each task, we generated 100,000 time steps for
training and a further 20,000 time steps for testing. The training
trajectories were divided into sequences of 200 time steps and presented
in batches of 16. Each task was trained for 2,000 epochs. The equations and
parameters of all eight systems are given below.

\paragraph{Van der Pol.} A second-order
nonlinear system exhibiting a stable limit cycle \citep{vanderPol1926}. It is governed by
\begin{align}
    \frac{dx}{dt} &= y, \\
    \frac{dy}{dt} &= \mu (1 - x^2) y - x,
\end{align}
where $\mu$ controls the strength of the nonlinear damping. Parameter
used: $\mu = 2.0$.

\paragraph{Lorenz-63.} A three-dimensional model
of atmospheric convection and a canonical example of deterministic chaos, exhibiting the well-known butterfly-shaped strange attractor for standard parameter settings ($\sigma = 10.0$, $\rho = 28$, $\beta = \frac{8}{3}$) \citep{Lorenz1963}. It is defined as
\begin{align}
    \frac{dx}{dt} &= \sigma (y - x), \\
    \frac{dy}{dt} &= x (\rho - z) - y, \\
    \frac{dz}{dt} &= xy - \beta z,
\end{align}
For the heterogeneous tasks, we used the standard parameterization described above. For the within-family Lorenz tasks, we varied the $\rho$ parameter across tasks, using $\rho \in \{28, 313, 200, 160\}$ while keeping the remaining parameters fixed.

\paragraph{R\"ossler.} A three-dimensional continuous
chaotic system designed to produce a simpler attractor topology than the Lorenz system while retaining chaotic behavior, characterized by a single spiral band with occasional folding \citep{ROSSLER1976}. It is given by
\begin{align}
    \frac{dx}{dt} &= -y - z, \\
    \frac{dy}{dt} &= x + ay, \\
    \frac{dz}{dt} &= b + z(x - c),
\end{align}
where $a$ and $b$ control the local spiraling dynamics and $c$ controls
the folding mechanism that produces chaos. Parameters used:
$a = 0.2$, $b = 0.2$, $c = 5.7$.

\paragraph{Chua.} A simple electronic circuit containing a nonlinear resistor, and is one of the most widely studied systems for generating a double-scroll chaotic attractor \citep{Chua1969}. It is defined as
\begin{align}
    \frac{dx}{dt} &= \alpha (y - x - f(x)), \\
    \frac{dy}{dt} &= x - y + z, \\
    \frac{dz}{dt} &= -\beta y,
\end{align}
with the piecewise-linear nonlinearity
\begin{align}
    f(x) = m_1 x + \tfrac{1}{2}(m_0 - m_1)\big(|x + 1| - |x - 1|\big),
\end{align}
where $\alpha$ and $\beta$ set the circuit's characteristic time scales
and $m_0$, $m_1$ define the slopes of the nonlinear resistor's
piecewise-linear characteristic. Parameters used: $\alpha = 15.6$,
$\beta = 28.0$, $m_0 = -\frac{8}{7}$, $m_1 = -\frac{5}{7}$.

\paragraph{Blasius.} A three-species food chain of vegetation $x$,
herbivores $y$ and predators $z$ that exhibits chaotic population
oscillations \citep{Blasius1999}. It is governed by
\begin{align}
    \frac{dx}{dt} &= a x - \frac{\alpha_1 x y}{1 + k_1 x}, \\
    \frac{dy}{dt} &= -b y + \frac{\alpha_1 x y}{1 + k_1 x} - \frac{\alpha_2 y z}{1 + k_2 y}, \\
    \frac{dz}{dt} &= -c (z - z^{*}) + \frac{\alpha_2 y z}{1 + k_2 y},
\end{align}
where $a$ is the vegetation growth rate, $b$ and $c$ are the mortality
rates of herbivores and predators, $\alpha_1$ and $\alpha_2$ are the
predation rates, $k_1$ and $k_2$ control the saturation of the functional
responses, and $z^{*}$ is a small predator immigration level. Parameters
used: $a = 1$, $b = 1$, $c = 10$, $\alpha_1 = 0.2$, $\alpha_2 = 1$,
$k_1 = 0.05$, $k_2 = 0$, $z^{*} = 0.006$.

\paragraph{Laser.} A three-dimensional chaotic system proposed as a model
of semiconductor laser dynamics \citep{ABOOEE2013}. 
It is governed by
\begin{align}
    \frac{dx}{dt} &= a (y - x) + b\, y z^2, \\
    \frac{dy}{dt} &= c\, x + d\, x z^2, \\
    \frac{dz}{dt} &= h\, z + k\, x^2,
\end{align}
where $a$ sets the linear coupling between $x$ and $y$, $b$ and $d$
control the cubic interaction with $z$, and $h$ and $k$ set the linear
damping of $z$ and its quadratic driving by $x$. Parameters used:
$a = 10$, $b = 1$, $c = 5$, $d = -1$, $h = -5$, $k = -6$.

\paragraph{Genesio--Tesi.} A third-order nonlinear feedback system with a
single quadratic nonlinearity that exhibits a chaotic attractor \citep{Genesio1992}. It is governed by
\begin{align}
    \frac{dx}{dt} &= y, \\
    \frac{dy}{dt} &= z, \\
    \frac{dz}{dt} &= -c x - b y - a z + x^2,
\end{align}
where $a$, $b$ and $c$ are coefficients of the linear feedback terms. Parameters used: $a = 0.44$, $b = 1.1$, $c = 1$.

\paragraph{Finance.} A macroeconomic model of the interplay between the
interest rate $x$, the investment demand $y$ and the price index $z$ that
exhibits chaotic fluctuations \citep{Cai2007}. It is governed by
\begin{align}
    \frac{dx}{dt} &= \left(\tfrac{1}{b} - a\right) x + z + x y, \\
    \frac{dy}{dt} &= -b y - x^2, \\
    \frac{dz}{dt} &= -x - c z,
\end{align}
where $a$ is the saving amount, $b$ the cost per unit of investment and
$c$ the elasticity of demand of commercial markets. Parameters used: $a = 0.001$, $b = 0.2$, $c = 1.1$.

\subsection{Evaluation Measures}
\label{AP:Evaluation_Meausres}
\paragraph{State-space divergence ($D_{\mathrm{stsp}}$).}
To quantify how well a reconstructed system reproduces the geometry of the ground-truth attractor, we use the state-space divergence $D_{\mathrm{stsp}}$ introduced by \citet{Koppe2019}. This measure compares the state-space distributions of trajectories generated by the ground-truth system, $p_{\mathrm{true}}(\vx)$, and by the reconstructed model, $p_{\mathrm{gen}}(\vx\mid\vz)$. For low-dimensional systems, these distributions are estimated by discretizing the state space into bins, yielding

\begin{equation}
D_{\mathrm{stsp}}
\left(
p_{\mathrm{true}}(\vx),
p_{\mathrm{gen}}(\vx \mid\mathbf{z})
\right)
\approx
\sum_{i=1}^{b^N}
\hat{p}^{(i)}_{\mathrm{true}}
\log
\left(
\frac{\hat{p}^{(i)}_{\mathrm{true}}}
{\hat{p}^{(i)}_{\mathrm{gen}}}
\right),
\end{equation}

where $\hat{p}^{(i)}_{\mathrm{true}}$ and $\hat{p}^{(i)}_{\mathrm{gen}}$ denote the relative frequencies with which the ground-truth and generated trajectories occupy bin $i$, respectively, $b$ denotes the number of total bins defined per dimension, and $N$ denotes the dimensionality of the dynamical system. If no generated samples fall within the occupied ground-truth bins, the measure is considered divergent. Before estimating these distributions, initial transients are discarded so that the trajectories represent the long-term behavior of the freely running systems.

For the final evaluation on synthetic dynamical systems, five initial observations are selected at evenly spaced indices along the test trajectory. From each initial condition, the model is rolled out autonomously for 40,000 steps, with the first 10,000 steps discarded as transients. The remaining 30,000 generated samples are compared with the full 20,000-sample reference test trajectory. The resulting scores are aggregated across the five initial conditions using the median, with divergent cases excluded.

Following \citet{Koppe2019}, the state space is partitioned over a range determined by the variability of the data. The bin resolution involves a trade-off: overly coarse bins may obscure relevant geometrical structure, whereas excessively fine bins lead to sparsely populated state spaces and unreliable distribution estimates. For the synthetic dynamical systems, we use $b=30$ bins per dimension, resulting in $30^2$ or $30^3$ total bins for two- or three-dimensional systems, respectively, providing sufficient resolution while avoiding excessive sparsity.


Direct binning becomes impractical as dimensionality increases
because the number of bins grows exponentially with $N$.
For the six-dimensional gait representation used here,
$b=30$ bins per dimension would require
$30^6\approx7.3\times10^8$ bins. We therefore estimate
state-space divergence using Gaussian mixture models (GMMs),
following \citet{Koppe2019,Brenner2022}. Each trajectory defines an
equally weighted mixture with one Gaussian component per
retained sample:
\begin{align}
\hat p_{\mathrm{true}}(\vx)
&= \frac{1}{T'}
\sum_{t=1}^{T'}\mathcal{N}(\vx;\vx_t,\Sigma),\\
\hat p_{\mathrm{gen}}(\vx)
&= \frac{1}{L'}
\sum_{l=1}^{L'}\mathcal{N}(\vx;\hat{\vx}_l,\Sigma),
\end{align}
where $\{\vx_t\}_{t=1}^{T'}$ and
$\{\hat{\vx}_l\}_{l=1}^{L'}$ are the reference and generated
trajectory samples, respectively. We use a shared isotropic
covariance $\Sigma=\sigma^2\mI$, with $\sigma^2=0.01$
in standardized observation coordinates. This variance
determines the spatial resolution of the density estimate
and is held fixed across evaluations.

We estimate
$\mathrm{KL}(\hat p_{\mathrm{true}}\Vert\hat p_{\mathrm{gen}})$ 
by Monte Carlo sampling:
\begin{equation}
\widetilde D_{\mathrm{stsp}}
=
\frac{1}{n}
\sum_{i=1}^{n}
\log
\frac{
\frac{1}{T'}\sum_{t=1}^{T'}
\mathcal{N}(\vx^{(i)};\vx_t,\Sigma)
}{
\frac{1}{L'}\sum_{l=1}^{L'}
\mathcal{N}(\vx^{(i)};\hat{\vx}_l,\Sigma)
},
\qquad
\vx^{(i)}\sim\hat p_{\mathrm{true}}.
\end{equation}
We draw $n=2000$ Monte Carlo samples by selecting reference
mixture components uniformly and adding Gaussian noise with
covariance $\Sigma$.

For the gait dataset we use the $\widetilde D_{\mathrm{stsp}}$ as the evaluation measure. Each autonomous rollout contains
12,500 transitions. After discarding the first 5,000 as a
transient, all $L'=7500$ generated samples are compared with
the $T'=5000$ reference test samples.
Smaller divergence values indicate closer agreement between
the estimated state-space distributions.

\paragraph{Hellinger distance.}
Following \citet{Mikhaeil2022}, we assess temporal agreement between ground-truth and generated trajectories by computing the Hellinger distance between their power spectra. The Hellinger distance is bounded by $0\leq D_H\leq1$, with $D_H=0$ indicating identical spectral distributions. We compute $D_H$ separately for each observed dynamical variable and report their average as $D_H$, providing a single measure of temporal reconstruction quality.

\subsection{Transfer Mechanism Hyperparameter Search}
\label{AP:Transfer mechanism}

To enable forward transfer, we allow units in the free pool to receive inputs from committed units, while preventing free units from influencing committed units. Specifically, the transfer weights from committed source units to free target units, denoted by \(W_{\mathrm{com}\rightarrow\mathrm{free}}\), remain plastic. Conversely, the reverse connections, \(W_{\mathrm{free}\rightarrow\mathrm{com}}\), are fixed to zero, ensuring that learning a new task cannot modify representations allocated to previous tasks.

Naively optimizing the transfer weights using gradient descent may cause the model to rely excessively on committed units. This issue becomes more pronounced as the committed pool grows, since the number of potential transfer connections can eventually exceed the number of connections among free units. We therefore regularize the transfer pathways using either \(L_0\) gating or \(L_1\)/\(L_2\) penalties.

Let
$\mathcal{T}_{ij}:=g_ig_jW_{ij}$
denote the transfer connection from committed unit \(j\) to free unit \(i\), after applying the target unit's own gate. This ordering is important: if a unit is about to be released (\(g_i < \tau_g\)), its incoming transfer connections are already scaled close to zero by its own gate. Consequently, the regularizer exerts essentially no pressure on such connections and instead concentrates on connections to units that will remain active.

We compare four ways of treating these connections: no regularization, \(L_0\) gating of committed source units, and \(L_1\) or \(L_2\) penalties on the effective transfer weights. The corresponding transfer losses are
\begin{align}
\mathcal{L}_{\mathrm{transfer}}^{k} =
\begin{cases}
0, & \text{unregularized}, \\[2pt]
\lambda_{\mathrm{tr}}
\displaystyle\sum_{j \in \sU^{\text{committed}}_{k-1}}
\sigma\!\left(
\log \alpha^{\mathrm{lat}}_{j} -
\log\frac{-\gamma}{\zeta}
\right), & L_0, \\[4pt]
\lambda_{\mathrm{tr}}
\displaystyle\sum_{i\in\sU^{\text{free}}_{k},\, j\in\sU^{\text{committed}}_{k-1}}
\lvert \mathcal{T}_{ij} \rvert,
& L_1, \\[2pt]
\lambda_{\mathrm{tr}}
\displaystyle\sum_{i\in\sU^{\text{free}}_{k},\, j\in\sU^{\text{committed}}_{k-1}}
\mathcal{T}_{ij}^{2},
& L_2.
\end{cases}
\label{eq:transfer_loss}
\end{align}
Here, \(\lambda_{\mathrm{tr}}\) controls the strength of transfer regularization. The \(L_0\) formulation differs structurally from the \(L_1\) and \(L_2\) alternatives: it introduces a second hard-concrete gate, \(g^{\mathrm{lat}}_j\), for each committed source unit \(j\). The lateral gate for each committed source unit $c$ is defined as:
\begin{align}
g_j^{\mathrm{lat}}
=
\operatorname{clip}_{[0,1]}
\left[
\sigma(\log\alpha_j^{\mathrm{lat}})(\zeta-\gamma)+\gamma
\right],
\qquad j\in\sU_{k-1}^{\mathrm{committed}},
\label{eq:lateral_gate}
\end{align}
where $\gamma=-0.1$ and $\zeta=1.1$. At the beginning of each task, we initialize
$\log\alpha_j^{\mathrm{lat}}=2$, giving
$g_j^{\mathrm{lat}}\approx0.957$. Therefore the transfer from the commited unit c to free unit i is:
\begin{align}
\mathcal T_{ij}:=g_ig_j^{\mathrm{lat}}W_{ij},
\end{align}
At task completion, lateral gates are thresholded using
$\tau_g=0.5$:
\begin{align}
\bar g_j^{\mathrm{lat}}
=
g_j^{\mathrm{lat}}
\mathbf{1}\!\left[g_j^{\mathrm{lat}}>\tau_g\right].
\label{eq:lateral_threshold}
\end{align}
Retained gates preserve their learned magnitudes rather than being
set to one. Together with the thresholded target-unit gates
$\bar g_i$, they are absorbed into the transfer weights:
\begin{align}
W_{ij}\leftarrow
\bar g_i\,\bar g_j^{\mathrm{lat}}W_{ij},
\qquad
i\in\mathcal \sU_k^{\mathrm{free}},\quad
j\in\sU_{k-1}^{\mathrm{committed}},
\label{eq:lateral_commit}
\end{align}
where $\bar g_i=g_i\mathbf{1}[g_i>\tau_g]$ for gated target units
and $\bar g_i=1$ for protected readout units.
Rows belonging to retained target units are subsequently frozen.
Closing a lateral gate removes the current task's access to that
source; it does not release the previously committed source unit.
Fresh lateral gate parameters are initialized for each new task.
For the first task, the committed pool is empty, so the transfer
contribution and transfer loss are both zero.
In contrast, \(L_1\) and \(L_2\) regularization shrink all transfer weights, which can limit excessive dependence on prior-task representations but may also attenuate useful transfer connections.

The Pareto fronts in Fig.~\ref{fig:DSR_task_pareto_fronts} show the Optuna trials for each transfer mechanism over the hyperparameter ranges reported in Table~\ref{tab:transfer-hyperparameters}, evaluated on both the heterogeneous-task and Lorenz-family benchmarks.

\begin{table}[htbp!]
\caption{Hyperparameter search ranges for the transfer mechanisms.}
\label{tab:transfer-hyperparameters}
\begin{center}
\begin{tabular}{cc}
\multicolumn{1}{c}{\bf Transfer Mechanism} & \multicolumn{1}{c}{\bf $\lambda_{\text{tr}}$} 
\\ \hline \\
Naive  & --  \\
$L_0$ & $[10^{-6},10^{-3}]$ \\
$L_1$ & $[10^{-6},10^{-2}]$ \\
$L_2$ & $[3\times10^{-5},3\times10^{-1}]$  \\
\end{tabular}
\end{center}
\end{table}

\subsection{CRUG on Synthetic Dynamical Systems}
\label{AP:CRUG on Synthetic Dynamical Systems}


To demonstrate that the performance of CRUG on continual learning of dynamical systems is not restricted to the specific systems tested above (Van der Pol, Lorenz-63, R\"ossler, and Chua), we conducted an additional experiment using four other chaotic systems from the database of \citet{Gilpin2023}: Blasius, Laser, Genesio--Tesi, and Finance, as described in Appendix~\ref{AP:Dynamical Systems}. The hyperparameter search ranges for CRUG with $L_2$ transfer regularization on both task sequences are reported in Table~\ref{tab:CRUG_L2_DSR}. As before, we implemented all methods on an AL-RNN with \(M=160\) and \(P=80\), trained sequentially on the four tasks. We used Optuna with 20 trials and 5 seeds per trial to fine-tune each method's
hyperparameters. The best trial for each method is reported in Table~\ref{tab:DSR_additional_tasks}.

In agreement with our previous findings on continual DSR of another set of dynamical systems, the parameter-regularization-based methods (EWC and SI) and the parameter-isolation method XdG, are unable to prevent forgetting. The replay-based methods achieve better performance and substantially reduce forgetting. Although CLNP does not exhibit forgetting, it fails to learn the tasks accurately in the first place ($D_{\mathrm{stsp}} = 1.1$–$3.2$ immediately after training). In contrast, CRUG learns all four tasks to $D_{\mathrm{stsp}} < 1.0$ with zero forgetting. CRUG thereby improves on CLNP by roughly an order of magnitude in overall $D_{\mathrm{stsp}}$ at comparable capacity ($98$ vs.\ $103$ committed units, $p=1.00$). In summary, on this benchmark CRUG is competitive with the replay methods on reconstruction while committing only ${\sim}60\%$ of the network's units and, unlike them, requiring no stored or generated replay data.

\begin{figure}[htbp!]
    \centering
    \includegraphics[width=0.95\linewidth]{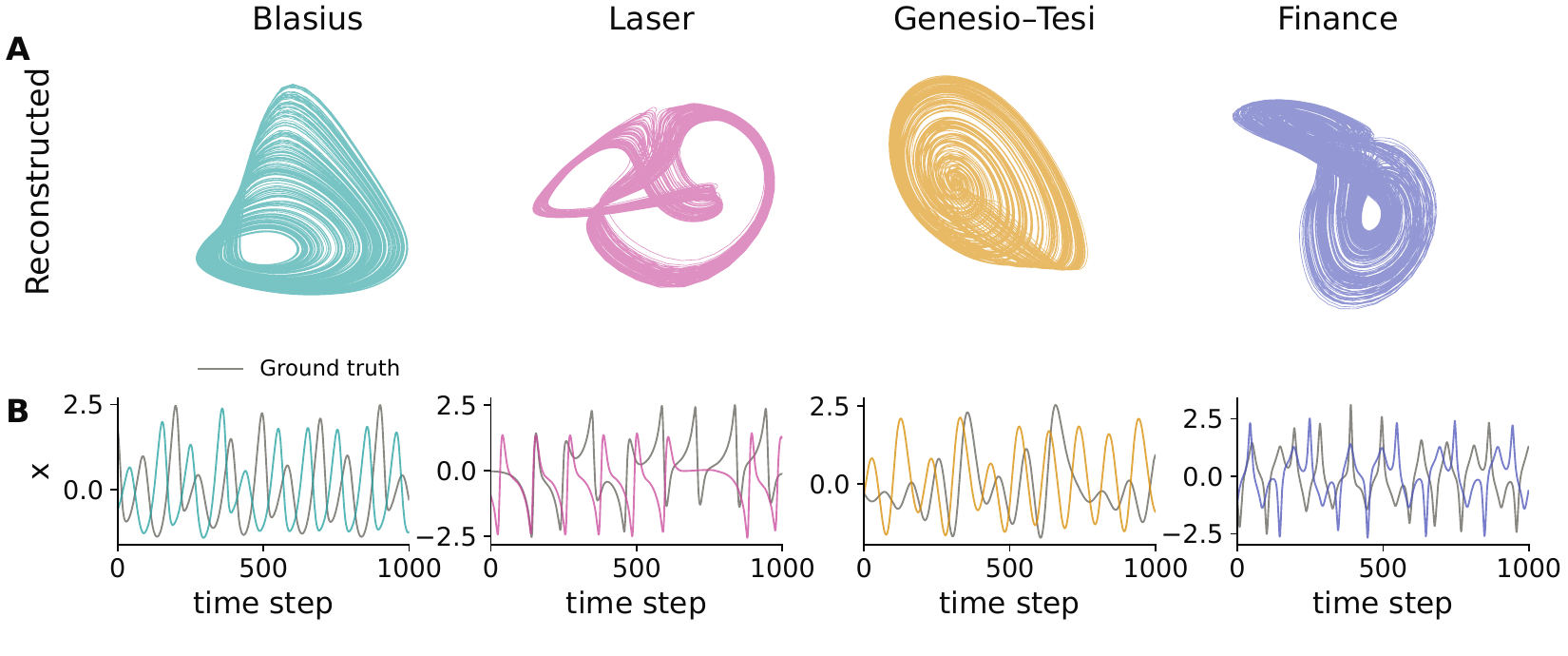}
    \caption{\textbf{cDSR on synthetic DS sequence 2.} Four additional chaotic dynamical systems were trained sequentially. 
    \textbf{A)} The final model was used to generate rollout trajectories for all four tasks. 
    \textbf{B)} Reconstructed and ground-truth $x$-dimensions are shown over 1000 time steps.}
    \label{fig:BLGF_DSR_tasks}
\end{figure}

The reconstructed trajectories produced by the model equipped with CRUG after learning all four tasks are compared with the corresponding ground-truth trajectories in Fig.~\ref{fig:BLGF_DSR_tasks}.

\begin{table}[htbp!]
\caption{\textbf{Benchmarking CRUG against other continual learning methods
on four additional DSR tasks} (Blasius $\rightarrow$
Laser $\rightarrow$ Genesio--Tesi $\rightarrow$ Finance).
Each entry in the four first columns reports the $D_{\mathrm{stsp}}$ measure immediately after the task’s
own training phase $\rightarrow$ after all tasks have been learned
sequentially (except for interleaved training). Values are reported in the format
$\text{median}^{\,Q_3-\text{median}}_{\,Q_1-\text{median}}$ across 5 runs
(lower is better). Divergent or incomplete runs are not omitted but counted as worst results in the median.
Overall performance is the maximum final task $D_{\text{stsp}}$ and  $D_{H}$ within each
run, summarized across all runs.
``div.'' denotes an infinite median, and ${}^\dagger$ denotes
an infinite upper quartile. Finance is learned last and therefore has no subsequent
forgetting period.}

\label{tab:DSR_additional_tasks}
\centering
\renewcommand{\arraystretch}{1.5}
\resizebox{\textwidth}{!}{%
\begin{tabular}{l|cccc|cc}

\hline

& \multicolumn{4}{c|}{\textbf{$D_{\mathrm{stsp}}$}}

& \multicolumn{2}{c}{\textbf{Overall performance}} \\

\cline{2-5}\cline{6-7}

\textbf{Method}

& \textbf{Blasius}

& \textbf{Laser}

& \textbf{Genesio--Tesi}

& \textbf{Finance}

& \textbf{$D_{\mathrm{stsp}}\downarrow$}

& \textbf{$D_H\downarrow$} \\

\hline
& & & & & & \\
Interleaved & $0.29^{+0.02}_{-0.10}\!\rightarrow\!0.29^{+0.02}_{-0.10}$ & $0.41^{+0.17}_{-0.11}\!\rightarrow\!0.41^{+0.17}_{-0.11}$ & $0.57^{+0.16}_{-0.20}\!\rightarrow\!0.57^{+0.16}_{-0.20}$ & $0.31^{+0.09}_{-0.00}$ & $0.73^{+0.14}_{-0.26}$ & $0.06^{+0.00}_{-0.01}$\\
Naive & $0.16^{+0.01}_{-0.01}\!\rightarrow\!\text{div.}$ & $0.20^{+0.02}_{-0.02}\!\rightarrow\!15.28^{\dagger}_{-0.06}$ & $0.29^{+0.07}_{-0.03}\!\rightarrow\!14.76^{\dagger}_{-0.25}$ & $0.24^{+0.00}_{-0.01}$ & $\text{div.}$ & $\text{div.}$\\
EWC & $0.15^{+0.02}_{-0.01}\!\rightarrow\!\text{div.}$ & $0.18^{+0.02}_{-0.05}\!\rightarrow\!15.23^{\dagger}_{-0.01}$ & $0.41^{+0.10}_{-0.12}\!\rightarrow\!14.86^{\dagger}_{-0.36}$ & $0.31^{+0.03}_{-0.03}$ & $\text{div.}$ & $\text{div.}$\\
SI & $0.17^{+0.02}_{-0.03}\!\rightarrow\!0.16^{+0.01}_{-0.00}$ & $15.25^{+0.02}_{-0.04}\!\rightarrow\!15.25^{+0.02}_{-0.04}$ & $14.50^{+0.35}_{-0.00}\!\rightarrow\!14.50^{+0.35}_{-0.00}$ & $15.01^{+0.13}_{-0.03}$ & $15.25^{+0.02}_{-0.04}$ & $\text{div.}$\\
ER & $0.14^{+0.02}_{-0.02}\!\rightarrow\!0.51^{+0.07}_{-0.34}$ & $0.18^{+0.06}_{-0.00}\!\rightarrow\!0.47^{+0.16}_{-0.06}$ & $0.38^{+0.04}_{-0.01}\!\rightarrow\!0.95^{+0.56}_{-0.63}$ & $0.31^{+0.01}_{-0.08}$ & $1.51^{+7.68}_{-0.87}$ & $0.09^{+0.11}_{-0.01}$\\
GR & $0.16^{+0.00}_{-0.02}\!\rightarrow\!0.19^{+0.02}_{-0.01}$ & $0.27^{+0.06}_{-0.11}\!\rightarrow\!0.43^{+0.15}_{-0.10}$ & $0.40^{+0.01}_{-0.09}\!\rightarrow\!0.58^{+0.03}_{-0.00}$ & $0.24^{+0.03}_{-0.01}$ & $\mathbf{0.61^{+0.05}_{-0.03}}$ & $\mathbf{0.05^{+0.01}_{-0.00}}$\\
XdG & $5.62^{+0.45}_{-3.27}\!\rightarrow\!5.62^{\dagger}_{-3.32}$ & $14.13^{+0.17}_{-1.38}\!\rightarrow\!14.13^{\dagger}_{-1.38}$ & $12.96^{+0.87}_{-11.55}\!\rightarrow\!12.96^{+0.87}_{-11.55}$ & $14.99^{+0.16}_{-2.16}$ & $15.24^{\dagger}_{-0.38}$ & $\text{div.}$\\
CLNP & $3.17^{+0.47}_{-0.46}\!\rightarrow\!3.17^{+0.47}_{-0.46}$ & $2.51^{+0.00}_{-0.65}\!\rightarrow\!2.51^{+0.00}_{-0.65}$ & $3.23^{+0.40}_{-1.21}\!\rightarrow\!3.23^{+0.40}_{-1.21}$ & $1.17^{+0.05}_{-0.37}$ & $4.72^{+0.38}_{-1.09}$ & $0.15^{+0.01}_{-0.01}$\\
\hline
CRUG & $0.77^{+0.21}_{-0.21}\!\rightarrow\!0.74^{+0.24}_{-0.18}$ & $0.64^{+0.02}_{-0.16}\!\rightarrow\!0.64^{+0.02}_{-0.11}$ & $0.64^{+0.07}_{-0.22}\!\rightarrow\!0.64^{+0.13}_{-0.22}$ & $0.62^{+0.36}_{-0.03}$ & $0.77^{+0.41}_{-0.13}$ & $0.07^{+0.02}_{-0.01}$\\
\hline
\end{tabular}}
\end{table}

\begin{table}[htbp!]
\centering
\renewcommand{\arraystretch}{1.3}
\footnotesize
\caption{\textbf{CRUG with $L_2$ transfer regularization: ten best configurations on the canonical and the additional DSR benchmark.} Top: Van der Pol $\to$ Lorenz-63 $\to$ R\"ossler $\to$ Chua, 40 Optuna trials, seeds 0--9. Bottom: Blasius $\to$ Laser $\to$ Genesio--Tesi $\to$ Finance, 20 Optuna trials, 10 seeds. Configurations are ranked by the median across seeds of the maximum final $D_{\mathrm{stsp}}$ over the four tasks. Task entries show own-training $\to$ final $D_{\mathrm{stsp}}$; the last task has only its own-training evaluation. Entries show $\text{median}^{\,Q_3-\text{median}}_{\,Q_1-\text{median}}$ across all seeds. ``div.'' denotes an infinite median. Overall $D_H$ is the maximum final $D_H$ over the four tasks within each run. Units kept is the total number of committed units across the four tasks (of $M=160$). Bold rows mark the configurations reported in the method comparison Tables~\ref{tab:DSR_canonical_tasks} and ~\ref{tab:DSR_additional_tasks}.}
\label{tab:CRUG_L2_DSR}
\resizebox{\textwidth}{!}{%
\begin{tabular}{c|llll|lll}
\hline
& \multicolumn{4}{c|}{\textbf{$D_{\mathrm{stsp}}$}}

& \multicolumn{3}{c}{\textbf{Overall performance}} \\

\cline{2-5}\cline{6-8}

{\bfseries\boldmath $(\lambda_{\mathrm{ReLU}},\lambda_{\mathrm{lin}},\lambda_{\mathrm{tr}})$}

& \textbf{Van der Pol}

& \textbf{Lorenz-63}

& \textbf{R\"ossler}

& \textbf{Chua}

& \textbf{$D_{\mathrm{stsp}}\downarrow$}

& \textbf{$D_H\downarrow$} 
& \textbf{Units kept}
\\
\hline
&&&&&&& \\
$(6.58\times10^{-4}, 2.43\times10^{-4}, 4.94\times10^{-2})$ & $0.06^{+0.00}_{-0.01}\to 0.06^{+0.00}_{-0.01}$ & $0.23^{+0.13}_{-0.03}\to 0.28^{+0.12}_{-0.08}$ & $0.91^{+0.72}_{-0.17}\to 0.83^{+0.95}_{-0.16}$ & $0.75^{+0.08}_{-0.09}$ & $0.92^{+0.85}_{-0.12}$ & $0.05^{+0.01}_{-0.00}$ & $116^{+13}_{-8}$ \\
$(7.66\times10^{-4}, 2.52\times10^{-4}, 8.47\times10^{-2})$ & $0.08^{+0.03}_{-0.02}\to 0.08^{+0.03}_{-0.02}$ & $0.29^{+0.09}_{-0.08}\to 0.34^{+0.13}_{-0.13}$ & $0.86^{+0.29}_{-0.13}\to 0.89^{+0.14}_{-0.15}$ & $0.79^{+0.09}_{-0.22}$ & $0.92^{+0.24}_{-0.08}$ & $0.06^{+0.01}_{-0.01}$ & $112^{+3}_{-4}$ \\
$(5.69\times10^{-4}, 1.88\times10^{-4}, 7.50\times10^{-2})$ & $0.07^{+0.04}_{-0.01}\to 0.07^{+0.04}_{-0.01}$ & $0.26^{+0.16}_{-0.07}\to 0.34^{+0.12}_{-0.12}$ & $0.77^{+0.58}_{-0.07}\to 0.77^{+0.16}_{-0.14}$ & $0.64^{+0.19}_{-0.05}$ & $0.93^{+0.21}_{-0.12}$ & $0.06^{+0.01}_{-0.01}$ & $124^{+3}_{-5}$ \\
$(5.87\times10^{-4}, 2.02\times10^{-4}, 2.40\times10^{-1})$ & $0.07^{+0.01}_{-0.01}\to 0.07^{+0.01}_{-0.01}$ & $0.27^{+0.14}_{-0.05}\to 0.24^{+0.16}_{-0.02}$ & $1.04^{+0.05}_{-0.14}\to 0.86^{+0.31}_{-0.21}$ & $0.73^{+0.02}_{-0.05}$ & $0.95^{+0.26}_{-0.06}$ & $0.05^{+0.02}_{-0.01}$ & $120^{+4}_{-2}$ \\
$(5.56\times10^{-4}, 2.34\times10^{-4}, 2.36\times10^{-1})$ & $0.06^{+0.02}_{-0.01}\to 0.06^{+0.02}_{-0.01}$ & $0.21^{+0.21}_{-0.01}\to 0.21^{+0.18}_{-0.02}$ & $0.93^{+0.98}_{-0.11}\to 0.90^{+0.74}_{-0.16}$ & $0.81^{+0.03}_{-0.14}$ & $0.96^{+0.68}_{-0.12}$ & $0.05^{+0.00}_{-0.00}$ & $116^{+6}_{-1}$ \\
$(5.20\times10^{-4}, 1.78\times10^{-4}, 1.24\times10^{-1})$ & $0.06^{+0.01}_{-0.02}\to 0.06^{+0.01}_{-0.02}$ & $0.25^{+0.10}_{-0.04}\to 0.28^{+0.12}_{-0.07}$ & $0.86^{+0.25}_{-0.08}\to 0.89^{+0.16}_{-0.11}$ & $0.72^{+0.05}_{-0.10}$ & $0.97^{+0.10}_{-0.17}$ & $0.05^{+0.02}_{-0.01}$ & $126^{+5}_{-4}$ \\
$(5.09\times10^{-4}, 1.77\times10^{-4}, 9.30\times10^{-3})$ & $0.08^{+0.06}_{-0.01}\to 0.08^{+0.06}_{-0.01}$ & $0.24^{+0.26}_{-0.05}\to 0.28^{+0.30}_{-0.08}$ & $1.20^{+0.22}_{-0.28}\to 0.97^{+0.47}_{-0.21}$ & $0.86^{+0.06}_{-0.23}$ & $0.97^{+0.47}_{-0.07}$ & $0.06^{+0.01}_{-0.01}$ & $128^{+8}_{-7}$ \\
$(8.75\times10^{-4}, 3.00\times10^{-4}, 9.86\times10^{-3})$ & $0.07^{+0.02}_{-0.00}\to 0.07^{+0.02}_{-0.00}$ & $0.40^{+0.13}_{-0.12}\to 0.38^{+0.15}_{-0.12}$ & $1.00^{+0.11}_{-0.12}\to 0.98^{+0.04}_{-0.07}$ & $0.86^{+0.09}_{-0.20}$ & $0.98^{+0.23}_{-0.07}$ & $0.07^{+0.02}_{-0.01}$ & $105^{+17}_{-4}$ \\
$(1.20\times10^{-3}, 3.90\times10^{-4}, 6.75\times10^{-2})$ & $0.09^{+0.02}_{-0.01}\to 0.09^{+0.02}_{-0.01}$ & $0.38^{+0.15}_{-0.13}\to 0.35^{+0.18}_{-0.10}$ & $0.93^{+0.14}_{-0.18}\to 1.02^{+0.24}_{-0.20}$ & $0.79^{+0.07}_{-0.13}$ & $1.02^{+0.24}_{-0.20}$ & $0.07^{+0.01}_{-0.01}$ & $100^{+6}_{-5}$ \\
$(5.64\times10^{-4}, 1.92\times10^{-4}, 1.18\times10^{-1})$ & $0.06^{+0.01}_{-0.01}\to 0.06^{+0.01}_{-0.01}$ & $0.30^{+0.20}_{-0.07}\to 0.43^{+0.22}_{-0.20}$ & $0.94^{+0.11}_{-0.13}\to 0.91^{+0.15}_{-0.13}$ & $0.66^{+0.13}_{-0.09}$ & $1.03^{+0.16}_{-0.22}$ & $0.05^{+0.03}_{-0.01}$ & $125^{+3}_{-5}$ \\
{\boldmath$(2.33\times10^{-3}, 1.53\times10^{-3}, 1.40\times10^{-2})$} & {\boldmath$0.15^{+0.16}_{-0.06}\to 0.15^{+0.16}_{-0.06}$} & {\boldmath$0.40^{+0.13}_{-0.12}\to 0.41^{+0.06}_{-0.12}$} & {\boldmath$1.22^{+1.02}_{-0.21}\to 1.26^{+1.16}_{-0.20}$} & {\boldmath$0.98^{+0.12}_{-0.17}$} & {\boldmath$1.26^{+1.16}_{-0.18}$} & {\boldmath$0.08^{+0.01}_{-0.01}$} & {\boldmath$63^{+3}_{-1}$} \\

\end{tabular}%
}
\vspace{1em}

\resizebox{\textwidth}{!}{%
\begin{tabular}{c|llll|lll}
\hline
& \multicolumn{4}{c|}{\textbf{$D_{\mathrm{stsp}}$}}

& \multicolumn{3}{c}{\textbf{Overall performance}} \\

\cline{2-5}\cline{6-8}

{\bfseries\boldmath $(\lambda_{\mathrm{ReLU}},\lambda_{\mathrm{lin}},\lambda_{\mathrm{tr}})$}

& \textbf{Blasius}

& \textbf{Laser}

& \textbf{Genesio--Tesi}

& \textbf{Finance}

& \textbf{$D_{\mathrm{stsp}}\downarrow$}

& \textbf{$D_H\downarrow$} 
& \textbf{Units kept}
\\
\hline
&&&&&&& \\
$(7.51\times10^{-4}, 7.05\times10^{-4}, 3.07\times10^{-2})$ & $0.44^{+0.48}_{-0.02}\to 0.41^{+0.50}_{-0.05}$ & $0.36^{+0.11}_{-0.05}\to 0.37^{+0.07}_{-0.01}$ & $0.34^{+0.23}_{-0.01}\to 0.39^{+0.03}_{-0.07}$ & $0.30^{+0.04}_{-0.00}$ & $0.45^{+0.46}_{-0.01}$ & $0.07^{+0.02}_{-0.00}$ & $130^{+0}_{-7}$ \\
$(6.83\times10^{-4}, 3.79\times10^{-4}, 3.40\times10^{-5})$ & $0.31^{+0.01}_{-0.02}\to 0.31^{+0.01}_{-0.00}$ & $0.33^{+0.04}_{-0.04}\to 0.30^{+0.00}_{-0.01}$ & $0.32^{+0.19}_{-0.04}\to 0.36^{+0.21}_{-0.06}$ & $0.36^{+0.00}_{-0.03}$ & $0.47^{+0.10}_{-0.02}$ & $0.07^{+0.00}_{-0.01}$ & $149^{+2}_{-17}$ \\
$(5.44\times10^{-4}, 3.30\times10^{-4}, 3.70\times10^{-5})$ & $0.36^{+0.13}_{-0.09}\to 0.36^{+0.13}_{-0.09}$ & $0.33^{+0.06}_{-0.02}\to 0.30^{+0.09}_{-0.01}$ & $0.36^{+0.01}_{-0.06}\to 0.36^{+0.02}_{-0.07}$ & $0.38^{+0.18}_{-0.11}$ & $0.52^{+0.32}_{-0.03}$ & $0.06^{+0.03}_{-0.01}$ & $150^{+2}_{-2}$ \\
$(8.19\times10^{-4}, 8.05\times10^{-4}, 1.23\times10^{-3})$ & $0.31^{+0.02}_{-0.04}\to 0.27^{+0.06}_{-0.00}$ & $0.38^{+0.02}_{-0.01}\to 0.37^{+0.01}_{-0.01}$ & $0.35^{+0.01}_{-0.02}\to 0.36^{+0.01}_{-0.05}$ & $0.34^{+0.06}_{-0.02}$ & $0.53^{+0.14}_{-0.12}$ & $0.06^{+0.02}_{-0.00}$ & $128^{+5}_{-2}$ \\
$(5.07\times10^{-4}, 2.97\times10^{-4}, 1.64\times10^{-2})$ & $0.40^{+0.03}_{-0.01}\to 0.42^{+0.03}_{-0.02}$ & $0.30^{+0.10}_{-0.01}\to 0.32^{+0.08}_{-0.02}$ & $0.40^{+0.02}_{-0.08}\to 0.32^{+0.11}_{-0.02}$ & $0.29^{+0.03}_{-0.01}$ & $0.55^{+0.01}_{-0.12}$ & $0.08^{+0.01}_{-0.01}$ & $150^{+2}_{-3}$ \\
$(6.00\times10^{-4}, 3.99\times10^{-4}, 5.44\times10^{-5})$ & $0.49^{+0.01}_{-0.26}\to 0.49^{+0.01}_{-0.26}$ & $0.33^{+0.07}_{-0.04}\to 0.33^{+0.07}_{-0.04}$ & $0.37^{+0.18}_{-0.01}\to 0.37^{+0.18}_{-0.01}$ & $0.37^{+0.05}_{-0.02}$ & $0.56^{+0.07}_{-0.15}$ & $0.10^{+0.01}_{-0.01}$ & $142^{+7}_{-2}$ \\
$(7.42\times10^{-4}, 2.34\times10^{-4}, 1.29\times10^{-2})$ & $0.36^{+0.56}_{-0.09}\to 0.35^{+0.57}_{-0.10}$ & $0.29^{+0.11}_{-0.00}\to 0.30^{+0.11}_{-0.02}$ & $0.40^{+0.14}_{-0.07}\to 0.41^{+0.16}_{-0.06}$ & $0.36^{+0.02}_{-0.00}$ & $0.57^{+0.36}_{-0.16}$ & $0.07^{+0.02}_{-0.00}$ & $148^{+4}_{-1}$ \\
$(5.08\times10^{-4}, 5.01\times10^{-4}, 1.39\times10^{-2})$ & $0.41^{+0.22}_{-0.23}\to 0.41^{+0.22}_{-0.23}$ & $0.30^{+0.03}_{-0.00}\to 0.30^{+0.03}_{-0.00}$ & $0.35^{+0.14}_{-0.04}\to 0.35^{+0.14}_{-0.04}$ & $0.34^{+0.05}_{-0.03}$ & $0.63^{+0.09}_{-0.12}$ & $0.06^{+0.01}_{-0.00}$ & $146^{+0}_{-14}$ \\
$(5.08\times10^{-4}, 1.59\times10^{-4}, 1.76\times10^{-2})$ & $0.25^{+0.32}_{-0.08}\to 0.25^{+0.32}_{-0.08}$ & $0.33^{+0.09}_{-0.03}\to 0.33^{+0.09}_{-0.03}$ & $0.36^{+0.02}_{-0.02}\to 0.36^{+0.02}_{-0.02}$ & $0.34^{+0.12}_{-0.02}$ & $0.63^{+0.00}_{-0.05}$ & $0.06^{+0.01}_{-0.01}$ & $154^{+0}_{-1}$ \\
$(1.05\times10^{-3}, 5.00\times10^{-4}, 1.13\times10^{-1})$ & $0.64^{+0.01}_{-0.21}\to 0.64^{+0.01}_{-0.21}$ & $0.35^{+0.11}_{-0.02}\to 0.35^{+0.11}_{-0.02}$ & $0.50^{+0.07}_{-0.05}\to 0.54^{+0.04}_{-0.08}$ & $0.41^{+0.02}_{-0.09}$ & $0.65^{+0.36}_{-0.12}$ & $0.06^{+0.01}_{-0.00}$ & $129^{+0}_{-1}$ \\
{\boldmath$(3.36\times10^{-3}, 2.94\times10^{-3}, 1.44\times10^{-1})$} & {\boldmath$0.77^{+0.21}_{-0.21}\to 0.74^{+0.24}_{-0.18}$} & {\boldmath$0.64^{+0.02}_{-0.16}\to 0.64^{+0.02}_{-0.11}$} & {\boldmath$0.64^{+0.07}_{-0.22}\to 0.64^{+0.13}_{-0.22}$} & {\boldmath$0.62^{+0.36}_{-0.03}$} & {\boldmath$0.77^{+0.41}_{-0.13}$} & {\boldmath$0.07^{+0.02}_{-0.01}$} & {\boldmath$98^{+7}_{-5}$} \\

\end{tabular}%
}
\end{table}

\subsection{Task Order}
\label{AP:task order}

To investigate the effect of task order, we further benchmarked three additional permutations of the four dynamical systems: Van der Pol, Lorenz-63, R\"ossler, and Chua. The evaluated task sequences are:
\begin{enumerate}
\item R\"ossler $\rightarrow$ Chua $\rightarrow$ Van der Pol $\rightarrow$ Lorenz-63 (RCVL),
\item Chua $\rightarrow$ Van der Pol $\rightarrow$ Lorenz-63 $\rightarrow$ R\"ossler (CVLR),
\item Lorenz-63 $\rightarrow$ R\"ossler $\rightarrow$ Chua $\rightarrow$ Van der Pol (LRCV).
\end{enumerate}
The results are reported in Table~\ref{tab:task_order}. Across all task-order permutations, the model commits between 61 and 70 units, while the overall $D_{\mathrm{stsp}}$ ranges from 1.10 to 2.21 and is consistently determined by the most challenging task, R\"ossler. These results indicate that the performance and capacity efficiency of the method are robust to the order in which tasks are learned. For context, all three alternative task orders yielded lower median overall \(D_{\mathrm{stsp}}\) than any of the other methods in Table~\ref{tab:DSR_canonical_tasks}.

\begin{table}[htbp!]
\caption{\textbf{Task Order Robustness.} Each entry reports performance immediately after the task's own training phase \(\rightarrow\) performance after completion of the full sequence. Values are reported in the format
$\text{median}^{\,Q_3-\text{median}}_{\,Q_1-\text{median}}$ across 10 runs
(lower is better). Overall performance is computed within each run as the maximum final
\(D_{\mathrm{stsp}}\) across tasks and then summarized across runs.
The final task has
no subsequent forgetting period, therefore reported as a single number.}
\label{tab:task_order}
\centering
\renewcommand{\arraystretch}{1.5}
\resizebox{\textwidth}{!}{%
\begin{tabular}{l|cccc|ccc}

\hline

& \multicolumn{4}{c|}{\textbf{$D_{\mathrm{stsp}}$}}

& \multicolumn{3}{c}{\textbf{Overall performance}} \\

\cline{2-5}\cline{6-8}

\textbf{Method}

& \textbf{Blasius}

& \textbf{Laser}

& \textbf{Genesio--Tesi}

& \textbf{Finance}

& \textbf{$D_{\mathrm{stsp}}\downarrow$}

& \textbf{$D_H\downarrow$}
& \textbf{Kept units}
\\

\hline
& & & & & & & \\
LRCV
&$0.12^{+0.03}_{-0.02}$ & $0.39^{+0.14}_{-0.10}\!\rightarrow\!0.42^{+0.11}_{-0.12}$ & $1.10^{+0.45}_{-0.16}\!\rightarrow\!1.10^{+0.75}_{-0.17}$ & $0.75^{+0.06}_{-0.10}\!\rightarrow\!0.79^{+0.03}_{-0.08}$ & $1.10^{+0.75}_{-0.13}$ & $0.06^{+0.02}_{-0.02}$ & $70.0^{+3.5}_{-6.5}$\\
RCVL
&$0.06^{+0.02}_{-0.01}\!\rightarrow\!0.06^{+0.02}_{-0.01}$ & $0.41^{+0.06}_{-0.12}$ & $2.21^{+0.93}_{-0.63}\!\rightarrow\!2.21^{+0.89}_{-0.49}$ & $0.97^{+0.22}_{-0.24}\!\rightarrow\!0.97^{+0.15}_{-0.24}$ & $2.21^{+0.89}_{-0.49}$ & $0.07^{+0.01}_{-0.02}$ & $61.0^{+3.0}_{-3.0}$\\
CVLR
&$0.10^{+0.04}_{-0.03}\!\rightarrow\!0.10^{+0.04}_{-0.02}$ & $0.31^{+0.12}_{-0.06}\!\rightarrow\!0.31^{+0.12}_{-0.06}$ & $1.15^{+0.16}_{-0.14}$ & $0.92^{+0.15}_{-0.17}\!\rightarrow\!0.86^{+0.12}_{-0.11}$ & $1.20^{+0.44}_{-0.11}$ & $0.07^{+0.01}_{-0.00}$ & $65.5^{+3.5}_{-2.5}$\\
VLRC
&$0.15^{+0.16}_{-0.06}\!\rightarrow\!0.15^{+0.16}_{-0.06}$ & $0.40^{+0.13}_{-0.12}\!\rightarrow\!0.41^{+0.06}_{-0.12}$ & $1.22^{+1.02}_{-0.21}\!\rightarrow\!1.26^{+1.16}_{-0.20}$ & $0.98^{+0.12}_{-0.17}$ & $1.26^{+1.16}_{-0.18}$ & $0.08^{+0.02}_{-0.01}$ & $63.0^{+3.0}_{-0.8}$\\

\end{tabular}}
\end{table}

\subsection{Continual Learning Methods}
\label{AP:Continual_Learning_Methods}
\paragraph{Elastic Weight Consolidation (EWC) \citep{Kirkpatrick2017}}
EWC is a successful CL method for feedforward networks \citep{Kirkpatrick2017,Huszar2018}. It penalizes changes to parameters deemed important for previously learned tasks, using the Fisher information matrix (FIM) to estimate parameter importance. For a multi-task setting, the loss is defined as:
\begin{align}
    \mathcal{L}^K =  \mathcal{L}_{\text{task}}^{K} + \lambda_{\mathrm{EWC}} \sum_{k=1}^{K-1} \sum_i F_i^{(k)} \left(\theta_i - \theta_i^{(k)}\right)^2,
    \label{eq:ewc_multitask}
\end{align}
where $\mathcal{L}_{\text{task}}^{K}$, for $K>1$, denotes the loss on the current task $K$, $\theta^{(k)}$ denotes the parameter values obtained after training on task $k$, and $F^{(k)}$ denotes the diagonal of the FIM computed at the end of task $k$:
\begin{align}
    F_i^{(k)} = \mathbb{E}_{x \sim \mathcal{D}_k}\!\left[\left(\frac{\partial \log p_\theta(y \mid x)}{\partial \theta_i}\right)^{2}\right]_{\theta = \theta^{(k)}}.
    \label{eq:fisher_diag}
\end{align}
We evaluated a range of $\lambda_{\text{EWC}}$ values to identify a regime in which the model retains sufficient plasticity to learn new tasks while maintaining sufficient stability to preserve previously acquired tasks. The results are reported in Table~\ref{tab:EWC_DSR_optuna}. No value of $\lambda_{\text{EWC}}$ provided a satisfactory balance between stability and plasticity. Small values were insufficient to prevent parameter overwriting and the resulting forgetting of previously learned tasks. In contrast, larger values increasingly constrained parameter updates, reducing the model’s plasticity and leading to poorer performance even immediately after the corresponding task’s own training, particularly for the third and fourth tasks. These results suggest that, despite its effectiveness in feedforward networks, EWC does not provide an effective continual learning strategy for the recurrent DSR setting considered here.
\begin{table}[htbp!]
\centering
\renewcommand{\arraystretch}{1.3}
\footnotesize

\caption{\textbf{EWC for different values of
$\boldsymbol{\lambda_{\mathrm{EWC}}}$.} For each task, $D_{\text{stsp}}$ is
reported right after that task's own training phase and after the fourth
(final) task's training phase, in the format own $\to$ final, given as
$\text{median}^{\,Q_3-\text{median}}_{\,Q_1-\text{median}}$ over 5
runs (diverged runs excluded; ``div.'' marks that all runs diverged).
For small $\lambda_{\mathrm{EWC}}$, the model can learn all four tasks
but substantially forgets earlier tasks by the end of training. As
$\lambda_{\mathrm{EWC}}$ increases, the model progressively loses
plasticity, impairing the learning of later tasks. Thus, none of the
tested values simultaneously provides sufficient stability and
plasticity. Since all displayed settings yield an infinite median
overall divergence, no setting is uniquely optimal; for the method
comparison, we selected $\lambda_{\mathrm{EWC}}4214.16$ (Optuna best) for the
first benchmark and $\lambda_{\mathrm{EWC}}=10$ for the second
benchmark. ${}^*$ Indicates that only one of five runs remained
non-divergent; hence, $Q_1$ and $Q_3$ are not reported.}
\label{tab:EWC_DSR_optuna}

\resizebox{\textwidth}{!}{%
\begin{tabular}{c|llll|l}
\multicolumn{1}{c}{$\boldsymbol{\lambda_{\mathrm{EWC}}}$}
& \multicolumn{1}{c}{\bf Van der Pol}
& \multicolumn{1}{c}{\bf Lorenz-63}
& \multicolumn{1}{c}{\bf R\"ossler}
& \multicolumn{1}{c}{\bf Chua}
& \multicolumn{1}{c}{\bf Overall $D_{\mathrm{stsp}}$}
\\ \hline
&&&&& \\

$2.325$ & $0.02^{+0.00}_{-0.00}\to \mathrm{div.}$ & $0.24^{+0.05}_{-0.05}\to \mathrm{div.}$ & $0.54^{+1.00}_{-0.08}\to \mathrm{div.}$ & $0.71^{+0.06}_{-0.08}$ & $\mathrm{div.}$ \\
$13.586$ & $0.02^{+0.00}_{-0.00}\to \mathrm{div.}$ & $0.21^{+0.10}_{-0.01}\to \mathrm{div.}$ & $0.75^{+0.23}_{-0.07}\to 12.25^{*}$ & $0.68^{+0.01}_{-0.00}$ & $\mathrm{div.}$ \\
$96.519$ & $0.02^{+0.01}_{-0.01}\to 17.69^{*}$ & $0.26^{+0.03}_{-0.10}\to \mathrm{div.}$ & $3.25^{+2.53}_{-1.22}\to 15.08^{*}$ & $0.74^{+0.15}_{-0.09}$ & $\mathrm{div.}$ \\
$1014.9$ & $0.02^{+0.00}_{-0.01}\to 16.07^{*}$ & $0.28^{+0.00}_{-0.03}\to 12.33^{*}$ & $4.24^{+0.04}_{-0.11}\to 14.80^{+0.18}_{-0.18}$ & $0.72^{+0.07}_{-0.11}$ & $\mathrm{div.}$ \\
$9792.4$ & $0.02^{+0.00}_{-0.01}\to 16.60^{+0.55}_{-0.55}$ & $0.25^{+0.14}_{-0.06}\to 13.73^{+0.12}_{-0.12}$ & $2.60^{+0.72}_{-0.01}\to 15.27^{+0.01}_{-0.94}$ & $1.00^{+0.01}_{-0.22}$ & $\mathrm{div.}$ \\
$89254$ & $0.02^{+0.00}_{-0.00}\to 16.37^{+0.66}_{-0.66}$ & $0.32^{+0.04}_{-0.10}\to 14.44^{+0.32}_{-0.32}$ & $4.79^{+1.21}_{-1.92}\to 15.29^{*}$ & $6.03^{+5.01}_{-3.39}$ & $\mathrm{div.}$ \\

\end{tabular}%
}

\vspace{1em}

\resizebox{\textwidth}{!}{%
\begin{tabular}{c|llll|l}
\multicolumn{1}{c}{$\boldsymbol{\lambda_{\mathrm{EWC}}}$}
& \multicolumn{1}{c}{\bf Blasius}
& \multicolumn{1}{c}{\bf Laser}
& \multicolumn{1}{c}{\bf Genesio--Tesi}
& \multicolumn{1}{c}{\bf Finance}
& \multicolumn{1}{c}{\bf Overall $D_{\mathrm{stsp}}$}
\\ \hline
&&&&& \\

$10^{0}$ & $0.15^{+0.01}_{-0.00}\to \mathrm{div.}$ & $0.21^{+0.02}_{-0.03}\to 15.27^{+0.01}_{-0.05}$ & $0.33^{+0.04}_{-0.05}\to \mathrm{div.}$ & $0.23^{+0.01}_{-0.02}$ & $\mathrm{div.}$ \\
$10^{1}$ & $0.15^{+0.02}_{-0.01}\to \mathrm{div.}$ & $0.18^{+0.02}_{-0.05}\to 15.22^{+0.02}_{-0.01}$ & $0.41^{+0.10}_{-0.12}\to 14.68^{+0.20}_{-0.18}$ & $0.31^{+0.03}_{-0.03}$ & $\mathrm{div.}$ \\
$10^{2}$ & $0.16^{+0.00}_{-0.03}\to \mathrm{div.}$ & $0.21^{+0.01}_{-0.00}\to 15.22^{+0.06}_{-0.00}$ & $0.40^{+0.10}_{-0.07}\to 14.68^{+0.09}_{-0.09}$ & $0.30^{+0.02}_{-0.00}$ & $\mathrm{div.}$ \\
$10^{3}$ & $0.17^{+0.00}_{-0.01}\to \mathrm{div.}$ & $0.19^{+0.03}_{-0.00}\to 15.27^{+0.01}_{-0.18}$ & $0.35^{+0.03}_{-0.02}\to 14.50^{+0.09}_{-0.01}$ & $0.83^{+0.12}_{-0.33}$ & $\mathrm{div.}$ \\
$10^{4}$ & $0.15^{+0.07}_{-0.00}\to \mathrm{div.}$ & $0.30^{+0.08}_{-0.04}\to 15.24^{+0.03}_{-0.20}$ & $0.90^{+0.04}_{-0.33}\to 14.50^{+0.00}_{-0.03}$ & $1.17^{+0.81}_{-0.34}$ & $\mathrm{div.}$ \\
$10^{5}$ & $0.15^{+0.01}_{-0.01}\to \mathrm{div.}$ & $0.27^{+0.07}_{-0.01}\to 14.84^{+0.20}_{-0.20}$ & $2.18^{+0.28}_{-1.04}\to \mathrm{div.}$ & $7.33^{+0.03}_{-0.41}$ & $\mathrm{div.}$ \\

\end{tabular}%
}

\end{table}
\paragraph{Synaptic Intelligence (SI) \citep{Zenke2017}}

\begin{table}[b!]
\centering
\renewcommand{\arraystretch}{1.3}
\footnotesize
\caption{\textbf{SI for different values of $\boldsymbol{\lambda_{\mathrm{SI}}}$.} For each task, the
$D_{\text{stsp}}$ metric is reported immediately after that task’s own
training phase and after training on the fourth (final) task, in the
format own $\to$ final. Values are reported in the format
$\text{median}^{\,Q_3-\text{median}}_{\,Q_1-\text{median}}$ across 5 runs
(lower is better). ``div.'' denotes an infinite median, and ${}^\dagger$ denotes
an infinite upper quartile. For small
$\lambda_{\text{SI}}$, the method is unable to prevent forgetting of
previously learned tasks and also interferes with learning the new
task. As $\lambda_{\text{SI}}$ increases, retention of the first task
improves, but the model progressively loses its plasticity and
eventually fails to learn even the second task: its ``own’’ value
remains high (around 15) rather than decreasing toward zero. Thus,
no value of $\lambda_{\text{SI}}$ achieves both stability and
plasticity simultaneously. ${}^*$ Indicates that only one of five runs remained non-divergent; hence, $Q_1$ and $Q_3$ are not reported.}
\label{tab:SI_DSR_optuna}

\resizebox{\textwidth}{!}{%
\begin{tabular}{c|llll|l}
\multicolumn{1}{c}{$\boldsymbol{\lambda_{\mathrm{SI}}}$}
& \multicolumn{1}{c}{\bf Van der Pol}
& \multicolumn{1}{c}{\bf Lorenz-63}
& \multicolumn{1}{c}{\bf R\"ossler}
& \multicolumn{1}{c}{\bf Chua}
& \multicolumn{1}{c}{\bf Overall $D_{\mathrm{stsp}}$}
\\ \hline
&&&&& \\

$10^{-2}$ & $0.02^{+0.00}_{-0.01}\to \mathrm{div.}$ & $1.06^{+0.28}_{-0.01}\to \mathrm{div.}$ & $7.60^{+1.17}_{-1.09}\to 15.25^{+0.01}_{-1.10}$ & $4.28^{+0.07}_{-0.23}$ & $\mathrm{div.}$ \\
$10^{-1}$ & $0.02^{+0.00}_{-0.01}\to 17.70^{*}$ & $8.83^{+0.17}_{-1.19}\to 15.06^{+0.01}_{-0.46}$ & $12.73^{+0.87}_{-0.40}\to \mathrm{div.}$ & $13.58^{+0.06}_{-0.08}$ & $\mathrm{div.}$ \\
$10^{0}$ & $0.02^{+0.00}_{-0.01}\to 16.45^{+0.63}_{-0.23}$ & $14.78^{+0.28}_{-0.02}\to 14.98^{+0.09}_{-0.12}$ & $14.25^{+0.04}_{-0.09}\to 14.52^{+0.36}_{-0.01}$ & $14.44^{+0.06}_{-0.03}$ & $\mathrm{div.}$ \\
$10^{1}$ & $0.02^{+0.00}_{-0.00}\to 10.07^{+0.64}_{-2.16}$ & $15.03^{+0.03}_{-0.14}\to 15.03^{+0.03}_{-0.11}$ & $14.21^{+0.95}_{-0.05}\to 15.14^{+0.05}_{-0.42}$ & $14.59^{+0.03}_{-0.07}$ & $15.25^{\dagger}_{-0.11}$ \\
$\mathbf{10^{2}}$ & $\mathbf{0.02^{+0.00}_{-0.00}\to 0.31^{+0.07}_{-0.07}}$ & $\mathbf{15.03^{+0.00}_{-0.03}\to 15.03^{+0.00}_{-0.04}}$ & $\mathbf{15.16^{+0.04}_{-0.34}\to 14.83^{+0.35}_{-0.34}}$ & $\mathbf{14.52^{+0.06}_{-0.19}}$ & $\mathbf{15.16^{+0.11}_{-0.13}}$ \\
$10^3$ & $0.02^{+0.00}_{-0.00}\to 0.31^{+0.11}_{-0.28}$ & $15.03^{+0.02}_{-0.00}\to 15.03^{+0.02}_{-0.00}$ & $14.84^{+0.35}_{-0.37}\to 15.17^{+0.03}_{-0.33}$ & $14.51^{+0.00}_{-0.17}$ & $15.24^{\dagger}_{-0.07}$ \\
$1.5\times10^{4}$ & $0.02^{+0.00}_{-0.00}\to 0.02^{+0.00}_{-0.00}$ & $15.03^{+0.02}_{-0.00}\to 15.03^{+0.02}_{-0.00}$ & $14.49^{+0.24}_{-0.08}\to 14.56^{+0.34}_{-0.07}$ & $14.62^{+0.02}_{-0.16}$ & $\mathrm{div.}$ \\
$3\times10^{5}$ & $0.02^{+0.00}_{-0.01}\to 0.02^{+0.00}_{-0.00}$ & $15.03^{+0.01}_{-0.04}\to 15.03^{+0.01}_{-0.04}$ & $15.25^{+0.01}_{-0.06}\to 15.25^{+0.01}_{-0.06}$ & $14.65^{+0.01}_{-0.01}$ & $15.26^{+0.03}_{-0.01}$ \\

\end{tabular}%
}

\vspace{1em}

\resizebox{\textwidth}{!}{%
\begin{tabular}{c|llll|l}
\multicolumn{1}{c}{$\boldsymbol{\lambda_{\mathrm{SI}}}$}
& \multicolumn{1}{c}{\bf Blasius}
& \multicolumn{1}{c}{\bf Laser}
& \multicolumn{1}{c}{\bf Genesio--Tesi}
& \multicolumn{1}{c}{\bf Finance}
& \multicolumn{1}{c}{\bf Overall $D_{\mathrm{stsp}}$}
\\ \hline
&&&&& \\

$10^{-2}$ & $0.17^{+0.03}_{-0.02}\to \mathrm{div.}$ & $0.82^{+0.02}_{-0.01}\to 15.22^{+0.03}_{-1.12}$ & $1.73^{+0.16}_{-0.08}\to 14.86^{*}$ & $2.56^{+0.43}_{-0.44}$ & $\mathrm{div.}$ \\
$10^{-1}$ & $0.14^{+0.02}_{-0.01}\to \mathrm{div.}$ & $12.08^{+0.48}_{-0.16}\to 13.22^{+0.35}_{-0.35}$ & $12.19^{+0.96}_{-2.67}\to 13.31^{+0.59}_{-0.59}$ & $8.83^{+1.28}_{-0.84}$ & $\mathrm{div.}$ \\
$10^{0}$ & $0.16^{+0.01}_{-0.04}\to 14.65^{*}$ & $14.16^{+0.31}_{-0.06}\to 14.62^{+0.03}_{-0.21}$ & $14.21^{+0.29}_{-0.00}\to 14.18^{+0.15}_{-0.03}$ & $14.76^{+0.24}_{-0.03}$ & $\mathrm{div.}$ \\
$10^{1}$ & $0.16^{+0.03}_{-0.04}\to 15.35^{+0.01}_{-0.00}$ & $15.22^{+0.03}_{-0.01}\to 15.22^{+0.05}_{-0.01}$ & $14.46^{+0.04}_{-0.08}\to 14.48^{+0.02}_{-0.02}$ & $15.19^{+0.06}_{-0.03}$ & $15.35^{+0.01}_{-0.00}$ \\
$10^{2}$ & $0.16^{+0.01}_{-0.00}\to 11.60^{+0.52}_{-0.95}$ & $15.24^{+0.03}_{-0.03}\to 15.27^{+0.01}_{-0.05}$ & $14.50^{+0.36}_{-0.00}\to 14.50^{+0.36}_{-0.00}$ & $15.19^{+0.07}_{-0.03}$ & $15.27^{+0.01}_{-0.01}$ \\
$10^{3}$ & $0.15^{+0.02}_{-0.01}\to 0.30^{+0.08}_{-0.13}$ & $15.25^{+0.02}_{-0.04}\to 15.25^{+0.02}_{-0.03}$ & $14.50^{+0.36}_{-0.00}\to 14.50^{+0.36}_{-0.00}$ & $15.18^{+0.00}_{-0.02}$ & $15.25^{+0.02}_{-0.03}$ \\
$\mathbf{10^{4}}$ & $\mathbf{0.17^{+0.02}_{-0.03}\to 0.16^{+0.01}_{-0.00}}$ & $\mathbf{15.25^{+0.02}_{-0.04}\to 15.25^{+0.02}_{-0.04}}$ & $\mathbf{14.50^{+0.35}_{-0.00}\to 14.50^{+0.35}_{-0.00}}$ & $\mathbf{15.01^{+0.13}_{-0.03}}$ & $\mathbf{15.25^{+0.02}_{-0.04}}$ \\
$10^{5}$ & $0.16^{+0.01}_{-0.01}\to 0.15^{+0.02}_{-0.01}$ & $15.25^{+0.01}_{-0.04}\to 15.25^{+0.01}_{-0.04}$ & $14.36^{+0.25}_{-0.12}\to 14.36^{+0.25}_{-0.12}$ & $14.96^{+0.02}_{-0.04}$ & $15.25^{+0.01}_{-0.04}$ \\

\end{tabular}%
}

\end{table}

This method is conceptually similar to EWC but
estimates parameter importance online from the optimization
trajectory rather than from a Fisher-information estimate.
Let $i$ index scalar trainable parameters and $K$ index tasks.
At every optimizer step $t$ within task $K$, a task-specific
accumulator is updated as
\begin{align}
\omega_i^{(K)}
\leftarrow
\omega_i^{(K)}
-g_i^{(K)}(t)\Delta\theta_i^{(K)}(t),
\label{eq:si_omega_update}
\end{align}
where $g_i^{(K)}(t)$ is the gradient of the combined task and
regularization loss before the update, and
$\Delta\theta_i^{(K)}(t)=\theta_i^{(K)}(t+1)-\theta_i^{(K)}(t)$
is the actual optimizer-induced parameter change.
The accumulator is initialized to zero at the start of each task.

At the task boundary, the accumulated contribution is normalized
and added to a persistent importance score:
\begin{align}
\Omega_i^{(\leq K)}
=
\Omega_i^{(<K)}
+
\max\left(
0,\,
\frac{\omega_i^{(K)}}
{(\theta_{i,\mathrm{end}}^{(K)}
-\theta_{i,\mathrm{start}}^{(K)})^2+\xi}
\right),
\label{eq:si_importance}
\end{align}
where $\xi=10^{-3}$, $\theta_{i,\mathrm{start}}^{(K)}$ and $\theta_{i,\mathrm{end}}^{(K)}$
are the parameter value at the beginning and end of task $K$, respectively. The nonnegative clamp discards negative contributions.

During task $K$, the regularized objective is
\begin{align}
\mathcal{L}^{K}
=
\mathcal{L}_{\mathrm{task}}^{K}
+
\lambda_{\mathrm{SI}}
\sum_{i=1}^{|\boldsymbol{\theta}|}
\Omega_i^{(<K)}
\left(
\theta_i-\theta_i^{\mathrm{prev}}
\right)^2,
\label{eq:si_loss}
\end{align}
where $\boldsymbol{\theta}$ contains the trainable entries of
$\va$, $\mW$, $\mC$, and $\vh$. The fixed encoders and decoders are
excluded. The importance vector accumulates contributions
from completed tasks, so the penalty sums over parameter
coordinates without an additional sum over previous tasks.
The anchor $\boldsymbol{\theta}^{\mathrm{prev}}$ is updated
to the selected parameter vector at each task boundary.
For the first task, the importance vector is zero.

Table~\ref{tab:SI_DSR_optuna} reports the search over
$\lambda_{\mathrm{SI}}$. Within the evaluated settings,
smaller regularization strengths permitted substantial forgetting,
whereas larger strengths restricted learning of subsequent tasks. This indicates that SI, like EWC, is not an effective
continual learning strategy for DSR tasks with RNNs.

\paragraph{Replay (ER and GR)}

Replay methods counteract forgetting by revisiting data from previously
learned tasks while a new task is trained. In \emph{Experience Replay}
(ER) \citep{Chaudhry2019}, the stored training data of earlier tasks are
used directly. In \emph{Generative Replay} (GR) \citep{shin2017}, the
real data of earlier tasks are discarded, and the model instead replays
trajectories that it generated itself at the end of each task. In both
variants, all parameters of the AL-RNN are shared across tasks, and
training on the current task $K$ is interleaved with replay steps.
After every $r$ gradient steps on the current task, one additional
gradient step is taken on an earlier task $j$, drawn uniformly at
random from previous tasks $j \sim \{1,\dots,K-1\}$. The replay frequency is thus set by a single
hyperparameter $r$: a smaller $r$ means more frequent replay.

The two variants differ only in the content of the replay buffer. For ER, the buffer for task $j$ contains its complete training trajectory ($10^5$ time steps). For GR, the buffer is generated once, immediately after training on task $j$, using free rollouts of the trained model.
We selected $r$ using an Optuna search (25 trials, 5 seeds per trial, $r \in [1,200]$). A summary of the results is reported in Table~\ref{tab:replay_DSR}. For both benchmarks, we selected $r=1$ as the best-performing replay frequency for both ER and GR and used this setting for the method comparisons in Tables~\ref{tab:DSR_canonical_tasks} and~\ref{tab:DSR_additional_tasks}. In the first benchmark, both replay variants substantially reduce forgetting of the first two tasks, with more frequent replay yielding better retention ($r=1$ performs best). However, no replay frequency preserves the R\"ossler system: its $D_{\text{stsp}}$ increases from approximately 2–4 immediately after training to 5–15 after the fourth task, even with frequent replay. In the second benchmark, frequent replay ($r=1$) substantially reduces forgetting across all tasks. GR performs comparably to ER despite requiring no stored real data, which is advantageous when retaining training data is impractical or prohibited. Moreover, because the AL-RNN is itself a generative model, GR does not require a separate generator for replay.

Overall, replay is more effective than the parameter-regularization methods considered here, but its effectiveness depends on the task sequence. In addition, replay incurs extra computational cost due to the replay steps and requires a replay buffer that grows with the number of previous tasks.
\begin{table}[t]
\centering
\renewcommand{\arraystretch}{1.25}
\setlength{\tabcolsep}{3pt}
\footnotesize

\caption{\textbf{Experience replay (ER) and generative replay (GR) for
different replay intervals $\mathbf{r}$.} For each task, $D_{\text{stsp}}$
is reported right after that task's own training phase and after
training on the fourth (final) task, in the format own $\to$ final,
given as $\text{median}^{\,Q_3-\text{median}}_{\,Q_1-\text{median}}$
over 5 seeds. ``div.'' denotes an infinite median, and ${}^\dagger$ denotes
an infinite upper quartile. Forgetting increases steadily with $r$.}
\label{tab:replay_DSR}

\resizebox{\textwidth}{!}{%
\begin{tabular}{c|l|llll|l}
\multicolumn{1}{c}{\bf Method} &
\multicolumn{1}{c}{$\boldsymbol{r}$} &
\multicolumn{1}{c}{\bf Van der Pol} &
\multicolumn{1}{c}{\bf Lorenz-63} &
\multicolumn{1}{c}{\bf R\"ossler} &
\multicolumn{1}{c}{\bf Chua} &
\multicolumn{1}{c}{\bf Overall $D_{\mathrm{stsp}}$}
\\ \hline
&&&&&& \\

\multirow{6}{*}{\rotatebox[origin=c]{90}{ER}}
& $\boldsymbol{1}$ & $\boldsymbol{0.03^{+0.01}_{-0.00}\to 0.03^{+0.25}_{-0.02}}$ & $\boldsymbol{0.27^{+0.01}_{-0.00}\to 0.27^{+0.01}_{-0.05}}$ & $\boldsymbol{3.84^{+2.40}_{-1.11}\to 10.82^{+0.67}_{-0.96}}$ & $\boldsymbol{1.15^{+0.05}_{-0.35}}$ & $\boldsymbol{10.82^{+0.67}_{-0.96}}$ \\
& $5$ & $0.02^{+0.00}_{-0.00}\to 0.08^{+1.21}_{-0.05}$ & $0.31^{+0.08}_{-0.09}\to 0.38^{+0.04}_{-0.13}$ & $2.36^{+0.21}_{-0.85}\to 11.99^{+0.68}_{-2.74}$ & $0.82^{+0.01}_{-0.13}$ & $11.99^{+0.68}_{-2.74}$ \\
& $37$ & $0.02^{+0.00}_{-0.00}\to 3.46^{+0.15}_{-0.60}$ & $0.31^{+0.04}_{-0.08}\to 1.34^{+0.39}_{-0.04}$ & $2.67^{+2.02}_{-0.93}\to 12.81^{+0.75}_{-1.21}$ & $0.83^{+0.00}_{-0.07}$ & $13.39^{+0.70}_{-1.15}$ \\
& $74$ & $0.02^{+0.00}_{-0.00}\to 6.40^{+0.83}_{-0.83}$ & $0.32^{+0.06}_{-0.10}\to 3.62^{+11.39}_{-0.26}$ & $2.90^{+1.21}_{-0.92}\to 14.10^{+0.09}_{-0.21}$ & $0.60^{+0.17}_{-0.01}$ & $\mathrm{div.}$ \\
& $146$ & $0.02^{+0.00}_{-0.00}\to 17.69^{+0.01}_{-0.16}$ & $0.29^{+0.16}_{-0.12}\to 15.07^{+0.00}_{-0.00}$ & $2.02^{+0.48}_{-0.40}\to 14.71^{+0.27}_{-0.27}$ & $0.58^{+0.14}_{-0.01}$ & $\mathrm{div.}$ \\
& $193$ & $0.02^{+0.01}_{-0.01}\to 17.69^{+0.00}_{-0.07}$ & $0.31^{+0.01}_{-0.09}\to 13.62^{+1.39}_{-1.48}$ & $2.55^{+0.27}_{-1.00}\to 13.48^{+1.74}_{-0.04}$ & $0.59^{+0.19}_{-0.02}$ & $17.70^{\dagger}_{-0.01}$ \\

\hline

\multirow{6}{*}{\rotatebox[origin=c]{90}{GR}}
& $\boldsymbol{1}$ & $\boldsymbol{0.02^{+0.00}_{-0.00}\to 0.11^{+0.06}_{-0.08}}$ & $\boldsymbol{0.25^{+0.08}_{-0.02}\to 0.35^{+0.05}_{-0.07}}$ & $\boldsymbol{3.72^{+0.35}_{-2.50}\to 5.14^{+4.98}_{-0.48}}$ & $\boldsymbol{0.71^{+0.06}_{-0.18}}$ & $\boldsymbol{10.12^{+0.60}_{-5.45}}$ \\
& $5$ & $0.02^{+0.00}_{-0.00}\to 0.27^{+0.51}_{-0.20}$ & $0.24^{+0.13}_{-0.04}\to 0.45^{+0.15}_{-0.11}$ & $4.81^{+0.14}_{-1.26}\to 11.30^{+1.94}_{-3.11}$ & $0.75^{+0.17}_{-0.01}$ & $11.30^{+1.94}_{-3.11}$ \\
& $36$ & $0.02^{+0.00}_{-0.01}\to 4.12^{+5.81}_{-3.29}$ & $0.30^{+0.13}_{-0.13}\to 1.83^{+0.87}_{-0.15}$ & $2.35^{+1.04}_{-1.07}\to 13.05^{+0.18}_{-0.53}$ & $0.67^{+0.01}_{-0.00}$ & $17.69^{\dagger}_{-4.28}$ \\
& $50$ & $0.02^{+0.00}_{-0.00}\to 9.92^{+3.46}_{-2.35}$ & $0.29^{+0.10}_{-0.02}\to 2.65^{+1.85}_{-0.32}$ & $3.12^{+0.35}_{-1.98}\to 13.62^{+0.18}_{-0.64}$ & $0.94^{+0.30}_{-0.20}$ & $13.62^{+4.07}_{-0.64}$ \\
& $74$ & $0.02^{+0.00}_{-0.00}\to 17.03^{+0.66}_{-0.52}$ & $0.33^{+0.01}_{-0.00}\to 6.69^{+1.71}_{-2.46}$ & $2.34^{+1.34}_{-1.52}\to 13.85^{+0.14}_{-0.41}$ & $0.62^{+0.11}_{-0.07}$ & $17.03^{+0.66}_{-0.52}$ \\
& $198$ & $0.03^{+0.03}_{-0.00}\to 14.42^{+2.77}_{-3.19}$ & $0.26^{+0.02}_{-0.01}\to 11.28^{+2.91}_{-2.43}$ & $2.39^{+2.65}_{-0.75}\to 13.65^{+0.20}_{-0.20}$ & $0.63^{+0.04}_{-0.13}$ & $\mathrm{div.}$ \\
\end{tabular}%
}

\vspace{1em}

\resizebox{\textwidth}{!}{%
\begin{tabular}{c|l|llll|l}
\multicolumn{1}{c}{\bf Method} &
\multicolumn{1}{c}{$\boldsymbol{r}$} &
\multicolumn{1}{c}{\bf Blasius} &
\multicolumn{1}{c}{\bf Laser} &
\multicolumn{1}{c}{\bf Genesio--Tesi} &
\multicolumn{1}{c}{\bf Finance} &
\multicolumn{1}{c}{\bf Overall $D_{\mathrm{stsp}}$}
\\ \hline
&&&&&& \\

\multirow{6}{*}{\rotatebox[origin=c]{90}{ER}}
& $\boldsymbol{1}$ & $\boldsymbol{0.14^{+0.02}_{-0.02}\to 0.51^{+0.07}_{-0.34}}$ & $\boldsymbol{0.18^{+0.06}_{-0.00}\to 0.47^{+0.16}_{-0.06}}$ & $\boldsymbol{0.38^{+0.04}_{-0.01}\to 0.95^{+0.56}_{-0.63}}$ & $\boldsymbol{0.31^{+0.01}_{-0.08}}$ & $\boldsymbol{1.51^{+7.68}_{-0.87}}$ \\
& $5$ & $0.13^{+0.02}_{-0.01}\to 0.75^{+0.94}_{-0.02}$ & $0.26^{+0.04}_{-0.03}\to 3.17^{+7.35}_{-2.36}$ & $0.38^{+0.10}_{-0.02}\to 1.21^{+1.03}_{-0.08}$ & $0.30^{+0.02}_{-0.02}$ & $3.17^{+7.35}_{-0.93}$ \\
& $25$ & $0.13^{+0.02}_{-0.01}\to 4.80^{+1.20}_{-0.36}$ & $0.23^{+0.03}_{-0.00}\to 13.01^{+0.02}_{-0.18}$ & $0.33^{+0.00}_{-0.02}\to 12.00^{+1.33}_{-1.56}$ & $0.31^{+0.01}_{-0.01}$ & $13.02^{+0.31}_{-0.02}$ \\
& $50$ & $0.13^{+0.03}_{-0.01}\to 11.89^{+1.43}_{-0.32}$ & $0.28^{+0.02}_{-0.01}\to 13.34^{+0.29}_{-0.17}$ & $0.43^{+0.01}_{-0.01}\to 13.04^{+0.56}_{-6.74}$ & $0.29^{+0.10}_{-0.02}$ & $13.60^{+0.03}_{-0.26}$ \\
& $100$ & $0.16^{+0.01}_{-0.02}\to 13.71^{+0.01}_{-0.01}$ & $0.27^{+0.00}_{-0.08}\to 14.26^{+0.08}_{-0.55}$ & $0.42^{+0.02}_{-0.11}\to 14.29^{+0.26}_{-0.05}$ & $0.26^{+0.08}_{-0.02}$ & $14.35^{+0.21}_{-0.08}$ \\
& $200$ & $0.15^{+0.01}_{-0.01}\to 13.97^{+0.05}_{-0.65}$ & $0.35^{+0.07}_{-0.03}\to 13.91^{+0.46}_{-0.06}$ & $0.37^{+0.05}_{-0.04}\to 14.36^{+0.03}_{-0.22}$ & $0.29^{+0.01}_{-0.02}$ & $14.40^{+0.35}_{-0.03}$ \\

\hline

\multirow{6}{*}{\rotatebox[origin=c]{90}{GR}}
& $\boldsymbol{1}$ & $\boldsymbol{0.16^{+0.00}_{-0.02}\to 0.19^{+0.02}_{-0.01}}$ & $\boldsymbol{0.27^{+0.06}_{-0.11}\to 0.43^{+0.15}_{-0.10}}$ & $\boldsymbol{0.40^{+0.01}_{-0.09}\to 0.58^{+0.03}_{-0.00}}$ & $\boldsymbol{0.24^{+0.03}_{-0.01}}$ & $\boldsymbol{0.61^{+0.05}_{-0.03}}$ \\
& $5$ & $0.15^{+0.01}_{-0.02}\to 1.66^{+0.01}_{-0.17}$ & $0.26^{+0.01}_{-0.09}\to 1.07^{+0.89}_{-0.58}$ & $0.31^{+0.09}_{-0.06}\to 1.68^{+1.16}_{-0.92}$ & $0.29^{+0.05}_{-0.00}$ & $1.96^{+0.89}_{-0.02}$ \\
& $25$ & $0.17^{+0.01}_{-0.01}\to 11.44^{+0.51}_{-9.74}$ & $0.22^{+0.02}_{-0.02}\to 12.71^{+0.45}_{-0.16}$ & $0.40^{+0.04}_{-0.02}\to 4.95^{+2.97}_{-1.53}$ & $0.30^{+0.00}_{-0.07}$ & $12.71^{+0.45}_{-0.16}$ \\
& $50$ & $0.14^{+0.00}_{-0.01}\to 12.76^{+0.64}_{-0.44}$ & $0.26^{+0.03}_{-0.05}\to 13.40^{+0.37}_{-0.65}$ & $0.35^{+0.07}_{-0.02}\to 13.71^{+0.10}_{-0.05}$ & $0.29^{+0.00}_{-0.03}$ & $13.81^{+0.38}_{-0.04}$ \\
& $100$ & $0.14^{+0.04}_{-0.01}\to 13.04^{+0.50}_{-0.15}$ & $0.26^{+0.01}_{-0.02}\to 13.33^{+0.16}_{-0.12}$ & $0.45^{+0.04}_{-0.10}\to 14.45^{+0.02}_{-0.01}$ & $0.35^{+0.00}_{-0.07}$ & $14.45^{+0.02}_{-0.01}$ \\
& $200$ & $0.17^{+0.00}_{-0.03}\to 14.10^{+0.01}_{-0.29}$ & $0.25^{+0.02}_{-0.04}\to 13.98^{+0.22}_{-0.14}$ & $0.36^{+0.16}_{-0.00}\to 14.23^{+0.14}_{-0.05}$ & $0.28^{+0.09}_{-0.02}$ & $14.66^{+0.11}_{-0.39}$ \\
\end{tabular}%
}

\end{table}

\paragraph{Context-dependent Gating (XdG) \citep{Masse2018}}
XdG is a random gating method: each task is assigned an independently
drawn binary mask over the latent units, which is fixed before training
on that task and never updated. Each task keeps its own readout units
(one per observed dimension), which are always active for that task and
never shared with other tasks. Each of the non-readout latent units is
independently active for a task with probability $1 - P_{\text{XdG}}$.
Inactive units are gated out of the latent dynamics and therefore
receive no gradient, while all parameters remain trainable. Only the
gating component of XdG is used here, without the additional synaptic
stabilization combined with it in \citet{Masse2018}.

Table~\ref{tab:XdG_DSR} reports results over a range of
$P_{\text{XdG}}$ for both benchmarks. For small $P_{\text{XdG}}$, the masks of different
tasks overlap strongly, so learning a new task overwrites the units of
earlier tasks. Increasing $P_{\text{XdG}}$
reduces the overlap, but even a small number of shared units suffices
to disrupt earlier tasks: at $P_{\text{XdG}} = 0.90$, two tasks share
on average only about 1.5 units, yet the Van der Pol model still
diverges in all seeds and Lorenz-63 is largely forgotten. At the same
time, the capacity available to each task shrinks (about 15 units at
$P_{\text{XdG}} = 0.90$ and 7 at $0.95$), and the tasks are no longer
learned well in the first place. At $P_{\text{XdG}} = 0.95$, the
performance measured after a task's own training changes little by the
end of training. No value of $P_{\text{XdG}}$ achieves both stability and plasticity.

The method is conceptually close to CRUG, in that
both restrict each task to a subset of latent units. However, XdG draws
its masks at random, does not guarantee disjoint masks across tasks,
and applies a single, fixed gating probability uniformly to every task,
regardless of task difficulty. In contrast, CRUG learns disjoint subspaces by construction and allocates
capacity adaptively, allowing the model to decide how many units each
task requires based on its difficulty.
\begin{table}[t]
\centering
\renewcommand{\arraystretch}{1.25}
\setlength{\tabcolsep}{3pt}
\footnotesize

\caption{\textbf{Context-dependent gating for different values of
$\mathbf{P_{\text{XdG}}}$.} The probability that a given unit is masked out
(inactive) for a task is determined by $P_{\text{XdG}}$. For each task, $D_{\text{stsp}}$ is reported
right after that task's own training phase and after the fourth
(final) task's training phase, in the format own $\to$ final, given as
$\text{median}^{\,Q_3-\text{median}}_{\,Q_1-\text{median}}$ over 5
seeds (diverged seeds excluded; ``div.'' marks that all seeds
diverged). For small $P_{\text{XdG}}$, task masks overlap
substantially, and training later tasks causes earlier tasks to
diverge. As $P_{\text{XdG}}$ increases, overlap decreases, but even a
few shared units still disrupt earlier tasks, while the capacity
available to each task becomes too limited for the tasks to be learned
well ($P_{\text{XdG}} = 0.90$ and $0.95$). ${}^*$ Indicates that only one of five runs remained non-divergent; hence, $Q_1$ and $Q_3$ are not reported.}
\label{tab:XdG_DSR}

\resizebox{\textwidth}{!}{%
\begin{tabular}{c|llll|l}
\multicolumn{1}{c}{$\boldsymbol{P_{\mathrm{XdG}}}$}
& \multicolumn{1}{c}{\bf Van der Pol}
& \multicolumn{1}{c}{\bf Lorenz-63}
& \multicolumn{1}{c}{\bf R\"ossler}
& \multicolumn{1}{c}{\bf Chua}
& \multicolumn{1}{c}{\bf Overall $D_{\mathrm{stsp}}$}
\\ \hline
&&&&& \\

$0.08$ & $0.01^{+0.00}_{-0.00}\to \mathrm{div.}$ & $0.37^{+0.07}_{-0.15}\to 14.54^{+0.26}_{-0.33}$ & $1.48^{+2.41}_{-0.14}\to \mathrm{div.}$ & $0.78^{+0.17}_{-0.20}$ & $\mathrm{div.}$ \\
$0.29$ & $0.04^{+0.01}_{-0.02}\to 17.23^{+0.23}_{-0.23}$ & $0.25^{+0.03}_{-0.02}\to 15.06^{+0.02}_{-0.04}$ & $4.71^{+1.70}_{-1.44}\to \mathrm{div.}$ & $0.78^{+0.05}_{-0.05}$ & $\mathrm{div.}$ \\
$0.49$ & $0.02^{+0.00}_{-0.00}\to \mathrm{div.}$ & $0.35^{+0.06}_{-0.04}\to 14.89^{+0.13}_{-0.36}$ & $2.23^{+0.48}_{-0.53}\to \mathrm{div.}$ & $0.64^{+0.05}_{-0.11}$ & $\mathrm{div.}$ \\
$0.70$ & $0.05^{+0.01}_{-0.00}\to \mathrm{div.}$ & $0.35^{+0.18}_{-0.11}\to 15.06^{+0.00}_{-0.39}$ & $2.60^{+0.28}_{-1.23}\to \mathrm{div.}$ & $0.60^{+0.03}_{-0.04}$ & $\mathrm{div.}$ \\
$0.80$ & $0.06^{+0.01}_{-0.03}\to \mathrm{div.}$ & $0.49^{+0.18}_{-0.06}\to 15.04^{+0.03}_{-0.35}$ & $2.98^{+0.39}_{-0.49}\to \mathrm{div.}$ & $0.53^{+0.13}_{-0.04}$ & $\mathrm{div.}$ \\
$0.85$ & $0.07^{+0.01}_{-0.02}\to 15.18^{*}$ & $0.85^{+0.13}_{-0.00}\to 14.97^{+0.04}_{-0.32}$ & $3.31^{+0.57}_{-1.72}\to \mathrm{div.}$ & $0.55^{+0.22}_{-0.02}$ & $\mathrm{div.}$ \\
$0.90$ & $1.58^{+1.22}_{-0.79}\to \mathrm{div.}$ & $2.25^{+2.32}_{-0.55}\to 11.06^{+2.84}_{-2.34}$ & $3.09^{+1.75}_{-0.99}\to 13.05^{+0.22}_{-0.51}$ & $0.69^{+0.09}_{-0.15}$ & $\mathrm{div.}$ \\
$\mathbf{0.94}$ & $\mathbf{14.04^{+0.73}_{-3.29}\to 16.47^{+0.61}_{-1.13}}$ & $\mathbf{10.39^{+3.83}_{-6.17}\to 10.39^{+3.86}_{-6.17}}$ & $\mathbf{4.66^{+1.04}_{-1.15}\to 4.66^{+1.04}_{-1.15}}$ & $\mathbf{14.65^{+0.01}_{-13.95}}$ & $\mathbf{17.70^{\dagger}_{-1.22}}$ \\
\end{tabular}%
}

\vspace{1em}

\resizebox{\textwidth}{!}{%
\begin{tabular}{c|llll|l}
\multicolumn{1}{c}{$\boldsymbol{P_{\mathrm{XdG}}}$}
& \multicolumn{1}{c}{\bf Blasius}
& \multicolumn{1}{c}{\bf Laser}
& \multicolumn{1}{c}{\bf Genesio--Tesi}
& \multicolumn{1}{c}{\bf Finance}
& \multicolumn{1}{c}{\bf Overall $D_{\mathrm{stsp}}$}
\\ \hline
&&&&& \\
$0.10$ & $0.21^{+0.05}_{-0.03}\to 15.36^{+0.01}_{-0.01}$ & $0.25^{+0.01}_{-0.02}\to 15.29^{*}$ & $0.29^{+0.02}_{-0.00}\to 14.50^{+0.00}_{-0.02}$ & $0.28^{+0.00}_{-0.01}$ & $\mathrm{div.}$ \\
$0.30$ & $0.23^{+0.00}_{-0.04}\to 15.35^{*}$ & $0.26^{+0.01}_{-0.06}\to 15.26^{+0.01}_{-0.01}$ & $0.41^{+0.06}_{-0.15}\to 14.64^{+0.24}_{-0.24}$ & $0.38^{+0.00}_{-0.11}$ & $\mathrm{div.}$ \\
$0.50$ & $0.34^{+0.01}_{-0.14}\to 15.35^{+0.02}_{-0.11}$ & $0.39^{+0.13}_{-0.01}\to 15.28^{*}$ & $0.30^{+0.02}_{-0.02}\to 14.49^{+0.12}_{-0.02}$ & $0.31^{+0.05}_{-0.01}$ & $\mathrm{div.}$ \\
$0.70$ & $0.26^{+0.12}_{-0.01}\to \mathrm{div.}$ & $1.18^{+3.18}_{-0.40}\to 15.22^{+0.03}_{-0.33}$ & $0.50^{+0.13}_{-0.13}\to 12.60^{+0.95}_{-0.18}$ & $0.62^{+0.55}_{-0.11}$ & $\mathrm{div.}$ \\
$0.80$ & $0.82^{+0.16}_{-0.24}\to \mathrm{div.}$ & $12.00^{+0.23}_{-1.83}\to 12.95^{+1.14}_{-1.14}$ & $0.50^{+0.33}_{-0.19}\to 13.21^{+0.32}_{-0.96}$ & $1.22^{+0.78}_{-0.42}$ & $\mathrm{div.}$ \\
$0.85$ & $1.04^{+0.01}_{-0.54}\to \mathrm{div.}$ & $12.36^{+0.23}_{-0.19}\to 13.84^{+0.17}_{-0.17}$ & $0.66^{+0.28}_{-0.15}\to 14.41^{+0.26}_{-0.15}$ & $4.62^{+5.04}_{-3.28}$ & $\mathrm{div.}$ \\
$0.90$ & $2.12^{+0.99}_{-1.24}\to 10.99^{+2.07}_{-2.07}$ & $12.18^{+0.12}_{-0.96}\to 13.87^{+0.76}_{-0.67}$ & $0.91^{+0.01}_{-0.50}\to 12.05^{+1.75}_{-4.03}$ & $8.38^{+1.18}_{-2.17}$ & $\mathrm{div.}$ \\
$\mathbf{0.95}$ & $\mathbf{5.62^{+0.45}_{-3.27}\to 3.96^{+1.77}_{-1.69}}$ & $\mathbf{14.13^{+0.17}_{-1.38}\to 13.44^{+0.97}_{-2.67}}$ & $\mathbf{12.96^{+0.87}_{-11.55}\to 12.96^{+0.87}_{-11.55}}$ & $\mathbf{14.99^{+0.16}_{-2.16}}$ & $\mathbf{15.24^{\dagger}_{-0.38}}$ \\
\end{tabular}%
}

\end{table}

\paragraph{Continual Learning with Neuron Pruning (CLNP) \citep{Golkar2019}}
CLNP is another method from the parameter-isolation family that assigns each task its own set of units and protects them from subsequent tasks. During training on a task, an $L_1$ penalty on the weights encourages units to become inactive. After training, units with low average activity are pruned, and the remaining active units are committed to the task. For the AL-RNN, the incoming parameters of latent unit $i$ comprise its diagonal term $A_i$, its row of recurrent weights $W_{i\cdot}$, its input weights $C_{i\cdot}$, and its bias $h_i$. The $L_1$ penalty is applied to the incoming parameters of all units that have not yet been committed,
\begin{align}
\mathcal{L}^k = \mathcal{L}_{\text{task}}^{k}
+ \alpha^{\mathrm{lin}} \sum_{i \in \sU_k^{\mathrm{lin}}} \Omega_i
+ \alpha^{\mathrm{ReLU}} \sum_{i \in \sU_k^{\mathrm{ReLU}}} \Omega_i,
\qquad
\Omega_i = |A_{ii}| + \sum_{j=1}^{M} |W_{ij}| + \sum_{q=1}^{Q}  |C_{iq}| + |h_i|,
\label{eq:clnp_loss}
\end{align}
where $\sU_k^{\mathrm{lin}}$ and $\sU_k^{\mathrm{ReLU}}$ denote the free linear and ReLU units, respectively. We use separate regularization coefficients for linear and ReLU units, analogous to those in CRUG, to account for their distinct roles in the AL-RNN; the original CLNP uses a single coefficient because its network contains only ReLU units. 

After training, the average activity
$\bar a_i = \mathbb{E}\,|z_i|$ of each free unit is measured over free-running trajectories of the model and normalized by the average activity of the task’s readout units. This normalization is necessary because, unlike ReLU units, linear units have no inactive state at zero, making a single absolute activity threshold inappropriate for both unit types. Unit $i$ is retained if $\bar a_i > \tau_a$, while the readout units of each task are never pruned. Following the graceful-forgetting procedure of \cite{Golkar2019}, $\tau_a$ is chosen separately for each task as the largest value on a logarithmic grid for which the $D_{\text{stsp}}$ of the pruned model remains within a relative margin $m$ of the corresponding unpruned model.

Once a task is completed, the incoming parameters of its retained units are frozen. Weights from committed to free units remain trainable, allowing later tasks to read from units assigned to earlier tasks and thereby enabling forward transfer. In contrast, weights from free to committed units are set to zero. The parameters of all pruned units are re-initialized and returned to the free pool for the next task. Consequently, training on later tasks cannot alter the dynamics of earlier tasks. We verified that the parameters of committed units remained exactly unchanged in every run.

We searched over $\alpha^{\text{ReLU}}$, $\alpha^{\text{lin}}$, and $m$ using Optuna, with 40 trials and 10 seeds per trial for the first benchmark and 20 trials and 5 seeds per trial for the second benchmark. Table~\ref{tab:CLNP_DSR} reports the 10 best-performing trials for each task sequence. For the first task sequence, we selected the second-best trial for Table~\ref{tab:DSR_canonical_tasks}, as its overall performance was comparable to that of the best trial ($3.72$ vs. $3.81$) while requiring fewer units ($123.5$ vs.\ $133.0$) and resulting in fewer exhausted seeds ($3/10$ vs.\ $4/10$). As expected, CLNP prevents forgetting: the $D_{\text{stsp}}$ of each task after training on all four tasks is essentially identical to its value immediately after its own training. However, no setting learns the tasks well in the first place. The best settings reach $D_{\text{stsp}}$ values of about 2–4 on Van der Pol and Lorenz-63, substantially higher than the values of about 0.02 and 0.2–0.3, respectively, achieved by the other methods immediately after training on these tasks. This poor initial learning reflects a conflict between the two roles of the $L_1$ penalty. For small $\alpha$, the penalty does not silence enough units: the first task retains most of the available units, leaving insufficient capacity for subsequent tasks. Such runs terminate before completing all tasks, which is why several settings contain results for only a subset of seeds. CRUG avoids this conflict because its $L_0$ gates penalize the number of active units directly, without shrinking the weights of units that are ultimately retained.
\begin{table}[htbp!]
\centering
\renewcommand{\arraystretch}{1.3}
\footnotesize

\caption{\textbf{CLNP for different $L_1$ coefficients and graceful-forgetting margins.} For each task, $D_{\text{stsp}}$ is reported
right after that task's own training phase and after the fourth
(final) task's training phase, in the format own $\to$ final, given as
$\text{median}^{\,Q_3-\text{median}}_{\,Q_1-\text{median}}$ over the seeds (out of 10) that completed training ($n$). Divergent or incomplete runs are not omitted but counted as worst results in the median.
Overall performance is the maximum final task $D_{\text{stsp}}$ within each run, summarized across all runs.
``div.'' denotes an infinite median, and ${}^\dagger$ denotes
an infinite upper quartile. The last column gives the
median total number of units kept across all four tasks (out of
$M = 160$).}
\label{tab:CLNP_DSR}


\resizebox{\textwidth}{!}{%

\begin{tabular}{c|c|llll|ll}

\multicolumn{1}{c}{$\boldsymbol{(\alpha_{\text{ReLU}},
\alpha_{\text{lin}},m)}$}

& \multicolumn{1}{c}{$\boldsymbol{n}$}

& \multicolumn{1}{c}{\bf Van der Pol}

& \multicolumn{1}{c}{\bf Lorenz-63}

& \multicolumn{1}{c}{\bf R\"ossler}

& \multicolumn{1}{c}{\bf Chua}

& \multicolumn{1}{c}{\bf overall $D_{\text{stsp}}$}

& \multicolumn{1}{c}{\bf Total units kept}

\\ \hline

& & & & & & & \\
$(1.76\times10^{-3}, 1.58\times10^{-3}, 0.0215)$ & $6/10$ & $3.62^{\dagger}_{-1.41}\to 3.62^{\dagger}_{-1.41}$ & $2.19^{\dagger}_{-0.27}\to 2.19^{\dagger}_{-0.27}$ & $3.31^{\dagger}_{-0.75}\to 3.31^{\dagger}_{-0.75}$ & $2.60^{\dagger}_{-0.46}\to 2.60^{\dagger}_{-0.46}$ & $3.72^{\dagger}_{-0.43}$ & $133.0^{\dagger}_{-58.8}$ \\
$\mathbf{(1.67\times10^{-3}, 1.56\times10^{-3}, 0.0200)}$ & $\mathbf{7/10}$ & $\mathbf{2.59^{\dagger}_{-1.54}\to 2.59^{\dagger}_{-1.54}}$ & $\mathbf{2.15^{\dagger}_{-0.51}\to 2.15^{\dagger}_{-0.51}}$ & $\mathbf{3.53^{\dagger}_{-1.22}\to 3.53^{\dagger}_{-1.29}}$ & $\mathbf{2.44^{\dagger}_{-0.07}\to 2.44^{\dagger}_{-0.07}}$ & $\mathbf{3.81^{\dagger}_{-1.09}}$ & $\mathbf{123.5^{\dagger}_{-47.0}}$ \\
$(1.55\times10^{-3}, 1.51\times10^{-3}, 0.0372)$ &  $7/10$ & $2.66^{\dagger}_{-0.93}\to 2.66^{\dagger}_{-0.93}$ & $2.03^{\dagger}_{-0.23}\to 2.00^{\dagger}_{-0.20}$ & $3.86^{\dagger}_{-1.32}\to 3.88^{\dagger}_{-1.24}$ & $2.28^{\dagger}_{-0.39}\to 2.17^{\dagger}_{-0.32}$ & $4.87^{\dagger}_{-1.45}$ & $98.0^{\dagger}_{-12.8}$ \\
$(1.73\times10^{-3}, 1.71\times10^{-3}, 0.0222)$ &  $7/10$ & $2.79^{+3.20}_{-1.48}\to 2.79^{+3.20}_{-1.48}$ & $1.58^{+10.20}_{-0.54}\to 1.58^{+10.20}_{-0.54}$ & $3.33^{\dagger}_{-0.98}\to 3.33^{\dagger}_{-0.98}$ & $2.61^{\dagger}_{-0.54}\to 2.61^{\dagger}_{-0.54}$ & $5.90^{\dagger}_{-2.42}$ & $106.5^{+37.2}_{-36.8}$ \\
$(3.77\times10^{-3}, 2.13\times10^{-3}, 0.0966)$ & $8/10$ & $5.28^{+3.30}_{-0.98}\to 5.28^{+3.30}_{-0.98}$ & $3.34^{+1.01}_{-0.72}\to 3.26^{+0.95}_{-0.65}$ & $3.98^{+1.12}_{-0.58}\to 3.98^{+1.14}_{-0.50}$ & $7.82^{+5.10}_{-2.62}\to 7.82^{+5.10}_{-2.76}$ & $10.82^{+3.81}_{-4.72}$ & $101.0^{+7.0}_{-22.8}$ \\
$(3.13\times10^{-3}, 1.77\times10^{-3}, 0.0749)$ &  $6/10$ & $4.10^{\dagger}_{-1.75}\to 4.10^{\dagger}_{-1.75}$ & $2.42^{\dagger}_{-0.23}\to 2.34^{\dagger}_{-0.11}$ & $5.75^{\dagger}_{-2.72}\to 4.56^{\dagger}_{-1.53}$ & $11.47^{\dagger}_{-7.94}\to 11.47^{\dagger}_{-7.89}$ & $11.47^{\dagger}_{-7.28}$ & $120.0^{\dagger}_{-40.8}$ \\
$(1.89\times10^{-3}, 1.80\times10^{-3}, 0.0234)$ & $6/10$ & $5.15^{\dagger}_{-2.32}\to 5.15^{\dagger}_{-2.32}$ & $8.98^{\dagger}_{-6.94}\to 9.02^{\dagger}_{-7.03}$ & $3.72^{\dagger}_{-0.30}\to 3.50^{\dagger}_{-0.39}$ & $5.96^{\dagger}_{-3.70}\to 5.93^{\dagger}_{-3.36}$ & $12.06^{\dagger}_{-8.46}$ & $110.5^{\dagger}_{-29.2}$ \\
$(2.36\times10^{-3}, 1.14\times10^{-3}, 0.0482)$ & $6/10$ & $4.32^{\dagger}_{-2.29}\to 4.32^{\dagger}_{-2.29}$ & $1.97^{\dagger}_{-0.35}\to 1.97^{\dagger}_{-0.35}$ & $3.59^{\dagger}_{-0.88}\to 3.59^{\dagger}_{-0.88}$ & $13.56^{\dagger}_{-10.78}\to 13.56^{\dagger}_{-10.78}$ & $13.56^{\dagger}_{-8.97}$ & $125.5^{\dagger}_{-16.8}$ \\
$(6.46\times10^{-3}, 3.80\times10^{-3}, 0.3347)$ & $6/10$ & $5.85^{+0.95}_{-1.21}\to 5.85^{+0.95}_{-1.21}$ & $3.65^{+0.32}_{-1.24}\to 3.65^{+0.32}_{-1.24}$ & $4.05^{+0.96}_{-0.90}\to 4.05^{+0.88}_{-0.90}$ & $14.39^{\dagger}_{-3.11}\to 14.39^{\dagger}_{-3.11}$ & $14.39^{\dagger}_{-3.11}$ & $56.0^{+12.0}_{-5.8}$ \\
$(9.90\times10^{-3}, 1.52\times10^{-3}, 0.0214)$ & $6/10$ & $11.26^{+2.85}_{-2.93}\to 11.26^{+2.85}_{-2.93}$ & $4.05^{+0.75}_{-0.68}\to 4.01^{+0.78}_{-0.64}$ & $5.62^{\dagger}_{-1.59}\to 5.62^{\dagger}_{-1.59}$ & $12.90^{\dagger}_{-1.71}\to 12.90^{\dagger}_{-1.73}$ & $15.66^{\dagger}_{-3.83}$ & $57.0^{+22.5}_{-11.2}$ \\
\hline
\end{tabular}%
}

\vspace{1em}

\resizebox{\textwidth}{!}{%

\begin{tabular}{c|c|llll|ll}

\multicolumn{1}{c}{$\boldsymbol{(\alpha_{\text{ReLU}},
\alpha_{\text{lin}},m)}$}

& \multicolumn{1}{c}{$\boldsymbol{n}$}

& \multicolumn{1}{c}{\bf Blasius}

& \multicolumn{1}{c}{\bf Laser}

& \multicolumn{1}{c}{\bf Genesio--Tesi}

& \multicolumn{1}{c}{\bf Finance}

& \multicolumn{1}{c}{\bf overall $D_{\text{stsp}}$}

& \multicolumn{1}{c}{\bf Total units kept}

\\ \hline

& & & & & & & \\
$\mathbf{(1.53\times10^{-3}, 1.18\times10^{-3}, 0.6615)}$ & $\mathbf{5/5}$ & $\mathbf{3.17^{+0.47}_{-0.46}\to 3.17^{+0.47}_{-0.46}}$ & $\mathbf{2.51^{+0.00}_{-0.65}\to 2.51^{+0.00}_{-0.65}}$ & $\mathbf{3.23^{+0.40}_{-1.21}\to 3.23^{+0.40}_{-1.21}}$ & $\mathbf{1.17^{+0.05}_{-0.37}\to 1.17^{+0.05}_{-0.37}}$ & $\mathbf{4.72^{+0.38}_{-1.09}}$ & $\mathbf{103.0^{+11.0}_{-15.0}}$ \\
$(1.51\times10^{-3}, 1.31\times10^{-3}, 0.5905)$ & $4/5$ & $3.06^{+0.30}_{-0.10}\to 3.06^{+0.30}_{-0.10}$ & $3.90^{+10.08}_{-0.92}\to 3.90^{+10.08}_{-0.92}$ & $4.45^{+1.70}_{-3.09}\to 4.45^{+1.70}_{-3.09}$ & $1.27^{+0.29}_{-0.06}\to 1.27^{+0.29}_{-0.06}$ & $6.15^{+7.83}_{-1.70}$ & $117.0^{+31.0}_{-24.0}$ \\
$(1.75\times10^{-3}, 6.63\times10^{-4}, 0.2991)$ & $3/5$ & $2.92^{\dagger}_{-0.18}\to 2.92^{\dagger}_{-0.18}$ & $7.64^{\dagger}_{-6.07}\to 7.64^{\dagger}_{-6.07}$ & $1.73^{\dagger}_{-0.36}\to 1.73^{\dagger}_{-0.36}$ & $1.37^{\dagger}_{-0.36}\to 1.37^{\dagger}_{-0.36}$ & $7.64^{\dagger}_{-4.91}$ & $130.0^{\dagger}_{-22.0}$ \\
$(1.77\times10^{-3}, 1.31\times10^{-3}, 0.0888)$ &$4/5$ & $3.22^{+0.15}_{-0.17}\to 3.22^{+0.15}_{-0.17}$ & $7.48^{+5.66}_{-1.79}\to 7.48^{+5.66}_{-1.79}$ & $2.57^{+5.97}_{-1.05}\to 2.57^{+5.97}_{-1.05}$ & $1.21^{+0.59}_{-0.34}\to 1.21^{+0.59}_{-0.34}$ & $8.54^{+4.60}_{-1.06}$ & $134.0^{+26.0}_{-55.0}$ \\
$(1.74\times10^{-3}, 1.63\times10^{-3}, 0.0235)$ & $4/5$ & $2.63^{+0.62}_{-0.13}\to 2.63^{+0.62}_{-0.13}$ & $8.87^{+3.65}_{-7.47}\to 8.87^{+3.65}_{-7.47}$ & $4.03^{+1.85}_{-2.15}\to 4.03^{+1.85}_{-2.15}$ & $1.00^{+0.13}_{-0.04}\to 1.00^{+0.13}_{-0.04}$ & $8.87^{+3.65}_{-3.00}$ & $160.0^{+0.0}_{-1.0}$ \\
$(2.10\times10^{-3}, 1.80\times10^{-3}, 0.4857)$ &  $4/5$ & $3.40^{+1.33}_{-0.14}\to 3.40^{+1.33}_{-0.14}$ & $10.15^{+0.46}_{-7.40}\to 10.15^{+0.46}_{-7.40}$ & $7.82^{+0.45}_{-5.96}\to 7.82^{+0.45}_{-5.96}$ & $1.47^{+0.22}_{-0.28}\to 1.47^{+0.22}_{-0.28}$ & $10.15^{+0.46}_{-1.89}$ & $84.0^{+27.0}_{-5.0}$ \\
$(5.26\times10^{-3}, 1.40\times10^{-3}, 0.2503)$ &  $5/5$ & $4.70^{+0.04}_{-0.50}\to 4.70^{+0.04}_{-0.50}$ & $4.59^{+1.01}_{-0.83}\to 4.59^{+1.01}_{-0.83}$ & $9.95^{+0.65}_{-0.61}\to 9.95^{+0.65}_{-0.61}$ & $1.39^{+0.21}_{-0.13}\to 1.39^{+0.21}_{-0.13}$ & $10.60^{+3.38}_{-0.65}$ & $83.0^{+17.0}_{-0.0}$ \\
$(3.63\times10^{-3}, 2.25\times10^{-3}, 0.1583)$ & $4/5$ & $4.11^{+1.03}_{-0.24}\to 4.11^{+1.03}_{-0.24}$ & $7.64^{+0.29}_{-1.44}\to 7.64^{+0.29}_{-1.44}$ & $11.52^{+2.29}_{-1.43}\to 11.52^{+2.29}_{-1.43}$ & $1.47^{+0.14}_{-0.06}\to 1.47^{+0.14}_{-0.06}$ & $11.52^{+2.29}_{-1.43}$ & $64.0^{+16.0}_{-2.0}$ \\
$(2.74\times10^{-3}, 1.03\times10^{-3}, 0.7645)$ & $3/5$ & $4.34^{\dagger}_{-0.39}\to 4.34^{\dagger}_{-0.39}$ & $11.88^{\dagger}_{-3.84}\to 11.88^{\dagger}_{-3.84}$ & $11.92^{\dagger}_{-2.25}\to 11.92^{\dagger}_{-2.25}$ & $0.98^{\dagger}_{-0.08}\to 0.98^{\dagger}_{-0.08}$ & $11.92^{\dagger}_{-2.25}$ & $110.0^{\dagger}_{-9.0}$ \\
$(5.55\times10^{-3}, 4.67\times10^{-3}, 0.0264)$ & $5/5$ & $4.45^{+1.05}_{-0.12}\to 4.45^{+1.05}_{-0.12}$ & $4.35^{+5.07}_{-1.25}\to 4.35^{+5.07}_{-1.25}$ & $13.75^{+0.01}_{-0.93}\to 13.75^{+0.01}_{-0.93}$ & $2.08^{+0.04}_{-0.17}\to 2.08^{+0.04}_{-0.17}$ & $13.75^{+0.01}_{-0.93}$ & $74.0^{+19.0}_{-16.0}$ \\

\end{tabular}%
}

\end{table}


\subsection{Transfer for DSR Tasks from the Lorenz Family}
\label{AP:Lorenz_family}
\begin{figure}[htbp!]
    \centering
    \includegraphics[width=\linewidth]{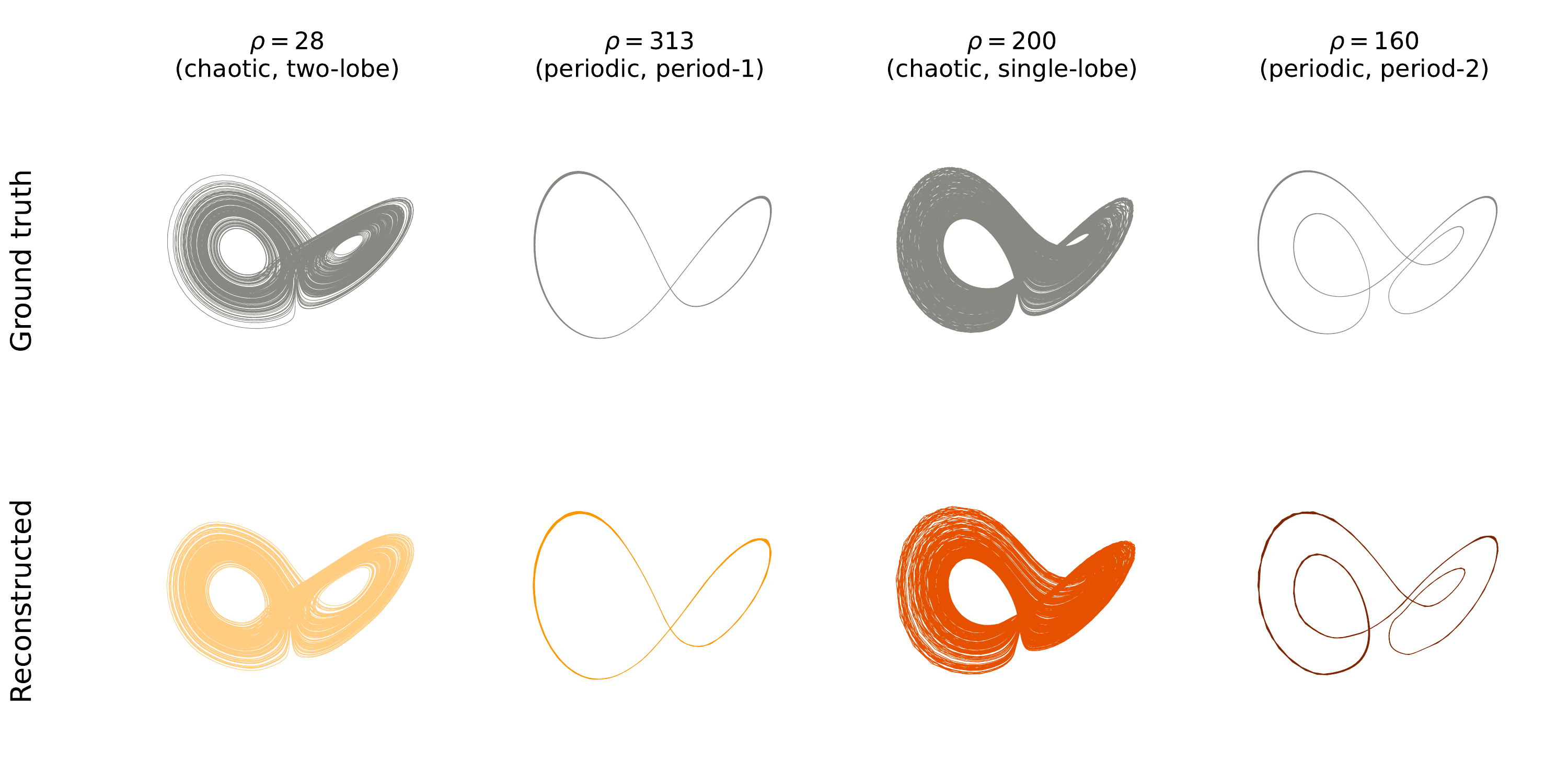}
    \caption{Reconstructions of the Lorenz-63 family DSR tasks ($\rho \in \{28, 313, 200, 160\}$) by a trained model (colored), shown together with the ground-truth trajectories (gray). The model successfully reconstructs all four dynamical systems.}
    \label{fig:Lorenz_family_dsr}
\end{figure}

To investigate how transfer behaves when tasks share dynamical similarities, we trained on an additional task sequence drawn from the same dynamical system family, the Lorenz-63 system, under different parametrizations ($\rho \in \{28, 313, 200, 160\}$). Fig.~\ref{fig:Lorenz_family_dsr} shows the final reconstructions of a trained model alongside the ground-truth data, demonstrating that the model successfully reconstructed all four dynamical systems.

We tuned the $L_1$, $L_2$, and $L_0$ transfer regularizations, running 40 trials with 10 seeds each, and compared them against the unregularized and no-transfer variants, using Pareto fronts (Fig.~\ref{fig:DSR_task_pareto_fronts}B). All transfer variants are clearly separated from the no-transfer baseline. In contrast, the corresponding plot for the heterogeneous task sequence (Fig.~\ref{fig:DSR_task_pareto_fronts}A) shows little difference between variants with transfer and no transfer. This indicates that transfer yields substantially greater performance gains when tasks are dynamically related than when they are heterogeneous. Moreover, the ratio $\mathrm{RMS}(\text{transfer block}) / \mathrm{RMS}(\text{own block})$ in Fig.~\ref{fig:DSR_task_pareto_fronts}C highlights that transfer connections play a larger role when tasks share dynamical similarities.

\subsection{Real-data experiments}
\label{AP:Gait_tasks}
Having evaluated CRUG on synthetic dynamical system benchmarks, we next applied it to real-world data. We used the \textit{Gait in Neurodegenerative Disease Database}, publicly available through PhysioNet \citep{Hausdorff2000,dataset}. It contains force-sensitive resistor recordings from 64 subjects walking at their usual pace for approximately 5 minutes, sampled at 300~Hz. The subjects belong to four groups: patients with Parkinson's disease (PD; $n=15$), Huntington's disease (HD; $n=20$), or amyotrophic lateral sclerosis (ALS; $n=13$), and healthy controls ($n=16$).

The two foot sensors provide a two-dimensional observation of each subject's gait dynamics. To obtain a richer representation of the underlying dynamics, we augmented these observations with four delay-embedding features, yielding six observed dimensions in total. We discarded the first and last 20~s of each recording, decimated the signals to 75~Hz, and split each recording into 14{,}000 training and 5{,}000 test time steps. The trajectories were divided into sequences of 200 time steps and presented to the model in minibatches of 16. We treated each subject as a separate task and trained a single AL-RNN continually across subjects, with 2{,}000 training epochs per subject. As in the synthetic-data experiments, we used STF with a forcing interval of 16 time steps and a forcing strength of $\alpha=1.0$ \citep{Mikhaeil2022, Brenner2022, Hess2023}.

Importantly, the model received no clinical group labels or other subject-level metadata; only the two foot-sensor recordings were used. We selected 12 subjects in total, three from each group, and trained an AL-RNN with $(M,P) = (1024,512)$ units, sequentially in the order
$\text{Control}_1 \rightarrow \text{Parkinson}_1 \rightarrow
\text{Huntington}_1 \rightarrow \text{ALS}_1 \rightarrow
\text{Control}_2 \rightarrow \cdots$. After training on all 12 subjects, we evaluated the reconstruction of each subject on its test trajectory. To quantify performance, we computed the GMM-based approximation of the state space divergence, $\widetilde D_{\mathrm{stsp}}$, between ground-truth and reconstructed trajectories, as well as the Hellinger distance $D_H$ between their power spectra (see Appendix~\ref{AP:Evaluation_Meausres}).

\begin{figure}[htbp!]
    \centering
    \includegraphics[width=\linewidth]{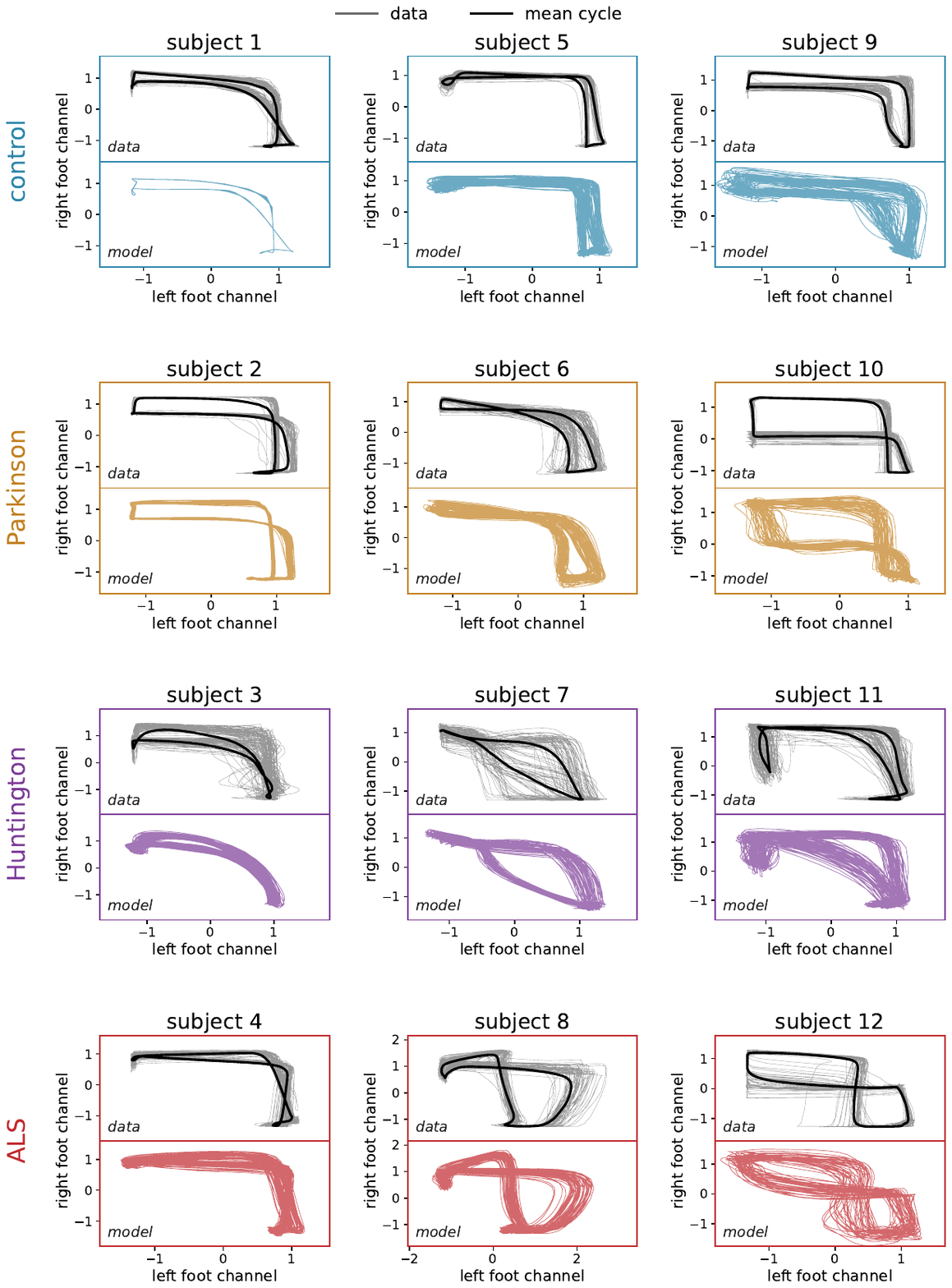}
    \caption{\textbf{Reconstruction of gait dynamics.} An AL-RNN using CRUG method was trained on 12 subjects from the gait dataset, three per group, in the order $\text{Control}_1 \rightarrow \text{Parkinson}_1 \rightarrow \text{Huntington}_1 \rightarrow \text{ALS}_1 \rightarrow \text{Control}_2 \rightarrow \cdots$. After all 12 subjects had been learned sequentially, the final model was used to reconstruct each subject's trajectories. For each subject, the top panel shows the test data (gray) with the mean trajectory (black), and the colored trajectory in bottom panel shows the model's reconstruction, demonstrating that the model successfully learned all subjects.}
    \label{fig:gait_reconstruction}
\end{figure}

\paragraph{Continual reconstruction across subjects.}
We optimized the CRUG hyperparameters, with the $L_2$ transfer regularization, with Optuna over 40 trials with three random seeds each. The results are reported in Table~\ref{tab:gait_joint_optuna}. The model learned all 12 subjects sequentially and reproduced the subject-specific gait dynamics (Fig.~\ref{fig:gait_reconstruction}). This demonstrates that CRUG can be applied to continual dynamical systems reconstruction from real-world recordings, maintaining multiple subject-specific dynamical models within a single RNN.


\begin{table}[t]
\centering
\renewcommand{\arraystretch}{1.3}
\setlength{\tabcolsep}{4pt}
\footnotesize

\caption{\textbf{Hyperparameter search for CRUG on the gait dataset.}
We ran a hyperparameter search with 40 trials (3 seeds each); the nine best trials are shown, ranked by overall $\widetilde D_{\mathrm{stsp}}$ aggregated over all 12 subjects. Entries show $\text{median}^{\,Q_3-\text{median}}_{\,Q_1-\text{median}}$ across seeds. Per-group columns report the median $\widetilde D_{\mathrm{stsp}}$ over the three subjects in each group (PD: Parkinson's disease; HD: Huntington's disease; ALS: amyotrophic lateral sclerosis). Overall $D_H$ and the total number of kept units indicate temporal agreement and capacity usage, respectively. Relative forgetting is the median over subjects of the relative increase in $\widetilde D_{\mathrm{stsp}}$ from immediately after training on a subject to the end of the sequence; positive values indicate forgetting. The selected configuration is highlighted in bold.}
\label{tab:gait_joint_optuna}

\resizebox{\textwidth}{!}{%
\begin{tabular}{c|cccc|cccc}
\hline
& \multicolumn{4}{c|}{{\bfseries\boldmath $\widetilde D_{\mathrm{stsp}}$ per group}} &
\multicolumn{4}{c}{\textbf{Overall}} \\
\cline{2-5}\cline{6-9}
{\boldmath$(\lambda_{\mathrm{ReLU}},\, \lambda_{\mathrm{lin}}/\lambda_{\mathrm{ReLU}},\, \lambda_{\mathrm{tr}})$} &
\textbf{Control} & \textbf{PD} & \textbf{HD} & \textbf{ALS} &
{\boldmath$\widetilde D_{\mathrm{stsp}}$} & {\boldmath$D_H$} & \textbf{Units} & \textbf{Rel.\ forgetting (\%)} \\
\hline
&&&&&&&&\\
$(1.12\times10^{-3},\, 0.20,\, 1.70\times10^{-4})$ & $1.72^{+0.34}_{-0.13}$ & $2.35^{+0.01}_{-0.17}$ & $4.25^{+0.46}_{-0.09}$ & $1.67^{+0.06}_{-0.01}$ & $2.04^{+0.16}_{-0.03}$ & $0.041^{+0.002}_{-0.000}$ & $730^{+2}_{-42}$ & $+0.0^{+0.2}_{-0.0}$ \\
$(1.09\times10^{-3},\, 0.21,\, 3.63\times10^{-5})$ & $1.97^{+0.25}_{-0.12}$ & $2.07^{+0.03}_{-0.08}$ & $5.16^{+0.26}_{-0.73}$ & $1.54^{+0.09}_{-0.00}$ & $2.07^{+0.04}_{-0.02}$ & $0.046^{+0.001}_{-0.004}$ & $690^{+31}_{-14}$ & $+0.3^{+0.1}_{-1.5}$ \\
$(1.09\times10^{-3},\, 0.21,\, 2.83\times10^{-4})$ & $1.88^{+0.20}_{-0.10}$ & $2.28^{+0.03}_{-0.14}$ & $4.49^{+0.22}_{-0.17}$ & $1.62^{+0.12}_{-0.10}$ & $2.10^{+0.10}_{-0.08}$ & $0.044^{+0.004}_{-0.002}$ & $741^{+4}_{-6}$ & $+0.0^{+0.1}_{-0.0}$ \\
$(2.06\times10^{-3},\, 0.30,\, 2.33\times10^{-4})$ & $2.24^{+0.09}_{-0.22}$ & $2.08^{+0.03}_{-0.08}$ & $4.51^{+0.24}_{-0.25}$ & $1.81^{+0.06}_{-0.07}$ & $2.19^{+0.11}_{-0.05}$ & $0.047^{+0.001}_{-0.001}$ & $555^{+7}_{-16}$ & $-0.4^{+0.2}_{-0.1}$ \\
{\boldmath$(2.52\times10^{-3},\, 0.36,\, 5.84\times10^{-4})$} & {\boldmath$1.65^{+0.22}_{-0.00}$} & {\boldmath$2.25^{+0.14}_{-0.10}$} & {\boldmath$4.94^{+0.11}_{-0.50}$} & {\boldmath$1.89^{+0.04}_{-0.05}$} & {\boldmath$2.20^{+0.06}_{-0.06}$} & {\boldmath$0.043^{+0.002}_{-0.001}$} & {\boldmath$487^{+12}_{-0}$} & {\boldmath$+0.0^{+0.2}_{-0.0}$} \\
$(2.69\times10^{-3},\, 0.25,\, 3.04\times10^{-5})$ & $1.83^{+0.08}_{-0.04}$ & $2.54^{+0.03}_{-0.12}$ & $4.48^{+0.90}_{-0.20}$ & $1.97^{+0.09}_{-0.04}$ & $2.27^{+0.06}_{-0.01}$ & $0.045^{+0.005}_{-0.002}$ & $504^{+16}_{-0}$ & $-0.2^{+0.2}_{-0.0}$ \\
$(1.88\times10^{-3},\, 0.28,\, 5.08\times10^{-4})$ & $1.97^{+0.26}_{-0.06}$ & $2.47^{+0.01}_{-0.17}$ & $4.00^{+0.93}_{-0.22}$ & $1.94^{+0.00}_{-0.05}$ & $2.27^{+0.11}_{-0.07}$ & $0.047^{+0.001}_{-0.002}$ & $561^{+16}_{-1}$ & $+0.0^{+0.0}_{-0.5}$ \\
$(1.98\times10^{-3},\, 0.31,\, 2.01\times10^{-4})$ & $2.02^{+0.42}_{-0.23}$ & $2.48^{+0.07}_{-0.11}$ & $5.07^{+1.12}_{-0.42}$ & $1.96^{+0.15}_{-0.02}$ & $2.30^{+0.17}_{-0.03}$ & $0.048^{+0.000}_{-0.003}$ & $549^{+2}_{-2}$ & $+0.1^{+1.3}_{-0.4}$ \\
$(1.80\times10^{-3},\, 0.29,\, 3.45\times10^{-4})$ & $1.91^{+0.00}_{-0.07}$ & $2.47^{+0.00}_{-0.07}$ & $4.21^{+0.51}_{-0.06}$ & $1.85^{+0.06}_{-0.06}$ & $2.32^{+0.02}_{-0.05}$ & $0.043^{+0.003}_{-0.001}$ & $571^{+6}_{-22}$ & $+0.0^{+0.5}_{-0.1}$ \\
\hline
\end{tabular}%
}
\end{table}

\paragraph{Joint versus group-specific models.}
A key practical advantage of learning all subjects within a single continually trained model is that the clinical group of a newly observed subject need not be known. In contrast, maintaining separate models for the four groups would require the group label to determine which model to update.
To compare the two approaches, we additionally trained four group-specific models: each on the same three subjects of its group as used for the joint model. The hyperparameters of each group model were again tuned with Optuna over 40 trials with three seeds each; the best trials for each group, ranked by overall $\widetilde D_{\mathrm{stsp}}$, are reported in Table~\ref{tab:gait_groups_hyperparameters}. Table~\ref{tab:gait_joint_vs_group} compares the best configuration of each group model with the selected best joint model. The joint model attains a lower median $\widetilde D_{\mathrm{stsp}}$ for every group, with significant differences for PD, HD and ALS, and it uses fewer units for every group.

\begin{table}[htbp!]
\centering
\renewcommand{\arraystretch}{1.3}
\setlength{\tabcolsep}{4pt}
\footnotesize

\caption{\textbf{Hyperparameter search for CRUG on the individual subject groups of the gait dataset.} Each block shows the best configurations for one subject group, trained sequentially on its three subjects and ranked by overall $\widetilde D_{\mathrm{stsp}}$. Entries show $\text{median}^{\,Q_3-\text{median}}_{\,Q_1-\text{median}}$ across 3 seeds.}
\label{tab:gait_groups_hyperparameters}

\begin{tabular}{c|ccc}
\hline
{\boldmath$(\lambda_{\mathrm{ReLU}},\, \lambda_{\mathrm{lin}}/\lambda_{\mathrm{ReLU}},\, \lambda_{\mathrm{tr}})$} &
{\boldmath$D_{\mathrm{stsp}}$} & {\boldmath$D_H$} & \textbf{Units} \\
\hline
\multicolumn{4}{l}{\textbf{(a) Control}} \\
\hline
&&&\\
{\boldmath$(1.09\times10^{-3},\, 0.21,\, 4.63\times10^{-4})$} & {\boldmath$1.91^{+0.03}_{-0.06}$} & {\boldmath$0.040^{+0.001}_{-0.001}$} & {\boldmath$176^{+0.5}_{-14.5}$} \\
$(1.02\times10^{-3},\, 0.22,\, 8.81\times10^{-4})$ & $1.92^{+0.28}_{-0.00}$ & $0.044^{+0.005}_{-0.005}$ & $163^{+7.0}_{-4.0}$ \\
$(1.08\times10^{-3},\, 0.20,\, 1.93\times10^{-4})$ & $1.94^{+0.16}_{-0.02}$ & $0.045^{+0.002}_{-0.002}$ & $164^{+6.5}_{-2.5}$ \\
$(1.28\times10^{-3},\, 0.26,\, 5.13\times10^{-4})$ & $1.95^{+0.16}_{-0.07}$ & $0.041^{+0.001}_{-0.001}$ & $151^{+2.5}_{-0.0}$ \\
$(1.18\times10^{-3},\, 0.22,\, 3.60\times10^{-3})$ & $2.00^{+0.12}_{-0.09}$ & $0.040^{+0.002}_{-0.001}$ & $158^{+1.5}_{-4.5}$ \\
$(1.34\times10^{-3},\, 0.29,\, 3.88\times10^{-3})$ & $2.02^{+0.22}_{-0.01}$ & $0.043^{+0.002}_{-0.002}$ & $141^{+4.5}_{-5.0}$ \\
$(1.01\times10^{-3},\, 0.22,\, 4.63\times10^{-3})$ & $2.03^{+0.06}_{-0.03}$ & $0.046^{+0.000}_{-0.001}$ & $170^{+7.5}_{-2.0}$ \\
$(1.09\times10^{-3},\, 0.21,\, 5.32\times10^{-4})$ & $2.06^{+0.05}_{-0.01}$ & $0.046^{+0.000}_{-0.005}$ & $168^{+1.0}_{-11.5}$ \\
\\
\multicolumn{4}{l}{\textbf{(b) Parkinson's disease}} \\
\hline
&&&\\
{\boldmath$(1.03\times10^{-3},\, 0.24,\, 3.77\times10^{-4})$} & {\boldmath$2.88^{+0.21}_{-0.00}$} & {\boldmath$0.038^{+0.001}_{-0.000}$} & {\boldmath$173^{+3.0}_{-10.5}$} \\
$(1.10\times10^{-3},\, 0.26,\, 9.86\times10^{-5})$ & $2.89^{+0.08}_{-0.06}$ & $0.034^{+0.003}_{-0.000}$ & $165^{+10.5}_{-2.0}$ \\
$(1.42\times10^{-3},\, 0.21,\, 5.91\times10^{-5})$ & $2.91^{+0.11}_{-0.14}$ & $0.038^{+0.006}_{-0.001}$ & $156^{+0.5}_{-6.5}$ \\
$(1.12\times10^{-3},\, 0.24,\, 9.51\times10^{-5})$ & $2.92^{+0.02}_{-0.12}$ & $0.036^{+0.005}_{-0.001}$ & $160^{+10.0}_{-0.5}$ \\
$(1.00\times10^{-3},\, 0.24,\, 5.61\times10^{-4})$ & $2.94^{+0.06}_{-0.05}$ & $0.048^{+0.002}_{-0.006}$ & $173^{+2.5}_{-17.5}$ \\
$(1.43\times10^{-3},\, 0.21,\, 6.42\times10^{-2})$ & $2.96^{+0.11}_{-0.05}$ & $0.036^{+0.001}_{-0.000}$ & $147^{+5.0}_{-2.5}$ \\
$(1.39\times10^{-3},\, 0.20,\, 2.68\times10^{-3})$ & $2.98^{+0.05}_{-0.02}$ & $0.033^{+0.003}_{-0.001}$ & $157^{+2.5}_{-3.0}$ \\
$(1.00\times10^{-3},\, 0.24,\, 4.90\times10^{-4})$ & $2.99^{+0.06}_{-0.00}$ & $0.039^{+0.000}_{-0.002}$ & $180^{+3.5}_{-10.5}$ \\
$(1.27\times10^{-3},\, 0.74,\, 1.70\times10^{-4})$ & $3.00^{+0.02}_{-0.00}$ & $0.037^{+0.000}_{-0.001}$ & $135^{+2.5}_{-2.0}$ \\
\\
\multicolumn{4}{l}{\textbf{(c) Huntington's disease}} \\
\hline
&&&\\
{\boldmath$(1.09\times10^{-3},\, 0.21,\, 3.69\times10^{-5})$} & {\boldmath$5.52^{+1.78}_{-0.23}$} & {\boldmath$0.046^{+0.024}_{-0.001}$} & {\boldmath$221^{+8.0}_{-3.5}$} \\
$(1.32\times10^{-3},\, 0.35,\, 4.03\times10^{-5})$ & $5.89^{+0.29}_{-0.05}$ & $0.062^{+0.007}_{-0.006}$ & $190^{+0.0}_{-6.0}$ \\
$(1.62\times10^{-3},\, 0.73,\, 1.12\times10^{-4})$ & $5.89^{+0.31}_{-0.21}$ & $0.052^{+0.006}_{-0.002}$ & $173^{+9.0}_{-1.5}$ \\
$(1.49\times10^{-3},\, 0.48,\, 6.09\times10^{-5})$ & $5.95^{+0.12}_{-0.34}$ & $0.061^{+0.005}_{-0.006}$ & $183^{+3.5}_{-12.0}$ \\
$(1.09\times10^{-3},\, 0.21,\, 4.63\times10^{-4})$ & $5.95^{+0.20}_{-0.04}$ & $0.067^{+0.000}_{-0.004}$ & $227^{+8.5}_{-0.0}$ \\
$(1.46\times10^{-3},\, 0.25,\, 2.03\times10^{-2})$ & $5.98^{+0.81}_{-0.21}$ & $0.053^{+0.002}_{-0.003}$ & $194^{+3.0}_{-5.5}$ \\
$(1.12\times10^{-3},\, 0.29,\, 4.29\times10^{-5})$ & $6.06^{+0.30}_{-0.36}$ & $0.049^{+0.009}_{-0.000}$ & $216^{+1.5}_{-6.5}$ \\
$(1.16\times10^{-3},\, 0.25,\, 3.38\times10^{-5})$ & $6.12^{+0.20}_{-0.14}$ & $0.072^{+0.001}_{-0.003}$ & $219^{+4.0}_{-9.5}$ \\
$(2.10\times10^{-3},\, 0.32,\, 4.53\times10^{-4})$ & $6.15^{+0.63}_{-0.36}$ & $0.068^{+0.006}_{-0.000}$ & $154^{+1.5}_{-0.0}$ \\
\\
\multicolumn{4}{l}{\textbf{(d) Amyotrophic lateral sclerosis}} \\
\hline
&&&\\
{\boldmath$(1.64\times10^{-3},\, 0.40,\, 9.79\times10^{-5})$} & {\boldmath$2.50^{+0.26}_{-0.02}$} & {\boldmath$0.037^{+0.005}_{-0.000}$} & {\boldmath$154^{+0.5}_{-5.0}$} \\
$(1.09\times10^{-3},\, 0.22,\, 3.63\times10^{-5})$ & $2.50^{+0.28}_{-0.06}$ & $0.035^{+0.001}_{-0.001}$ & $184^{+5.0}_{-4.5}$ \\
$(1.09\times10^{-3},\, 0.21,\, 3.69\times10^{-5})$ & $2.52^{+0.35}_{-0.02}$ & $0.040^{+0.001}_{-0.001}$ & $175^{+0.0}_{-16.5}$ \\
$(1.24\times10^{-3},\, 0.38,\, 6.48\times10^{-5})$ & $2.55^{+0.20}_{-0.04}$ & $0.037^{+0.006}_{-0.001}$ & $148^{+17.0}_{-0.0}$ \\
$(1.25\times10^{-3},\, 0.56,\, 6.19\times10^{-5})$ & $2.59^{+0.19}_{-0.00}$ & $0.035^{+0.000}_{-0.001}$ & $167^{+0.5}_{-3.0}$ \\
$(1.64\times10^{-3},\, 0.44,\, 1.13\times10^{-4})$ & $2.60^{+0.31}_{-0.03}$ & $0.039^{+0.000}_{-0.001}$ & $140^{+1.5}_{-5.0}$ \\
$(1.25\times10^{-3},\, 0.80,\, 5.93\times10^{-5})$ & $2.61^{+0.18}_{-0.03}$ & $0.039^{+0.001}_{-0.003}$ & $161^{+8.5}_{-2.5}$ \\
$(1.52\times10^{-3},\, 0.44,\, 9.27\times10^{-5})$ & $2.61^{+0.00}_{-0.04}$ & $0.039^{+0.005}_{-0.000}$ & $147^{+2.5}_{-7.5}$ \\
$(1.62\times10^{-3},\, 0.73,\, 1.12\times10^{-4})$ & $2.63^{+0.35}_{-0.09}$ & $0.039^{+0.001}_{-0.001}$ & $143^{+2.5}_{-6.5}$ \\

\end{tabular}
\end{table}

\begin{table}[htbp!]
\centering
\renewcommand{\arraystretch}{1.3}
\setlength{\tabcolsep}{4pt}
\footnotesize

\caption{\textbf{Joint versus per-group training of CRUG on the gait dataset.} Joint: a single model trained sequentially on all 12 subjects. Group: a separate model trained for each subject group. Entries show $\text{median}^{\,Q_3-\text{median}}_{\,Q_1-\text{median}}$; $\widetilde D_{\mathrm{stsp}}$ is pooled over subjects and seeds (9 values per group, 36 for all subjects), and kept units over seeds (3 values). $p$-values are from two-sided Mann--Whitney $U$ tests comparing joint and per-group training; with three seeds per condition, $0.10$ is the smallest attainable $p$-value for kept units.}
\label{tab:gait_joint_vs_group}
\begin{tabular}{l|ccc|ccc}
\hline
& \multicolumn{3}{c|}{\boldmath$\widetilde D_{\mathrm{stsp}}$} & \multicolumn{3}{c}{\textbf{Kept units}} \\
\cline{2-4}\cline{5-7}
\textbf{Group} & \textbf{Joint} & \textbf{Group} & \boldmath$p$ & \textbf{Joint} & \textbf{Group} & \boldmath$p$ \\
\hline
Control         & $\mathbf{1.65}^{+1.20}_{-0.04}$ & $1.91^{+0.70}_{-0.09}$ & 0.80   & $\mathbf{113}^{+0.0}_{-2.0}$  & $176^{+0.5}_{-14.5}$ & 0.10 \\
PD              & $\mathbf{2.25}^{+0.38}_{-0.76}$ & $2.95^{+0.27}_{-0.08}$ & 0.040  & $\mathbf{118}^{+3.0}_{-7.0}$  & $173^{+3.0}_{-10.5}$ & 0.10 \\
HD              & $\mathbf{4.94}^{+0.86}_{-0.82}$ & $5.83^{+4.48}_{-0.78}$ & 0.077  & $\mathbf{152}^{+2.0}_{-9.0}$  & $221^{+8.0}_{-3.5}$  & 0.10 \\
ALS             & $\mathbf{1.84}^{+0.12}_{-0.18}$ & $2.56^{+0.53}_{-0.10}$ & 0.0005 & $\mathbf{120}^{+2.0}_{-1.0}$  & $154^{+0.5}_{-5.0}$  & 0.10 \\
\hline
All 12 subjects & $\mathbf{2.12}^{+1.20}_{-0.47}$ & $2.92^{+0.79}_{-0.60}$ & 0.021  & $\mathbf{487}^{+12.0}_{-0.0}$ & $697^{+19.5}_{-0.5}$ & 0.10 \\
\hline
\end{tabular}
\end{table}
Together, these results show that CRUG can learn multiple subjects continually within a single RNN without requiring clinical group labels. The joint model can share information across subjects without prior knowledge of their similarity, while using less total capacity than separate group-specific models. This provides a practical framework for incrementally incorporating new recordings into a surrogate dynamical model, without retraining it from scratch whenever a new recording becomes available.

\subsection{CRUG on Cognitive Tasks}
\label{AP:Cognitive_Tasks}

Finally, to demonstrate that our method is not specific to continual DSR tasks, we evaluated it on a battery of nine sequential cognitive tasks introduced by \citet{Driscoll2024}: delayed-response pro/anti, reaction-time pro/anti, category-response pro/anti, context-integration pro/anti, and Go/NoGo. Together, these tasks probe a range of computational demands, including working memory, stimulus--response mapping, categorical decision making, evidence integration, and response inhibition.

\paragraph{Task definitions.} In the delayed-response pro/anti tasks, a directional stimulus is presented briefly and must be retained over a subsequent delay period. At the response cue, the model must report either the remembered stimulus direction (pro) or the opposite direction (anti). The reaction-time pro/anti tasks use the same directional mappings but remove the memory requirement: the model must respond immediately after stimulus onset, toward the stimulus direction (pro) or away from it (anti). In the category-response pro/anti tasks, the continuous stimulus angle $\theta$ is assigned to one of two semicircular categories, and the model must report the category containing the stimulus (pro) or the opposite category (anti). The context-integration pro/anti tasks present two noisy input modalities together with a contextual cue indicating which modality is relevant. The model must integrate evidence from the cued modality while ignoring the other, and produce either the corresponding directional response (pro) or its opposite (anti). Finally, in the Go/NoGo task, the model must determine whether the stimulus amplitude exceeds a predefined threshold. On Go trials, it responds in a fixed direction; on NoGo trials, it withholds the response and maintains fixation.

\paragraph{Initialization and training.}
We used an AL-RNN with $M=120$ units, comprising 60 linear and 60 nonlinear units. The model was trained sequentially on all nine tasks for 50 epochs per task. Each task comprised 1{,}000 training trials and 200 evaluation trials, and training used minibatches of 64 trials.
Each trial consists of a fixation phase followed by stimulus presentation and, depending on the task, either a delay period or an immediate response period. The input at each time step ($s_t$ in Eq.~\ref{eq:al-rnn}) consists of the nine raw stimulus channels (channel 0: fixation $f_t$; channels 1 and 2: $\sin\theta$ and $\cos\theta$; channels 3--8: task-rule flags), concatenated with a task-identity vector. There is no task-specific encoder, and the latent state is initialized at zero. Only the first three channels are used as outputs. For CRUG and CLNP, the readout for task $k$ is a fixed random matrix $D_k$. The remaining methods (Naive, EWC, SI, ER, GR, and XdG) do not partition the network and therefore use a single fixed random readout that is dense over all $M$ units and shared across all nine tasks.

Parameters are initialized as $A_i \sim \mathcal{N}(0.9, 0.1^2)$,
$W_{ij} \sim \mathcal{N}(0, (0.1/\sqrt{M})^2)$, $C_{ij}, (D_k)_{ij} \sim \mathcal{N}(0, 0.1^2)$,
and $h = 0$. Released units are reset to this untrained state before training on the next task. For CRUG, we applied $L_2$ regularization to the transfer connections.

\paragraph{Evaluation.} Performance is measured as the fraction of correctly answered trials,
\begin{align}
\mathrm{acc} = \frac{1}{T} \sum_{t=1}^{T} \mathbf{1}_{\{\text{trial } t \text{ correct}\}},
\label{eq:acc_cognitive}
\end{align}
where $T$ denotes the number of trials. At the final time step, the network outputs
$\hat{y}_t = (\hat{f}_t, \sin\hat{\theta}, \cos\hat{\theta})$, i.e., a fixation channel and a two-dimensional response direction. The correctness criterion is task-specific.

For the eight tasks whose response is a direction (delayed response, reaction time, category response, and context integration, each in its pro and anti variant), a trial is counted as correct if the circular distance between the predicted and target response angles is below $\delta = 30^\circ$. Although the two category tasks are binary, their categories are encoded as antipodal target directions, so the same angular criterion implements the categorical decision.

For Go/NoGo, the decision is carried by the fixation channel rather than by the response direction: the target is $f_t = 0$ on Go trials and $f_t = 1$ on NoGo trials. Additionally on Go trials, a trial is counted as correct if the predicted decision $\mathbf{1}_{\{\hat{f}_t > 0.5\}}$ matches the target, while the NoGo target direction is the zero vector and therefore carries no class information.

\paragraph{Hyperparameter tuning and results.} For each method, we performed an Optuna hyperparameter search with 25 trials for EWC, SI, XdG, ER, and GR, and 40 trials each for CLNP and CRUG, using 10 seeds per trial (Table~\ref{tab:hyperparameter_cognitive}). The best result for each method is reported in Table~\ref{tab:cognitive_tasks}. Naive fine-tuning exhibits substantial catastrophic forgetting, whereas all continual learning methods mitigate forgetting to varying degrees. This suggests that preserving performance across these cognitive tasks is less challenging than preserving long-term autonomous dynamics in the DSR benchmarks. Notably, the regularization-based methods EWC and SI, as well as the parameter-isolation method XdG, perform reasonably well on the cognitive tasks despite being largely ineffective in the DSR setting. The replay-based methods achieve higher accuracy still, and the two remaining parameter-isolation methods, CLNP and CRUG, attain the highest final accuracies. Although CLNP reaches accuracy comparable to CRUG, it is considerably less reliable, as 4 of 10 seeds exhausted the unit pool before completing the task sequence, and it requires substantially more network capacity (Table~\ref{tab:cognitive_tasks}).

\begin{table}[htbp!]
\caption{\textbf{Benchmarking CRUG against other continual learning methods on sequential cognitive tasks.} The model is trained sequentially on nine cognitive tasks. Naive fine-tuning, without any continual learning mechanism, exhibits pronounced catastrophic forgetting. All continual learning methods substantially outperform naive fine-tuning, although their effectiveness varies: EWC, SI and XdG show the largest performance degradation, the replay-based methods (ER, GR) perform considerably better, and the capacity-allocating methods perform best. CRUG and CLNP achieve comparable accuracy, but CRUG attains the highest mean final accuracy while using the least network capacity (58.4\%), whereas CLNP occupies almost the entire network (94.2\%). For CLNP, 4 of the 10 seeds exhausted the unit pool before completing the task sequence and therefore produced no final model; these runs are excluded from the reported mean and standard deviation. Capacity usage is not reported (``--'') for methods that do not isolate task-specific units and instead operate on all recurrent units.}
\label{tab:cognitive_tasks}

\begin{center}
\renewcommand{\arraystretch}{1.2}
\begin{tabular}{lcc}
\multicolumn{1}{c}{\bf Method} &\multicolumn{1}{c}{\bf Accuracy (\%) $\uparrow$} &\multicolumn{1}{c}{\bf Used Capacity (\%) $\downarrow$}
\\ \hline \\
Naive & $40.19 \pm 5.00$  & -- \\
EWC   & $93.78 \pm 7.17$  & -- \\
SI    & $80.74 \pm 13.89$ & -- \\
ER    & $98.11 \pm 0.98$  & -- \\
GR    & $97.46 \pm 1.69$  & -- \\
XdG   & $79.14 \pm 6.40$  & -- \\
CLNP $(6/10)$ & $98.79 \pm 0.27$ & $94.2 \pm 3.7$ \\
\hline
CRUG  & $99.26 \pm 0.23$  & $58.4 \pm 6.9$ \\
\hline
\end{tabular}
\end{center}
\end{table}

\begin{table}[t]
\centering\scriptsize
\caption{\textbf{Hyperparameter tuning of continual learning methods on the cognitive-task benchmark.} For each seed, accuracy is averaged over the nine tasks after training on the full task sequence. Values are reported as mean $\pm$ standard error of the mean across seeds, and rows are ranked by mean accuracy; only the ten best trials of each study are shown. Each study comprises 25 trials for EWC, SI, XdG, ER, and GR, and 40 trials for CLNP and CRUG, with 10 seeds per trial. Seeds that exhausted the available network capacity are excluded; for CLNP the number of contributing seeds varies between 5 and 9 across the listed trials. All listed CRUG configurations completed on all ten seeds. Bold rows mark the configurations reported in Table~\ref{tab:cognitive_tasks}.}
\label{tab:hyperparameter_cognitive}

\begin{minipage}[t]{0.32\textwidth}\centering
\textbf{(a) EWC}\\[2pt]
\begin{tabular}{cc}
\hline
$\lambda_{\mathrm{EWC}}$ & Final accuracy (\%) \\
\hline\\
$4136$  & $\boldsymbol{93.78\pm7.17}$ \\
$6107$  & $93.33\pm8.64$ \\
$19550$ & $93.22\pm8.14$ \\
$1963$  & $92.57\pm4.93$ \\
$9806$  & $92.51\pm8.69$ \\
$3230$  & $92.26\pm7.12$ \\
$20490$ & $91.96\pm10.84$ \\
$1859$  & $91.59\pm3.69$ \\
$7505$  & $91.53\pm14.05$ \\
$7220$  & $91.25\pm16.09$ \\
\hline
\end{tabular}
\end{minipage}\hfill
\begin{minipage}[t]{0.32\textwidth}\centering
\textbf{(b) SI}\\[2pt]
\begin{tabular}{cc}
\hline
$\lambda_{\mathrm{SI}}$ & Final accuracy (\%) \\
\hline\\
$0.4608$ & $\boldsymbol{80.74\pm13.89}$ \\
$0.6308$ & $79.72\pm14.01$ \\
$0.4903$ & $79.64\pm12.95$ \\
$0.3630$ & $79.39\pm13.95$ \\
$0.5845$ & $78.68\pm14.20$ \\
$0.1538$ & $78.52\pm12.99$ \\
$0.5652$ & $78.51\pm13.45$ \\
$0.1073$ & $77.53\pm13.04$ \\
$2.037$  & $77.40\pm14.30$ \\
$0.1033$ & $77.24\pm11.91$ \\
\hline
\end{tabular}
\end{minipage}\hfill
\begin{minipage}[t]{0.32\textwidth}\centering
\textbf{(c) XdG}\\[2pt]
\begin{tabular}{cc}
\hline
$p_{\mathrm{gate}}$ & Final accuracy (\%) \\
\hline\\
$0.9347$ & $\boldsymbol{79.14\pm6.40}$ \\
$0.9075$ & $79.06\pm5.04$ \\
$0.9366$ & $78.45\pm6.73$ \\
$0.9405$ & $78.23\pm6.70$ \\
$0.8404$ & $77.18\pm3.40$ \\
$0.9453$ & $76.99\pm6.64$ \\
$0.9173$ & $76.79\pm5.30$ \\
$0.8021$ & $75.98\pm4.26$ \\
$0.8728$ & $75.73\pm4.67$ \\
$0.8526$ & $75.16\pm4.50$ \\
\hline
\end{tabular}
\end{minipage}

\par\bigskip

\begin{minipage}[t]{0.48\textwidth}\centering
\textbf{(d) ER}\\[2pt]
\begin{tabular}{cc}
\hline
Replay ratio & Final accuracy (\%) \\
\hline\\
$0.4434$ & $\boldsymbol{98.11\pm0.98}$ \\
$0.1995$ & $98.04\pm1.04$ \\
$0.2338$ & $97.99\pm1.33$ \\
$0.4052$ & $97.89\pm0.91$ \\
$0.1661$ & $97.62\pm1.37$ \\
$0.5180$ & $97.59\pm1.32$ \\
$0.2907$ & $97.58\pm1.66$ \\
$0.3497$ & $97.57\pm1.30$ \\
$0.4475$ & $97.52\pm1.37$ \\
$0.3518$ & $97.50\pm1.61$ \\
\hline
\end{tabular}
\end{minipage}\hfill
\begin{minipage}[t]{0.48\textwidth}\centering
\textbf{(e) GR}\\[2pt]
\begin{tabular}{cc}
\hline
Replay ratio & Final accuracy (\%) \\
\hline\\
$0.1916$ & $\boldsymbol{97.46\pm1.69}$ \\
$0.3815$ & $97.42\pm1.96$ \\
$0.1837$ & $97.41\pm1.82$ \\
$0.1644$ & $97.40\pm2.17$ \\
$0.2609$ & $97.38\pm2.52$ \\
$0.3381$ & $97.29\pm2.25$ \\
$0.2907$ & $97.27\pm2.96$ \\
$0.2830$ & $97.24\pm2.00$ \\
$0.1486$ & $97.20\pm2.62$ \\
$0.5800$ & $97.15\pm2.08$ \\
\hline
\end{tabular}
\end{minipage}

\vspace{10pt}

\begin{minipage}[t]{\textwidth}\centering
\textbf{(f) CLNP}\\[2pt]
\begin{tabular}{ccccc}
\hline
$\alpha_{\mathrm{ReLU}}$ & Linear:ReLU ratio & $m$ & Capacity (\%) & Final accuracy (\%) \\
\hline\\
$0.050$ & $0.942$ & $0.010$ & $94.17\pm3.73$ & $\boldsymbol{98.79\pm0.27}$ \\
$0.061$ & $0.593$ & $0.006$ & $94.27\pm7.54$ & $98.49\pm0.74$ \\
$0.072$ & $0.658$ & $0.010$ & $94.52\pm6.84$ & $98.46\pm0.39$ \\
$0.055$ & $0.975$ & $0.009$ & $97.08\pm2.46$ & $98.40\pm0.69$ \\
$0.133$ & $0.797$ & $0.007$ & $93.61\pm4.79$ & $97.85\pm1.16$ \\
$0.103$ & $0.951$ & $0.011$ & $90.28\pm8.19$ & $97.85\pm1.31$ \\
$0.181$ & $0.689$ & $0.019$ & $94.54\pm5.88$ & $97.79\pm0.94$ \\
$0.090$ & $0.975$ & $0.005$ & $97.67\pm2.31$ & $97.74\pm1.75$ \\
$0.121$ & $0.650$ & $0.014$ & $93.81\pm2.30$ & $97.67\pm0.94$ \\
$0.079$ & $0.881$ & $0.013$ & $95.71\pm3.89$ & $97.67\pm1.29$ \\
\hline
\end{tabular}
\end{minipage}

\vspace{10pt}

\begin{minipage}[t]{\textwidth}\centering
\textbf{(g) CRUG}\\[2pt]
\begin{tabular}{ccccc}
\hline
$\lambda_{\mathrm{ReLU}}$ & $\lambda_{\mathrm{lin}}$ & $\lambda_{\mathrm{tr}}$ ($\times10^{-3}$) & Capacity (\%) & Final accuracy (\%) \\
\hline\\
$0.087$ & $0.040$ & $5.42$  & $74.58\pm7.52$ & $99.38\pm0.29$ \\
$0.066$ & $0.028$ & $4.70$  & $88.75\pm6.67$ & $99.35\pm0.26$ \\
$0.091$ & $0.032$ & $0.66$  & $78.58\pm6.07$ & $99.33\pm0.23$ \\
$0.070$ & $0.046$ & $1.29$  & $76.67\pm8.64$ & $99.33\pm0.30$ \\
$0.088$ & $0.035$ & $1.01$  & $77.25\pm7.08$ & $99.32\pm0.25$ \\
$0.109$ & $0.042$ & $2.57$  & $68.75\pm5.14$ & $99.32\pm0.27$ \\
$0.069$ & $0.043$ & $0.97$  & $77.42\pm7.73$ & $99.31\pm0.25$ \\
$0.079$ & $0.038$ & $1.13$  & $75.92\pm7.05$ & $99.29\pm0.40$ \\
$0.063$ & $0.037$ & $23.72$ & $83.67\pm5.90$ & $99.29\pm0.28$ \\
$0.084$ & $0.030$ & $0.42$  & $82.92\pm5.92$ & $99.27\pm0.21$ \\
\hline
{\boldmath$0.133$} & {\boldmath$0.060$} & {\boldmath$3.94$} & {\boldmath$58.42\pm6.92$} & {\boldmath$99.26\pm0.23$} \\
\hline
\end{tabular}
\end{minipage}
\end{table}

\end{document}